\documentclass[10pt,journal,compsoc]{IEEEtran}

\ifCLASSOPTIONcompsoc
  \usepackage[nocompress]{cite}
\else
  \usepackage{cite}
\fi
\ifCLASSINFOpdf
\else
\fi
\usepackage{amsmath}
\usepackage{amsfonts}

\usepackage{hyperref}
\usepackage{graphics}
\usepackage{graphicx}
\usepackage{subfiles}
\usepackage{wrapfig}
\usepackage{cleveref}
\usepackage{tabularx}
\usepackage[dvipsnames]{xcolor}
\usepackage{optidef}
\usepackage{bm}
\usepackage{dsfont}
\usepackage[algo2e,ruled,linesnumbered]{algorithm2e}

\usepackage[a4paper,
            left=1.2cm,
            right=1.2cm, 
            top = 1.4cm,
            bottom = 1.4cm, footskip=.05cm]{geometry}

\usepackage{letltxmacro}
\LetLtxMacro{\originaleqref}{\eqref}
\renewcommand{\eqref}{Eq.~\originaleqref}
\renewcommand*{\eqref}[1]{Eq.~\originaleqref{#1}}

\usepackage{thmtools}

\newcommand\numberthis{\addtocounter{equation}{1}\tag{\theequation}}

\newcommand{\rtypo}[1]{#1}

\newcommand{\rref}[1]{#1}

\newcommand{\rother}[1]{#1}
\renewcommand\marginpar[1]{}

\usepackage{array,multirow}
 \usepackage{float}

\usepackage{amssymb}

\newcommand*{\fixedref}[1]{#1}

\crefformat{section}{\S#2#1#3} 
\crefformat{subsection}{\S#2#1#3}
\crefformat{subsubsection}{\S#2#1#3}

\newcommand{\etal}{\textit{et al.}}

\usepackage{upgreek} 
\renewcommand{\vector}[1]{\bm{\lowercase{#1}}}
\renewcommand{\matrix}[1]{\bm{\uppercase{#1}}}
\newcommand{\rv}[1]{\mathsf{#1}}

\begin{document}

\title{\rtypo{Scalable Optimal Transport Methods in Machine Learning: A Contemporary Survey}}

\author{Abdelwahed~Khamis, Russell~Tsuchida,
        Mohamed~Tarek, Vivien~Rolland, and~Lars~Petersson 
}

\markboth{IEEE Transactions on Pattern Analysis and Machine Intelligence, March.~2024}%
{Shell \MakeLowercase{\textit{et al.}}: Bare Demo of IEEEtran.cls for Computer Society Journals}

\IEEEtitleabstractindextext{%
\begin{abstract}

Optimal Transport (OT) is a mathematical framework that first emerged in the eighteenth century and has led to a plethora of methods for answering many theoretical and applied questions. The last decade has been a witness to the remarkable contributions of this classical optimization problem to machine learning. This paper is about where and how optimal transport is used in machine learning with a focus on the question of \rtypo{scalable} optimal transport. We provide a comprehensive survey of optimal transport while ensuring an accessible presentation as permitted by the nature of the topic and the context. First, we explain the optimal transport background and introduce different flavors (i.e. mathematical formulations), properties, and notable applications. We then address the fundamental question of how to scale optimal transport to cope with the current demands of big and high dimensional data. We conduct a systematic analysis of the methods used in the literature for scaling OT and present the findings in a unified taxonomy. We conclude with presenting some open challenges and discussing potential future research directions. A live repository of related OT research papers is maintained in \url{https://github.com/abdelwahed/OT_for_big_data.git}
\end{abstract}

}

\maketitle

\IEEEdisplaynontitleabstractindextext

\IEEEpeerreviewmaketitle

\vspace{-3em}
\section{Introduction}\label{sec:introduction}

Selecting the right notion of discrepancy or distance between objects is at the heart of many machine learning problems. 
This work is about \rref{Optimal Transport (OT) \cite{Peyre2018ComputationalTransport}}, a field that defines many such notions between a variety of mathematical objects such as probability histograms, shapes, and point clouds. 
OT views objects as heaps of sand (mass) and quantifies the distance between them as the \marginpar{R1} \rtypo{least costly way to transform the first heap into the second}. 
This seemingly abstract concept has found applications in problems including \rref{\marginpar{R1} domain adaptation \cite{Courty2017OptimalAdaptation}}
, object detection \cite{GeOta:Detection}, reinforcement learning \cite{Haldar2022WatchTransport}, \rref{graph representation and matching \cite{PeyreGABRIELPEYRE2016Gromov-WassersteinMatrices,XuScalableMatching, petric2019got}} 
, feature matching \cite{SunLoFTR:Transformers}, analyzing \rref{deep learning generalization \cite{ChuangMeasuringTransport,bach2022gradient, chizat2018global} }, \rref{generative modeling \cite{Arjovsky2017WassersteinNetworks}}
, knowledge distillation \cite{ChenWassersteinDistillation}, fairness \cite{GordalizaObtainingTheory,BuylOptimalFairness} and many others.

Interestingly, the simple idea of moving mass in an optimal way has a rich history. It evolved over the years, branched into many fields, and produced a wealth of theoretical and practical knowledge.
\marginpar{R1}\rtypo{The recent treatment of the topic was shaped by the work of Leonid Kantorovich when a practical wartime optimal allocation problem confronted him.  He realized that the problem was actually of broader interest and devised a ``simple general method of solving this group of problems`` \cite{Kantorovich1960MathematicalProduction} \footnote{\rtypo{His method is regarded as a variant of the simplex method proposed in the late 1940s by Dantzig; who advanced the field of linear programming \cite{vershik2013long}.}}.}
By doing so, he unknowingly \cite{Kantorovich2006ONMONGE} revisited a similar transport problem studied 150 years earlier by the renowned mathematician Gaspard Monge; the problem of soil transport for construction purposes. Ultimately, Kantorovich's solutions constituted the main tools of linear programming (i.e., linearly constrained linear optimization) and won him a Nobel Prize in economic sciences in 1975. This historical connection between the theory of OT and its applications persists until today, fostering successful developments on both fronts. On the theoretical side, the success can be associated with OT's connection with several branches of mathematics. The parallel practical success is reflected in the growing adoption of OT in diverse applications.

\begin{figure}[!t]
    \centering
    \includegraphics[width= 1\linewidth]{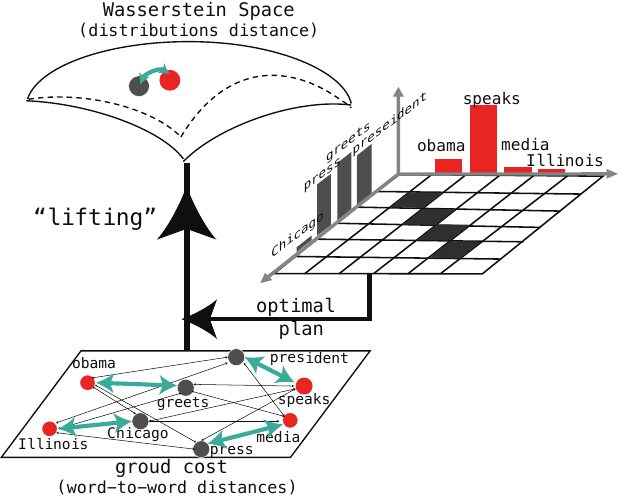}
    \vspace{-1.8em}
    \caption{ \textbf{Optimal Transport geometry awareness}. Individual word-to-word distances may not capture the semantic similarity between two documents due to the lack of common words. OT leverages the underlying geometry (captured by the \textit{ground cost}) and lifts it to the Wasserstein space where the distance represents the cost of the optimal transportation of a whole document (distribution) to another.}
    \label{fig:gcost_lifting}
    \vspace{-1.5em}

\end{figure}

\textit{So, what does OT bring to the Machine Learning toolbox?} Many machine learning problems boil down to comparing probability distributions. 
Consider the problem of assessing the similarity of two documents containing \textit{``Obama speaks to the media in Illinois''} and \textit{``The president greets the press in Chicago''} \cite{KusnerFromDistances}. 
A similarity notion that relies on direct pairwise comparisons of the \textbf{individual elements} (i.e. words) would render the documents dissimilar, due to a lack of common words. 
However, our common sense tells us that the documents are actually similar.
Looking at the two sets of words \textbf{as a whole}, the first sentence is semantically very close to the second. 
The latter insight motivates the OT perspective \cite{WernerSpeedingEmbeddings, YurochkinHierarchicalRepresentation}.
OT accounts for the shapes of the distributions by computing a distance that measures how much effort is needed to align a distribution to another (Fig.~\ref{fig:gcost_lifting}) while allowing the distributions to have different supports. 
Among all possible ways to do such ``global'' alignment of distributions, OT picks the one with the least effort (\textit{optimal plan}) guided by a notion of ``local'' distance/cost between the basic features of the distributions' supports (e.g. words).
This local cost captures the geometry of the data and is usually called the \textit{ground cost}\footnote{Note that other competing distances such as total-variation or $\chi^2$-distances are oblivious to any notion of similarity on the ground space.}.
In summary, OT lifts the basic word-to-word ground cost to a global distance in the distributional space. The appealing properties of OT can be summarized in the following two main points.

\textbf{Underlying geometry awareness}. Allowing geometric information of the support to be taken into account (unlike fairly common distances/divergences) can result in meaningful and smooth divergences, even when the two distributions don't overlap. 
This powerful feature is leveraged in many applications including OT-based embedding \cite{Frogner2019LearningSpaces} and generative modeling \cite{Arjovsky2017WassersteinNetworks}. 
Geometry awareness also means that OT borrows properties of the underlying space \cite{Peyre2018ComputationalTransport}. 
For example, concepts such as interpolation and barycenter from Euclidean ground space carry on to the Wasserstein space. 
Thus, one can perform smooth interpolation between probability distributions \cite{BunneProximalDynamics} or even perform regression on histograms \cite{Bonneel2016WassersteinTransport}.

\textbf{Explicit Correspondence}: in addition to distances between distributions, OT typically provides correspondences through the optimal plan. 
Such correspondences can be used for applications such as object matching and registration \cite{shen2021accurate, puy2020flot}, and interpreting machine learning results (\cref{sec:domain_adaptation}).

\begin{figure}[!t]
    \centering
    \includegraphics[width= 1\linewidth]{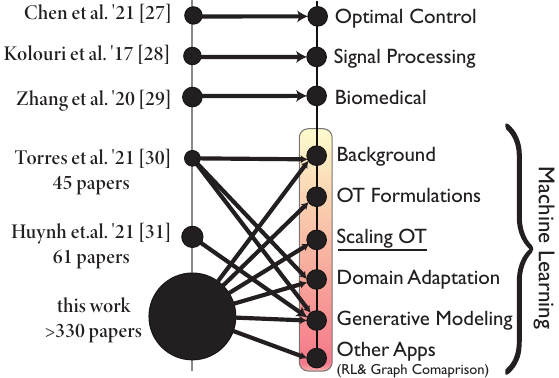}
    \vspace{-1.5em}
    \caption{ \textbf{Survey Position.} Among OT \rtypo{review papers} \cite{chen2021optimal, kolouri2017optimal, zhang2021review, torres2021survey, huynh2021optimal}, this work presents a very comprehensive and updated coverage of OT in machine learning. The circles on the left column represent the existing reviews with their size indicating the number of the reviewed papers. The connections between the left and right columns depict the topics covered in each considered review article. The underlined \underline{``Scaling OT''} is a key focus of this survey.
    }
    
    \label{fig:rel_surveys}
\vspace{-1.2em}
\end{figure}

 \begin{figure*}[!tb]
    \centering
    \includegraphics[width=0.95\linewidth]{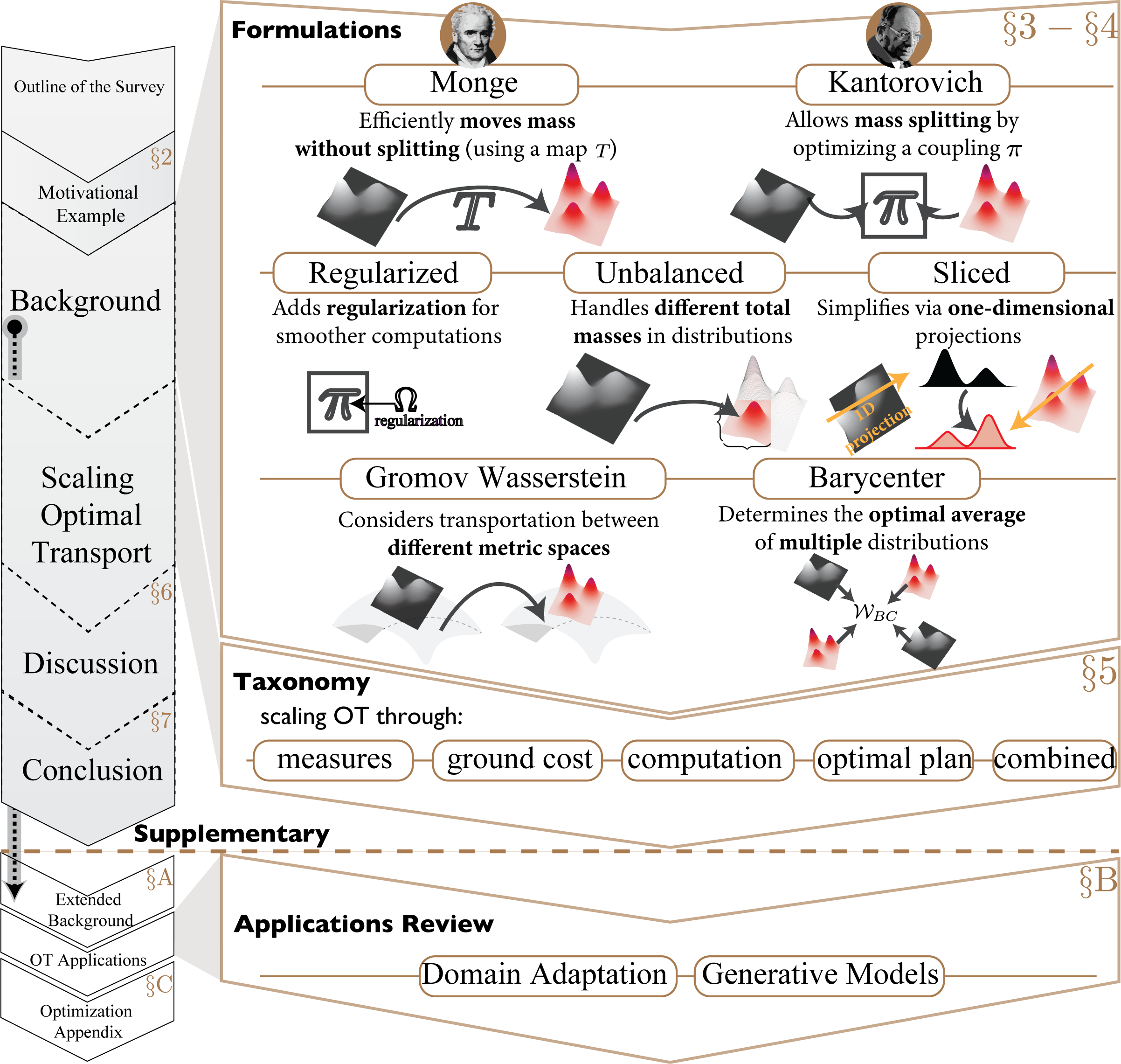}
    \vspace{-1.2em}
    \caption{\textbf{Outline of the Survey}. We start by \textbf{(\cref{sec:motivational_example} Motivational Example}) introducing the reader to OT through a motivational example that paves for \textbf{(\cref{sec:nota_background}, \cref{sec:common_ot_formulations} Background}) the discussion of OT formulations. Next, \textbf{(\cref{sec:scaling_ot} Scaling OT}) we present a taxonomy for scaling OT methods to big data regimes. Then, open issues and future research directions are discussed in \textbf{(\cref{sec:future_directions} Discussion}). In the supplementary material, we extend the background discussion \textbf{(\cref{sec:ot_background} Extended Background)}  and include a summary of OT applications in machine learning\textbf{(\cref{sec:ot_applictions} Applications}).  All the figures are best viewed in color. 
    }
    \label{fig:outline}
    \vspace{-1.5em}
\end{figure*}

\textit{What is the price tag?} The above-mentioned desirable characteristics come with longstanding computational and statistical challenges. 
First, OT-based distances are hard to compute, and explicit closed-form expressions for OT-based distances are rare beyond Gaussian and one-dimensional cases. 
In its original formulation, OT-based distance computation requires solving a linear program with a super-cubic complexity in the number of data points. 
Reducing this problematic complexity using various computational techniques and workarounds is an ongoing and interesting trend in the OT literature. 
We review many of these techniques  (\cref{sec:scaling_ot}) and categorize them in a taxonomy that reveals the connections among various techniques. 
Second, in high dimensional settings \marginpar{R3}\rtypo{ and without applying any scaling techniques (Sec. \cref{sec:scaling_ot})}, OT suffers from a severe ``curse of dimensionality''. The sample complexity\footnote{Sample complexity, informally, is the number of samples needed to estimate a function within a certain level of accuracy.} of estimating the Wasserstein distance
using a finite sample from the distributions grows exponentially in the dimension\cite{WeedSharpDistance, Lei2020ConvergenceSpaces}. 
To understand the gravity of the issue, contrast OT with another divergence, the Maximum Mean Discrepancy \cite{Gretton2022ATest}, whose sample complexity is dimension independent! 
Finally, there are limitations inherent to the vanilla OT formulation itself such as sensitivity to outliers \cite{Mukherjee2021Outlier-RobustTransport}, inability to incorporate context/structure in the optimal plan \cite{LimOrderTransport}, strong reliance on the chosen ground cost \cite{Dhouib2020ATransport}, assuming a common ground cost \cite{Zheng2022ADistance} and the inability to handle measures (generalized, un-normalized distributions) of non-equivalent masses (normalization constants) \cite{SatoFastTree}. 
Unsatisfied with these limitations, researchers have produced a diverse spectrum of OT formulations over the years with \rtypo{the potential} to meet a broader range of applications and requirements. 
Indeed, new formulations are still evolving as of this writing. 
For each formulation, we briefly discuss the characteristics, computational methods for solving it, and ML applications in which the formulation was used.

In the last decade, many significant developments in OT have taken place. 
New mathematical formulations, highly scalable algorithms (coping with datasets with millions of samples), interactions with deep learning, the rise of learned neural solvers, and empirical advancements in domain adaptation and generative modeling are a few developments to name. 
These developments call for a refreshed review of the field that sheds light on recent developments and identifies emerging trends and paradigms.
\marginpar{R1} \hypertarget{r11}{\rother{On one hand, there are excellent textbooks \cite{Peyre2018ComputationalTransport, Santambrogio15} that discuss the computational aspects of OT in great detail. However, their discussions of OT applications in ML and its scalability are dated.}} 
\rtypo{On the other hand, the existing review papers} (Fig.~\ref{fig:rel_surveys}) either have a narrow coverage of OT in machine learning or a specialized coverage irrelevant to machine learning. Our goal is to present a timely and comprehensive survey of the important topic of OT and OT applications in modern machine learning.

We summarize the contributions of this work as follows: \begin{enumerate}
    \item We present a comprehensive survey on the topic of OT with a focus on the applied side of the literature in a way that best targets the machine learning research audience.
    \item We compose a high-level view that organizes the techniques in literature for addressing the scalability issues of optimal transport (i.e. \rtypo{computational challenges}) in a single framework (\cref{sec:scaling_ot}). 
    \item We highlight literature gaps and discuss promising future research directions in OT and its applications.
\end{enumerate}

The remainder of this paper is organized as outlined in Fig.~\ref{fig:outline}. 
We took some efforts to make the paper self-contained. 
It is difficult to fit the OT-related optimization background into a few pages; we suggest that \cref{sec:app_optimization} can help the uninitiated readers before seeking other resources.

\section{A Motivational Example}\label{sec:motivational_example}

\begin{figure*}[!h]
    \centering
    \includegraphics[width= 0.95\linewidth]{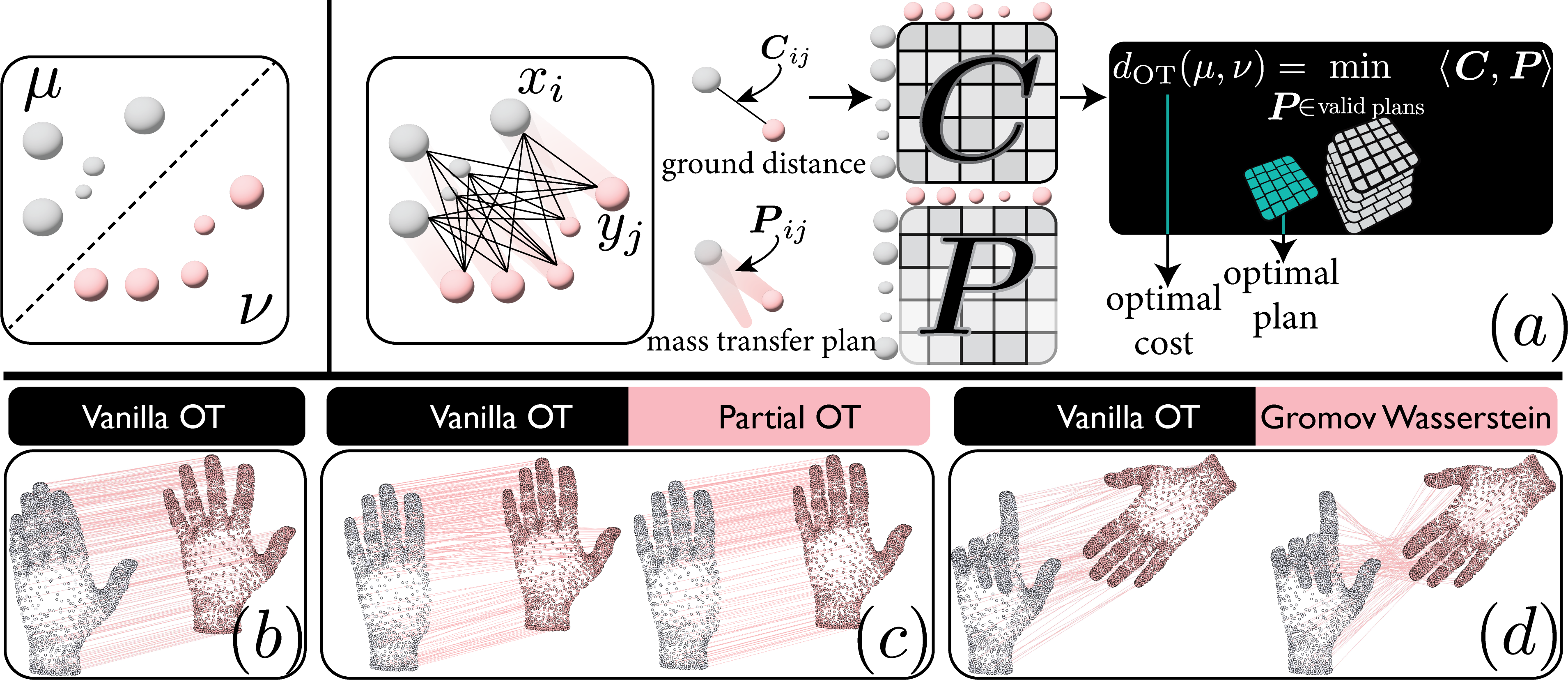}
    \vspace{-1em}
    \caption{ \textbf{Conceptual Depiction of Optimal Transport:}  \textbf{(a) }The optimal transport problem is stated as follows. Given two points cloud objects $\mu$ and $\nu$ and the knowledge of point-wise distances (i.e. the ground distance), find a legitimate way (i.e. valid plan) to redistribute the mass of $\mu$ into $\nu$ in the least costly way \rtypo{$\matrix{P}^*$}. The optimal transport cost is then  given by \rtypo{$d_\text{OT}(\mu,\nu) = \langle \matrix{C}, \matrix{P}^* \rangle$.} \textbf{(b)} Matching of two point cloud objects using optimal transport. The point correspondences estimated by the optimal plan are shown as straight pink lines. OT  literature is filled with formulations that extend this key concept to new situations and applications.  For example, formulations that allow \textbf{(c)} partial transportation or \textbf{(d)} transportation between incomparable spaces.
    }
    \label{fig:mot_example}
    \vspace{-1.5em}
\end{figure*}

In this section, we give a simple motivational example to introduce OT. 
Consider the problem of comparing point cloud objects. 
The objects are composed of 3D points that can be acquired by contemporary sensors such as Lidar, Radar, or depth sensors. 
Figure \ref{fig:mot_example}(a) shows examples of point cloud objects. 
To compare these objects, we seek a matching that redistributes the first object into the second in the least costly way (i.e. optimal transportation). 
Let's assume that there are $n$ and $m$ points in the objects $\mu$ and $\nu$; respectively. 
Furthermore, there is some mass (depicted as the volume of the point) that can be exchanged between the two objects during transportation. 
This mass can be, for example, the \textit{intensity} of the point \footnote{For example, in radar's point cloud the intensity is the strength of Radio Frequency (RF) reflection from the sensed point \cite{Khamis2020RFWash:Technique}.}. 
The term ``mass'' may be also used to refer to abstract or conceptual quantities. 
When unknown, we can assume that the masses are uniformly distributed. 
The mass of one point in the source can be exchanged with one or multiple points in the target.

Regardless of the notion of mass used, we assume that the points
$\{\vector{x}_i\}_{i=1}^{n}$ of the object $\mu$ have non-negative mass each that sum up to $1$ (i.e. normalized). 
We denote the vector of point masses in $\mu$ by \rtypo{$\vector{a} \in \mathbb{R}^n$}. 
The same assumption is made for the object $\nu$ whose points $\{\vector{y}_i\}_{i=1}^{m}$ are assumed to have a normalized mass vector \rtypo{$\vector{b}\in \mathbb{R}^m$}.

Suppose we wanted to transport the mass from the particles in $\mu$ to the particles in $\nu$ in a way that minimizes some total measure of cost, e.g. the sum of ``mass moved'' $\times$ ``the distance it moved''. 
More rigorously, we would like to decide on how much mass to move from point \rtypo{$\vector{x}_i$} to point \rtypo{$\vector{y}_j$}; \rtypo{$\matrix{P}_{ij}$}, considering the distance between \rtypo{$\vector{x}_i$ and $\vector{y}_j$; $\matrix{C}_{ij}$}. 
The first \rtypo{$\matrix{P}_{ij}$} is sometimes known as a transfer \textit{plan} which is to be identified while the latter \rtypo{$\matrix{C}_{ij}$} is known as \textit{ground distance} (e.g. the squared Euclidean distance $\Vert \vector{x}_i - \vector{y}_j\Vert^2$) which is given. 
Let \rtypo{$\matrix{P} \in \mathbb{R}^{n \times m}$} and \rtypo{$\matrix{C} \in \mathbb{R}^{n \times m}$} denote these quantities. 
The goal of OT is to find an optimal plan \rtypo{$\matrix{P}^*$} that dictates the mass distribution plan from the source points to the destination points in a way that minimizes the total transport cost. 
The set of all valid plans (formally defined in \hyperlink{couplings}{\textbf{Couplings}}) is described as follows:
\marginpar{R1} 
\rtypo{ 
\begin{align*}
\text{valid plans} = \{ \matrix{P} \in \mathbb{R}_{+}^{n \times m} | \sum_j \matrix{P}_{ij} = a_i ,  \sum_i \matrix{P}_{ij} = b_j \}.
\end{align*}}
The constraints ensure that \rtypo{$\matrix{P}$} preserves the mass so no mass is created or destroyed during transportation. 
In light of this, the optimal plan \rtypo{$\matrix{P}^*$} would be one that minimizes
\rtypo{\begin{align*}
\normalfont
d_{\text{OT}}(\mu, \nu) = 
\min\limits_{\matrix{P} \in \text{valid plans}}
\langle \matrix{C}, \matrix{P} \rangle,
\end{align*}}
where $\langle \cdot,\cdot \rangle$ is the Frobenius inner product. 
Applying the formulation above on point cloud objects (Fig \ref{fig:mot_example}(b) hand gestures from the CAPOD dataset \cite{papadakis2014canonically}), we can learn about the global proximity of the two shapes through the distance $d_{\text{OT}}$. 
Moreover, we learn the point correspondences conveyed by the optimal plan \rtypo{$\matrix{P}^*$} (visualized as straight lines).  

Posed this way, the OT problem (sometimes known as the Kantorovich formulation) possesses a useful set of characteristics as well as limitations, which motivate other formulations of the problem. 
We discuss many of them in \cref{sec:common_ot_formulations}. 
\rtypo{In some cases, one may wish to relax the mass preservation constraint.} 
For example, as shown in Fig. \ref{fig:mot_example}(c), \marginpar{R1}\rtypo{the pink hand object can only be \textbf{partially} matched to the incomplete (grey) hand copy as the thumb is missing. The vanilla OT forces the transportation of all points, resulting in inaccurate matching and (unrealistically) increased cost. Note the incorrect matching of the thumb's points.} A partial OT formulation (\cref{sec:unbalanced_OT}), however, is a better alternative as it has the flexibility \rtypo{of leaving some points out (the thumb) from the transportation}. Another limitation is that transportation that relies only on point correspondences conveyed by the ground distance (without considering the inter-points relations) can be misleading. In Fig. \ref{fig:mot_example}(d), we see a simple rotation transformation of the target object causes a degradation in the alignment by vanilla OT. This can be clearly seen in the incorrect correspondences of the index and the thumb fingers. This is not the case with a relational formulation (Gromov Wasserstein \cref{sec:gromov_wasserstein}) that handles this issue well at the expense of increased computation. This formulation can also compare points defined on
distinct (incomparable) spaces which can be useful for comparing graphs or measurements from different modalities.

\section{Background}
\label{sec:nota_background}
We begin with notation and background of the key OT formulations. We state the Monge and Kantorovich formulations below, and refer to an extended discussion of Kantorovich formulation in \cref{sec:ot_background}.

\subsection{Notation}

\textbf{Probability measures.} Let $\mathbb{X}$ and $\mathbb{Y}$ be two non-empty sets. 
Denote by $\mathcal{P}(\mathbb{X})$ the set of probability measures supported on $\mathbb{X}$. 
Let $\mu \in \mathcal{P}(\mathbb{X})$ and $\nu \in \mathcal{P}(\mathbb{Y})$. 
A function $\phi:\mathbb{X}\to\mathbb{R}$ is said to be integrable with respect to $\mu$ if it has finite expectation under $\mu$, i.e. \rtypo{${\mathbb{E}_{\vector{\rv{X}} \sim \mu} \big| \phi(\vector{\rv{x}}) \big| = \int_\mathbb{X} \vert \phi(\vector{x}) \vert \, d\mu(\vector{x}) < \infty}$}.
The space of functions integrable with respect to $\mu$ is denoted $${L^1(\mu) = \{ \phi:\mathbb{X} \to \mathbb{R} \mid \mathbb{E}_{\vector{\rv{x}} \sim \mu} \big|\phi(\vector{\rv{x}})\big| < \infty \}}.$$

The mathematical objects that we discuss can take on a discrete, continuous or mixed nature.
Averages can be \marginpar{R1}\rtypo{computed} using one or more of sum, integral or expectation notation.
While in principle these objects can be dealt with using one unified notation as an integral with respect to a measure, for ease of reading we sometimes prefer to explicitly write out the same equation using multiple notations.
To simplify notation, we will assume that every probability measure has an associated probability density function or probability mass function.

\textbf{Nonnegative \rtypo{unnormalized} measures.}
Most of the time, we will be interested in distances defined over the space of probability measures. 
Unless otherwise stated, all measures are probability measures.
In some cases (for example, \S~\ref{sec:unbalanced_OT}), these distances \rtypo{generalize} to spaces containing nonnegative and finite measures that are not necessarily probability measures.
Where we deal with such nonnegative \rtypo{unnormalized} measures, we will mention this close to where they are introduced.
In such cases, the expectation notation \rtypo{$\mathbb{E}_{\vector{\rv{x}} \sim \mu} \cdot$} is inappropriate and we resort to integrals \rtypo{$\int_{\mathbb{X}} \cdot d \mu(\vector{x}) $} or sums.
We denote by $\mathcal{M}_+(\mathbb{X})$ the space of finite nonnegative \rtypo{unnormalized} measures supported over $\mathbb{X}$.

\hypertarget{couplings}{\textbf{Coupling.}} A useful notion appearing in OT (for example, in the Kantorovich formulation~\eqref{eq:kantorovich}) is that of coupling.
Given two random vectors \rtypo{$\vector{\rv{x}}$ and $\vector{\rv{y}}$}, we may construct a random vector \rtypo{$(\vector{\rv{x}},\vector{\rv{y}})$} having marginal distributions that are the same as the distributions of \rtypo{$\vector{\rv{x}}$ and $\vector{\rv{y}}$}, and a non-unique joint distribution consistent with those marginals. 
The space $\Pi$ of all such joint distributions constitutes a useful space over which to search.
Define the set $\Pi(\mu, \nu)$ of joint probability measures with marginals $\mu$ and $\nu$.
In other words, $\Pi(\mu, \nu)$ is the set of plausible plans
\rtypo{\begin{align*}
    &\Pi(\mu, \nu) = \Big\{ \pi \in \mathcal{P}(\mathbb{X} \times \mathbb{Y}) \mid \forall A \subseteq \mathbb{X}, B \subseteq \mathbb{Y},  \\
    &\int_{A \times \mathbb{Y}} d \pi (\vector{x}, \vector{y}) = \int_A d\mu(\vector{x}),  \int_{\mathbb{X} \times B} d \pi (\vector{x}, \vector{y}) = \int_B d\nu(\vector{y}) \Big\}.
\end{align*}}

\textbf{Matrix representations for discrete and finite support.} 
It is useful to instantiate measures and couplings in terms of vectors and matrices in the case where the random variables in question are discrete and have finite support.

Let $\widetilde{a}$ and $\widetilde{b}$ be the probability mass functions corresponding \marginpar{R1}\rtypo{to} $\mu$ and $\nu$. Let the support of $\widetilde{a}$ and $\widetilde{b}$ be finite, that is $d_{\mathbb{X}} = \big| \mathbb{X}  \big| < \infty$ and $d_{\mathbb{Y}} = \big| \mathbb{Y}  \big| < \infty$. Define the probability simplex \rtypo{$\Sigma_d = \{ \vector{x} \in [0,1]^d \mid \vector{x}^\top \mathbf{1} = 1 \}.$} Let \rtypo{$\vector{a} \in \Sigma_{d_{\mathbb{X}}}$} and \rtypo{$\vector{b} \in \Sigma_{d_{\mathbb{Y}}}$} be vector representations of $\widetilde{a}, \widetilde{b}$.
That is, for some assignment \rtypo{$\{ \vector{x}_i \mid 1 \leq i \leq d_{\mathbb{X}}\} = \mathbb{X}$ and $\{ \vector{y}_i \mid 1 \leq i \leq d_{\mathbb{Y}}\} = \mathbb{Y}$, set $a_i = \widetilde{a}(\vector{x}_i)$} and \rtypo{$b_i = \widetilde{b}(\vector{y}_i)$}. 
The set $\Pi(\mu, \nu)$ corresponds with a set of finite-dimensional matrices $U(\vector{a}, \vector{b})$ representing joint probability mass functions consistent with marginals $\mu$ and $\nu$. Define
\rtypo{\begin{align*}
    U(\vector{a},\vector{b}) = \{ \matrix{P} \in [0,1]^{d_{\mathbb{X}} \times d_{\mathbb{Y}}} \mid \matrix{P} \mathbf{1} = \vector{a}, \matrix{P}^\top \mathbf{1} = \vector{b}\},
\end{align*}}
and note that for any \rtypo{$\matrix{P} \in U(\vector{a},\vector{b})$}, we have that the $ij$th entry of \rtypo{$\matrix{P}$} satisfies \rtypo{$\matrix{P}_{ij} = \text{Pr}(\vector{\rv{x}} = \vector{x}_i, \vector{\rv{y}} = \vector{y}_j)$} under some joint probability mass function for random variables \rtypo{$(\vector{\rv{x}}, \vector{\rv{y}})$} with marginal probability mass functions $\widetilde{a}$ and $\widetilde{b}$.

\textbf{Empirical measures}
When dealing with real-world data and problems, we usually don't have access to explicit forms of distributions $\mu$ and $\nu$.
Instead, we have samples $\{\vector{x}_i\}_{i=1}^{m} \sim \mu$ and $\{\vector{y}_j\}_{j=1}^{n} \sim \nu$.
In these settings, it is useful to define and work with the empirical measures \rtypo{$\mu_n = 1/n \sum_i \delta_{\vector{x}_i}$} and \rtypo{$\nu_n = 1/n \sum_j \delta_{\vector{y}_j}$.}
Here \rtypo{$\delta_{\vector{z}}$} is a unit mass centered at \rtypo{${\vector{z}}$}.

\subsection{Monge Formulation (Optimal Map)}
 Given probability measures $\mu$ and $\nu$, the OT map \rtypo{${\vector{t}^\ast}:\mathbb{X} \to \mathbb{Y}$}is defined to be one that maps random variables \rtypo{$\vector{\rv{x}}$ following $\mu$ to random variables $\vector{\rv{y}}$ following $\nu$ with the minimum expected cost between $\vector{\rv{x}}$ and ${\vector{t}^\ast}(\vector{\rv{x}})=\vector{\rv{Y}}$}. The Monge formulation \cite[Chapter 1]{Santambrogio15} is :
\rtypo{
\begin{align*}
    \mathfrak{M}_c(\mu,\nu) & = \inf\limits_{\vector{t} \in \mathcal{T}_{\mu\nu}} \int_{\mathbb{X}} c\big(\vector{x},\vector{t}(\vector{x})\big) d\mu(\vector{x}) \\  
    & = \inf\limits_{\vector{t} \in \mathcal{T}_{\mu\nu}}  \mathbb{E}_{\vector{\rv{x}} \sim \mu} \ [ c(\vector{\rv{x}},\vector{t}(\vector{\rv{x}}) \big) ]. \numberthis \label{eq:monge}
\end{align*}
}

Here \rtypo{$\mathcal{T}_{\mu\nu} =\{\vector{t}:\mathbb{X} \to \mathbb{Y} \mid \vector{t}_\#(\mu) = \nu\} $} constitutes the set of all possible mappings.  \rtypo{$\vector{t}_\#(\mu)$} is the pushforward measure of $\mu$ under \rtypo{$\vector{t}$}. In the context of probability measures this means that \rtypo{$\vector{t}_\#(\mu)$} is the probability measure of the random variable \rtypo{$\vector{t}(\vector{\rv{x}})$}. \rtypo{$c$} is the ground cost responsible for transporting one unit of mass from \rtypo{$\vector{\rv{x}}$ to $\vector{\rv{y}}$}. The choice of $c$ is typically application-dependent and influenced by domain knowledge and awareness of data. 

The Monge formulation~\eqref{eq:monge} may be ill-posed; there may not exist any mapping \rtypo{$\vector{T}$} such that \rtypo{$\vector{T}_\#(\mu) = \nu$.} For example, if $\mu$ places probability mass over $2$ states, it can never be mapped to a measure that places probability mass over $3$ states using a deterministic function. The Kantorovich formulation, introduced next, does not suffer from this issue.

\subsection{Kantorovich Formulation (Optimal Plan)}
\label{sec:back_kantorovich}

Given $\mu$ and $\nu$ and some cost function $c$, the Kantorovich OT problem \cite[Chapter 1]{Santambrogio15} (the general formulation of the discrete example in \cref{sec:motivational_example}) is defined as:
\rtypo{\begin{align*}
    \mathcal{K}_c(\mu, \nu) = 
 & \inf\limits_{\pi \in \Pi(\mu, \nu)} \int_{\mathbb{X} \times \mathbb{Y}}  c(\vector{x},\vector{y}) d\pi(\vector{x},\vector{y}) \\
 =& \inf\limits_{\pi \in \Pi(\mu, \nu)} \mathbb{E}_{(\vector{\rv{X}},\vector{\rv{Y}}) \sim \pi} \ [c(\vector{\rv{X}},\vector{\rv{Y}})].
 \numberthis 
\label{eq:kantorovich}
\end{align*}}
Any $\pi^\ast$ that attains this infimum is called an OT plan.
In the case that $\vector{\rv{x}}$ and $\vector{\rv{y}}$ are discrete random variables with probability mass vectors $\vector{a}$ and $\vector{b}$, we write
\rtypo{\begin{align}
    \mathcal{K}_c(\mu, \nu)  &=  \min\limits_{\matrix{P} \in U(\vector{a},\vector{b})} \langle \matrix{C}, \matrix{P} \rangle
\label{eq:kantorovich_discrete}
\end{align}}
where $\matrix{C} = [c(x_i, y_j)]_{ij}$ is the matrix containing the pairwise costs between \rtypo{$\vector{\rv{x}}$ and $\vector{\rv{y}}$}.
Interestingly, \eqref{eq:kantorovich}, known as the \textit{primal} formulation, can be written in a number of equivalent formulations (such as the \textit{dual}  formulation). \marginpar{R2} \rref{Note that interesting theoretical and practical connections were made between Monage and Kanorovich formulations. Brenier’s theorem \cite[Theorem 2.1]{Peyre2018ComputationalTransport} establishes the equivalence of Kantorovich and Monge's problem in the Euclidean space where the optimal map is the gradient of convex functions. Also, some works \cite{perrot2016mapping} augment Kanorovich's formulation with a Monge map variable and jointly learn the optimal coupling and an approximation of the transport map.}
\marginpar{R1}\rtypo{ Check \cref{sec:ot_background} for intuitive and comprehensive exposition on alternative Kantorovich formulations and a discussion of Hierarchical Optimal Transport formulations.}

\textbf{Wasserstein Distance:} It may be shown that if \rtypo{$c(\vector{x},\vector{y}) = d(\vector{x},\vector{y})^p$} for some distance $d$ on $\mathbb{S} = \mathbb{X} \cup \mathbb{Y}$, i.e. $d:\mathbb{S} \times \mathbb{S} \to \mathbb{R}$ and some $p \geq 1$, then the $p$-Wasserstein distance \cite[Definition 6.1]{Berlin2009OptimalNew}  :
\begin{align}
    W_p(\mu, \nu) &= \Big( \mathcal{K}_{d^p}(\mu, \nu)  \Big)^{1/p}
\label{eq:wasserstein_distance}
\end{align}
is a distance metric over the space of probability measures with finite moments up to order $p$ (Theorem 7.3 (i)~\cite{VillaniTopicsTransportation}). In the case where $0 \leq p < 1$, $W_p^p(\mu, \nu)$ is a distance metric over the space of probability measures with finite moments up to order $p$ (Theorem 7.3 (ii)~\cite{VillaniTopicsTransportation}).

\section{Common OT Formulations}
\label{sec:common_ot_formulations}

Here we discuss OT formulations with a focus on the balance between proper coverage and concise treatment.

\subsection{Regularized OT}
\label{sec:regularized_ot}

\textit{Formulation and Characteristics}: Some of the computational and statistical limitations of the original OT formulation ( \eqref{eq:kantorovich}) can be mitigated by augmenting the original objective with a regularization term. Regularized OT was brought to the forefront of applied machine learning after Cuturi published his seminal work on the topic ten years ago \cite{CuturiSinkhornTransport}. 
The work demonstrated the computational superiority of the entropic regularized formulation and its compatibility with modern ML pipelines (i.e., parallelism and differentiability). 
Yet, the formulation can be traced back to the works of Erwin Schr\"odinger in 1930s \cite{ChenStochasticBridge}\footnote{Recently, OT and Schr\"odinger bridge have provided insights into diffusion-based generative models~\cite{BortoliDiffusionModeling,Maoutsa2022TransportBridges}. }.  
The regularized OT problem is defined as:
\rtypo{\begin{equation}
\normalfont
{\mathcal{OT}}_{\Omega, \lambda}(\mu, \nu) = 
\inf\limits_{\pi \in \Pi(\mu, \nu)}
\int_{\mathbb{X} \times \mathbb{Y}} c(\vector{x}, \vector{y}) d\pi(\vector{x}, \vector{y}) 
+ \lambda \ \Omega(\pi),
\label{eq:cont_regularized_primal}
\end{equation}}
where $\Omega$ is a regularization operator and $\lambda$ is a positive regularization coefficient. The regularized discrete OT problem is similarly given by:
\rtypo{\begin{equation}
\normalfont
{\mathcal{OT}}_{\Omega, \lambda}(\mu, \nu) = 
\inf\limits_{\matrix{P} \in U(\vector{a},\vector{b})}
\langle \matrix{C}, \matrix{P} \rangle 
+ \lambda \ \Omega(\matrix{P}).
\label{eq:regularized_primal}
\end{equation}}

One choice for $\Omega$ is the KL divergence between the joint distribution \rtypo{$\matrix{P}$} (or measure $\pi$) and the product measure $\vector{m}_1 \times \vector{m}_2$ for a set of reference measures $\vector{m}_1$ and $\vector{m}_2$. That is, $\Omega(\matrix{P}) = \text{KL}(\matrix{P} \mid\mid \vector{m}_1 \times \vector{m}_2)$.
When additionally the reference measures $\vector{m}_1$ and $\vector{m}_2$ are uniform, the \rtypo{regularization} term (up to a constant) is called entropic \rtypo{regularization}:
\rtypo{\begin{align*}
    \Omega(\matrix{P}) = -H(\matrix{P}) = \sum_{ij} \matrix{P}_{ij} (\log \matrix{P}_{ij} - 1),
\end{align*}}
where $H$ is the Boltzmann-Shannon entropy function.
Adding the negative of the Boltzmann-Shannon entropy function as a regularization gives us the famous entropic-regularized OT:
\rtypo{\begin{equation}
\normalfont
\mathcal{OT}_{\lambda}(\mu, \nu) = 
\inf\limits_{\matrix{P} \in U(\vector{a},\vector{b})}
\langle \matrix{C}, \matrix{P} \rangle 
- \lambda H(\matrix{P}),
\label{eq:ent_regularized_primal}
\end{equation}}
which is a version of the original problem (\eqref{eq:kantorovich_discrete}) with an additional strictly convex regularization term (negative of the Boltzmann-Shannon entropy). Intuitively, this regularization biases the optimal plan towards the uniform distribution. Interestingly, entropic-regularized OT (whose optimal objective value is also known as the Sinkhorn ``distance'')
can be rewritten as minimization of a Kullback-Leibler (KL) divergence,
\rtypo{\begin{equation}
\mathcal{OT}_{\lambda}(\mu, \nu) = \lambda \inf\limits_{\matrix{P} \in U(\vector{a},\vector{b})} \text{KL}(\matrix{P} \mid\mid \matrix{K}),
\label{eq:ent_regularized_bregman}
\end{equation}}
where the matrix \rtypo{$\matrix{K}$} \marginpar{R1}\rtypo{(also known as Gibbs kernel)} is the element-wise negative exponential of the scaled ground cost \rtypo{$\matrix{K}_{ij} = \exp({-\matrix{C}_{ij}/\lambda})$. The KL divergence is known to be a strictly convex function of $\matrix{P}$ for a fixed $\matrix{K}$, which means that there exists a unique $\matrix{P}$ that minimizes the above expression and that is differentiable with respect to $\matrix{K}$. Note that the Sinkhorn distance, despite not being a proper ``distance'', is symmetric in $\mu$ and $\nu$, i.e., $\mathcal{OT}_{\lambda}(\mu, \nu) = \mathcal{OT}_{\lambda}(\nu, \mu)$, if the cost function $c(\vector{x}, \vector{y})$ is symmetric in $\vector{x}$ and $\vector{y}$.}

\textit{Computational Motivation:} While the regularization term can be motivated by application requirements through the inclusion of an additional prior \cite{Ferradans2014RegularizedTransport}, one can safely argue that the computational advantage of entropic regularization is a key appealing benefit. Specifically, the formulation in  \eqref{eq:ent_regularized_primal}, which was popularized in the GPU era by Cuturi \cite{cuturi13}, can be solved efficiently using an iterative matrix scaling algorithm known as the Sinkhorn-Knopp\cite{Knight2008TheApplications} algorithm, often shortened to the Sinkhorn algorithm (Alg.~\ref{alg:sinkhorn}). \rtypo{This algorithm boils down the process of computing the optimal plan to a series of matrix-vector multiplications. The kernel matrix capturing the measures $\mu$ and $\nu$ dissimilarity is denoted as $\matrix{K}_{ij}=e^{-\frac{\matrix{C}_{ij}}{\lambda}}$ where $\matrix{C}$ is the ground cost matrix and $\lambda$ is the regularization coefficient in (\eqref{eq:regularized_primal}). The algorithm seeks the unique optimal solution that must satisfy the following:}
\rtypo{\begin{align}
    \matrix{P}_\lambda = \text{diag}(\vector{u})\matrix{K}\text{diag}(\vector{v})
    \label{eq:sk_theorem} \tag{\text{Sinkhorn theorem}}
\end{align}}
\rtypo{\begin{equation}
    \matrix{P}_\lambda \mathds{1}_m= \vector{a} \ , \matrix{P}_\lambda^{T} \mathds{1}_n= \vector{b}
    \label{eq:opt_1} \tag{\text{optimality condition}}
\end{equation}}
\rtypo{Finding the vectors $u$ and $v$ that satisfy the requirements above is achieved simply by performing repeated alternate scaling (line 3 in Alg. ~\ref{alg:sinkhorn}) until convergence. The number of iterations needed to converge to a specific tolerance $\delta$ is controlled by the scale of elements in $\matrix{C}$ relative to $\lambda$ \cite{FranklinOnMatrices}. The iterations are merely applying kernel matrix $\matrix{K}$ (or its transpose) to vectors $u$ and $v$. } The Sinkhorn algorithm has a worst-case quadratic complexity in the number of points per iteration in its base form (with many enhancements possible, e.g. \cite{AltschulerMassivelyMethod}) and linear convergence rate. This is much more efficient and scalable than the linear programming methods employed by solvers for the un-regularized formulation.

\begin{algorithm2e}
\SetAlgoLined
\textbf{Inputs:} $\matrix{K},\vector{a},\vector{b},\delta, \vector{u}$\\
\Repeat{$r<\delta$}
{
    $\vector{v}\gets \vector{b}/\matrix{K}^T \vector{u},\;\vector{u}\gets \vector{a}/\matrix{K}\vector{v}$ \\
    $r\gets\|\vector{u}\odot \matrix{K}\vector{v} - \vector{a}\|_1 +\|\vector{v}\odot \matrix{K}^T \vector{u} - \vector{b}\|_1$
}
\KwResult{$\vector{u},\vector{v}$}
\caption{$\text{Sinkhorn}(\matrix{K},\vector{a},\vector{b},\delta)$ \label{alg:sinkhorn}}
\end{algorithm2e}

\textit{Sinkhorn Divergence:} Despite the computational appeal of the the \rtypo{Sinkhorn} ``distance'', two limitations arise in  \eqref{eq:ent_regularized_primal}. First, the regularized OT optimal value is not a distance.
\marginpar{R1}\rtypo{ Second, the formulation induces a bias in the minimizer known as the entropic bias \cite{Feydy2019InterpolatingDivergences}}.
These issues are handled by the Sinkhorn divergence \cite{Feydy2019InterpolatingDivergences, Sejourne2021SinkhornTransport}; another variant of regularized OT that builds on Sinkhorn "distance" and is defined as follows:
\begin{equation}
\overline{\mathcal{OT}}_{\lambda}(\mu, \nu) =  \mathcal{OT}_{\lambda}(\mu, \nu) - \frac{1}{2} \Big(\mathcal{OT}_{\lambda}(\mu, \mu) + \mathcal{OT}_{\lambda}(\nu, \nu)\Big).
\label{eq:sinkhorn_divergence}
\end{equation}
Sinkhorn divergence is smooth and differentiable in the inputs and enjoys a better sample complexity \cite{Genevay2019SampleDivergences} compared to regular OT. Yet, it can be seen that it requires three evaluations of $\mathcal{OT}_{\lambda}$  rather than one, although the symmetric Sinkhorn algorithm can be used to compute $\mathcal{OT}_{\lambda}(\mu, \mu)$ and $\mathcal{OT}_{\lambda}(\nu, \nu)$ more quickly than the regular Sinkhorn algorithm. 
\marginpar{R3}\rref{Altschuler \etal \cite{altschuler2017near} provides analysis of the computational complexity of Sinkhorn and the faster variant called Greenkhorn.} Note that we discuss scalable variants of Sinkhorn later in \cref{sec:scalingot_cost_approx} and \cref{sec:scaling_stochastic}.

\textit{Applications:} Entropic-regularized OT and the Sinkhorn ``distance'' have been extensively used in many machine learning applications. Some examples include \rtypo{self-labelling} \cite{AsanoSelf-labellingLearning}, adversarial examples \cite{Wong2019WassersteinIterations},  permutations learning \cite{MenaLEARNINGNETWORKS}\cite{Cruz2019VisualLearning } in deep models, initializing graph correspondence in graph matching learning \cite{FeyDEEPCONSENSUS}, differentiable sorting and ranking \cite{Cuturi2019DifferentiableTransport}\cite{Adams2011RankingPropagation} and enforcing doubly stochastic attention in transformers \cite{Zuo2020Sinkformers:Attention}. \rother{Sinkhorn was also employed in graph comparison pipelines (check \cref{sec:app_other_applications})}. Since the kind of regularization used in OT can be adapted depending on the application requirements, some works investigate alternatives to the entropy term in \eqref{eq:ent_regularized_primal}. Graph-inspired regularization was used in domain adaptation \cite{Courty2017OptimalAdaptation} and color transfer application\cite{Ferradans2014RegularizedTransport}. This was extended by Laplacian OT \cite{OptimalInria} and \cite{Courty2017OptimalAdaptation}. Other regularization terms that reflect various biases/prior, such as \rref{squared Euclidean distance regularization \cite{Blondel2018SmoothTransport}}, Lasso and \marginpar{R1} \rref{group Lasso \cite{Courty2017OptimalAdaptation}},  also exist in the literature. Temporal regularization \cite{Su2017Order-preservingMatching} was used in human pose alignment \cite{Su2017Order-preservingMatching} and \cite{KumarUnsupervisedClustering} weakly supervised action segmentation.

\textit{Limitations}: A known limitation of entropic regularized OT is the difficulty of setting the regularization strength $\lambda$. Big computational gains are tied to a high level of regularization (i.e., bigger $\lambda$), \rtypo{which can cause overspreading }(i.e., blurry optimal plan) \cite{Blondel2018SmoothTransport}. For example, in the point matching example (\cref{sec:motivational_example}), high regularization translates to dense matching where most points are connected to each other. This can hurt the correspondence interpretation. On the other hand, \rtypo{concentrated sparse plans} can theoretically be obtained for very small $\lambda$ \rtypo{, but in practice} this causes numerical issues \cite{Computing2019StabilizedProblems,Courty2017OptimalAdaptation}.

\subsection{Unbalanced and Partial OT}
\label{sec:unbalanced_OT}

\textit{Motivation:} The mass preservation constraint in the Kantorovich formulation \eqref{eq:kantorovich}  requires that the total mass between the two probability distributions be the same. Unbalanced OT (UOT) refers to formulations that relax this constraint to allow \rtypo{the transporting of} arbitrary (unnormalized) measures or partial masses. This is useful in applications like multi-object tracking \cite{Le2021UnbalancedApplications}, and crowd counting \cite{WanALocalization, Ma2021LearningTransport}.

\textit{Formulation and Characteristics}: One way of extending OT  to arbitrary positive measures is by adopting \textit{marginals relaxation}, in which the hard marginal constraints \eqref{eq:kantorovich} are removed, and the objective is augmented with soft regularization that penalizes the mass variation. Typically, this is done using Csiszar divergence $D_\phi$ \cite{ChizatUnbalancedFormulations}: 
\rtypo{\begin{align}
\normalfont
\begin{split}\label{eq:unbalanced_ot}
\mathcal{UOT}_{\phi}(\mu, \nu) = & 
\inf\limits_{\pi \in \mathcal{M}_{+}(\mathbb{X} \times \mathbb{Y})}
\int_{\mathbb{X} \times \mathbb{Y}} c(\vector{x}, \vector{y}) d\pi(\vector{x}, \vector{y}) \\ 
 & + \lambda_1 \ D_{\phi}(\pi_1 | \mu) + \lambda_2 \ D_{\phi}(\pi_2 | \nu). 
 \end{split}
\end{align}}
where $\mathcal{M}_{+}(\mathbb{X} \times \mathbb{Y})$ is the space of finite non-negative measures over $\mathbb{X} \times \mathbb{Y}$ and $D_\phi$ is a divergence induced by $\phi$ that quantifies mass variation between the marginals of the plan's marginals $( \pi_1, \pi_2)$ and the measures $(\mu, \nu)$.  The penalization strength is controlled by $(\lambda_1,\lambda_2)$. An obvious consequence of replacing constraints with penalties is that $\pi$'s marginals no longer need to be equal to $(\mu, \nu)$. Particular instances of $D_{\phi}$ include KL divergence \cite{Liero2018OptimalMeasures} and Total Variation  \cite{Sejourne2022UnbalancedNumerics}. Recently, \cite{Manupriya2020AMetrics} proposed a regularization based on Maximum Mean Discrepancy (MMD) and \rtypo{proved} that the modified formulation has desirable properties such as dimension-free sample complexity.  

Another way of extending OT to unbalanced measures, usually leveraged in discrete cases, is \textit{partial assignment}. The formulation is called Partial OT (POT). When transporting from $n$ to $m$ points, one can adapt the formulation \eqref{eq:kantorovich} by \rtypo{augmenting the plan $\matrix{P} \in \mathbb{R}^{n \times m}$  to be $\widetilde{\matrix{P}} \in \mathbb{R}^{(n+1) \times (m+1)}$ 
and similarly for the ground distance matrix $\matrix{C}$  to be $\widetilde{\matrix{C}} \in \mathbb{R}^{(n+1) \times (m+1)}$.}  The additional row and column are called \textit{dustbin} \cite{SarlinSuperGlue:Networks, Dang2020LearningRegistration} or \textit{dummy points}\cite{Chapel2020PartialLearning} and should absorb the mass from unmatched points. This, in effect, turns the formulation into a balanced one, and thus, the traditional computation of \eqref{eq:kantorovich} can be recycled.

\textit{Computation:} Entropic regularization may be combined with UOT \cite{Pham2020OnAlgorithm} or POT \cite{Dang2020LearningRegistration}. This allows one to invoke the Sinkhorn algorithm and speed up the computation.

\textit{Applications:} The balanced OT formulation in \eqref{eq:kantorovich} can be highly non-robust when there are outliers. Since it endeavors to transport all the mass from $\mu$ to $\nu$, a single contaminated data point can increase the OT cost arbitrarily. UOT, on the other hand, can be less sensitive to this issue by assigning small masses to outliers. This makes UOT more suitable for machine learning applications that deal with corrupted and noisy data. OT variants that detect and don't transport outliers build on the unbalanced formulation. For example, ROBust OT (ROBOT) \cite{Mukherjee2021Outlier-RobustTransport} performs outlier detection using a formulation that builds on UOT. 

\marginpar{R3}\rother{\textit{Limitations:} Unbalanced OT performance can be sensitive to the choice of the mass variation parameters ($\lambda_1$ and $\lambda_2$). Additionally, when regularization is integrated, tuning the regularization coefficient is challenging and application-dependent \cite{sejourne2022unbalanced}.  }

\subsection{Sliced OT} 
\label{sec:sliced_wasserstein}

\textit{Motivation}: One approach to speed up the OT computation relies on the idea of low-dimensional projections by aggregating OT distances computed on 1D projections of the data. \marginpar{R3}\rref{What motivates this approach is the fact that solving OT problems in the univariate/1D case is cheap \cite{KolouriGeneralizedDistances}}. Specifically, we obtain a representation for the p-Wasserstein distance between one-dimensional $\mu$ and $\nu$ with quantile functions of $F_\mu^{-1}$ and $F_\nu^{-1}$, respectively, as follows: 
\begin{equation} 
    \mathcal{W}_p(\mu, \nu) = \Big( \int_{0}^1 c^p \Big( F_\mu^{-1}(t) , F_\nu^{-1}(t) \Big) dt \Big)^{\frac{1}{p}}
\label{eq:wasserstein_1d}
\end{equation}
where $c^p(.,.)$ is the ground cost raised to the $p^{th}$ power. When $\mu$ and $\nu$ are empirical distributions with $n$ and $m$ samples, \eqref{eq:wasserstein_1d} can be computed very efficiently through simple sorting with a complexity of \rtypo{$\mathcal{O}(n \log (n) + m \log (m))$}.

\textit{Formulation and Characteristics}: Sliced Wasserstein (SW) \cite{Bonneel2015SlicedMeasures} is motivated by explicit utilization of the above computational efficiency. SW is computed as the average of infinitely many Wasserstein distances between one-dimensional projections of high-dimensional distributions. Formally, for any $\mu$ and $\nu$ 
the Sliced Wasserstein is:

\rtypo{\begin{equation} 
    \mathcal{SOT}_p^p(\mu, \nu) = \mathbb{E}_{{\vector{\rv{\uptheta}}} \sim \text{Unif}(\mathbb{S}^{d-1})} \Big[ \mathcal{W}_p^p({g_{\vector{\rv{\uptheta}}}}_{\#} \mu, {g_{\vector{\rv{\uptheta}}}}_{\#}  \nu) \Big] 
\label{eq:wasserstein_sliced}
\end{equation}}
where \rtypo{$g_{{\vector{\rv{\uptheta}}}}(\vector{\rv{x}})$} 
is the linear projection map parameterized by \rtypo{${\vector{\rv{\uptheta}}}$}.
Here \rtypo{${g_{\vector{\rv{\uptheta}}}}_{\#} \mu$} denotes the pushforward measure of $\mu$, i.e. the probability measure associated with the random variable \rtypo{$g_{\vector{\rv{\uptheta}}}(\vector{\rv{x}})$}, where \rtypo{$\vector{\rv{x}}$} is a random variable with probability measure $\mu$.
$\text{Unif}(\mathbb{S}^{d-1})$  means uniformly distributed over the surface of a \rtypo{$d$}-dimensional unit sphere $\mathbb{S}^{d-1}$. Since the projections of measures on the picked direction  \rtypo{(${g_{{\vector{\rv{\uptheta}}}}}_{\#} \mu$ and  ${g_{{\vector{\rv{\uptheta}}}}}_{\#} \nu$ )} are one-dimensional, the $\mathcal{W}_p^p$ in \eqref{eq:wasserstein_sliced} can be efficiently calculated using \eqref{eq:wasserstein_1d}. In practice, however, acquiring an infinite number of projections is not feasible, and thus \eqref{eq:wasserstein_sliced} is usually approximated using a Monte Carlo scheme. The integral is replaced with the average calculated over \rtypo{a finite} number of $L$ random projection directions: 
\rtypo{\begin{equation} 
    \widehat{\mathcal{SOT}}_p^p(\mu, \nu) = \frac{1}{L} \sum_{l=1}^L  \mathcal{W}_p^p({g_{{\vector{\rv{\uptheta}}}_l}}_{\#} \mu, {g_{{\vector{\rv{\uptheta}}}_l}}_{\#}  \nu) 
\label{eq:wasserstein_sliced_mc}
\end{equation}}
where ${\vector{\rv{\uptheta}}}_1, \cdots , {\vector{\rv{\uptheta}}}_L \sim \text{Unif}(\mathbb{S}^{d-1})$. 
It might be surprising that OT on the real line (i.e., sliced OT) can convey geometric information of high dimensional distributions. However, theoretical studies have shown that SW satisfies the metric axioms \cite{Bonnotte2013UnidimensionalTransportation} and has convergence properties similar to that of the Wasserstein distance \cite{Bayraktar2021StrongType, NadjahiAsymptoticDistance}. Further theoretical results \cite{NadjahiStatisticalDivergences} show that SW has better sample complexity that does not depend on the problem dimension.

\textit{Applications:} The Sliced Wasserstein is easier to compute and enjoys theoretical properties similar to that of the Wasserstein distance. Thus, it is used in many ML applications as a better alternative. A few examples include defining kernels \cite{Kolouri2016SlicedDistributions},  generative models \cite{DeshpandeMax-SlicedGANs, HeitzASynthesis}, pooling \cite{NaderializadehPoolingEmbedding}, neural texture synthesis \cite{HeitzASynthesis}, model selection \cite{TranEURECOMSimoneRossiEURECOMDimitriosMiliosModelAutoencoders}, learning representations of 3D point clouds \cite{NguyenPoint-setClouds} and \marginpar{R1}\rref{sets \cite{NaderializadehSetEmbeddings}}, and domain adaptation \cite{LeeSlicedAdaptation}. Additionally, it has been employed \rtypo{for alleviating} the computational burden of expensive OT formulations such as Wasserstein barycenter \cite{Bonneel2015SlicedMeasures} and Gromov Wasserstein \cite{VayerSlicedGromov-Wasserstein} (\cref{sec:gromov_wasserstein}).

\noindent\textit{Limitations:} Research on extensions of SW is mostly concerned with two questions linked to limitations of \rtypo{\eqref{eq:wasserstein_sliced_mc}: 1) how do we better determine directions for projections ${\vector{\rv{\uptheta}}}_l$? and 2) what could be a better mapping than the linear $g_{{\vector{\rv{\uptheta}}}}$?}

What motivates the first question is the increased \textit{projection complexity} of randomly chosen ${\vector{\rv{\uptheta}}}_l$. In other words, to achieve a good approximation of \eqref{eq:wasserstein_sliced_mc} we need a larger number of projections $L$ \cite{DeshpandeMax-SlicedGANs}.
\rtypo{One can consider a few informative projections to overcome this limitation instead of using all random ones. }Following this idea, Max Sliced Wasserstein (Max-SW)\cite{DeshpandeMax-SlicedGANs} advocates for the single ``best direction''. For order $p=2$, it is defined as follows:
\rtypo{\begin{equation} 
    \max \mathcal{SOT}_2^2(\mu, \nu) = \max_{\vector{\theta} \in \mathbb{S}^{d-1}} \mathcal{W}_2^2({g_{\vector{\theta}}}_{\#} \mu, {g_{\vector{\theta}}}_{\#}  \nu) 
\label{eq:max_sliced_wasserstein}
\end{equation}}
where the best direction is one that yields the largest Wasserstein distance between the projected measures. Finding this best direction isn't trivial. In practice, it is replaced with direction, resulting in the largest difference between the means of the projected measures. Distributional Sliced Wasserstein (DSW) \cite{DistributionalOpenReview} claims that focusing only on the most important direction (e.g., Max-SW) ignores other potentially relevant directions.  Thus, it takes a middle ground between SW and Max-SW by searching for a ``distribution'' of the important directions on the unit sphere.

The second question regarding SW limitations concerns the linear \rtypo{$g_{\vector{\theta}}$}. What motivates this is the hypothesis that nonlinear mappings could be better in high-dimensional settings. Generalized Sliced Wasserstein (GSW) \cite{KolouriGeneralizedDistances} formalizes this idea. Also, it is possible to combine nonlinear projection with better projection complexity. Max Generalized Sliced Wasserstein (max-GSW)\cite{KolouriGeneralizedDistances} is an example of this trend.

\subsection{Wasserstein Barycenter}
\label{sec:WBC}
\textit{Motivation}:
The barycenter of a collection of measures is an intuitive notion of an ``average" of the measures. For a set of vectors, the usual average of the vectors is the single vector $v$ that minimizes the sum of squared Euclidean distances between $v$ and every vector in the set. The barycenter of a set of measures provides a useful first-order summary statistic of a set of measures, just as the average of a set of vectors provides a useful summary statistic of those vectors.

\textit{Formulation and Characteristics}:
For a set of measures $\{ \mu_i \}_{i=1}^N$, we write the weighted Wasserstein barycenter $\mu_\ast$
\begin{align*}
    \mu_\ast = \inf_{\mu} \sum_{i=1}^N w_i W_2^2(\mu, \mu_i) \numberthis \label{eq:barycenter}
\end{align*}
with nonnegative weights $\{ w_i \}_{i=1}^N$ such that $\sum_{i=1}^N w_i = 1$. Note that each Wasserstein distance itself involves an infimum. The existence and uniqueness of solutions to~\eqref{eq:barycenter} appears to have first been studied by~\cite{Agueh2011BarycentersSpace}. They also provide a dual formulation and provide solutions to the problem when each $\mu_i$ is zero-mean Gaussian.

\textit{Computation}:
Unfortunately, the Wasserstein Barycenter inherits the usual computational cost associated with optimal transport distances. One way to circumvent this issue is to consider entropy regularized divergences (see \S~\ref{sec:regularized_ot}) instead of $W_2^2$, so that the Sinkhorn algorithm can be exploited. Regularization in this manner is considered by~\cite{Janati2020DebiasedBarycenters}. They show that in the Gaussian case, entropic regularization with uniform reference measures~\eqref{eq:ent_regularized_primal} leads to a blurry (or biased) Barycenter, but entropic regularization of the form~\eqref{eq:sinkhorn_divergence} does not suffer from this problem. \marginpar{R1}\rref{\cite{cuturi2014fast, benamou2015iterative} proposed efficient algorithms for Wasserstein barycenter. Claici \etal \cite{claici2018stochastic} proposed a stochastic algorithm that can extend to continuous distributions. Leveraging deep Euclidean embeddings and auto-encoders, \cite{courty2018learning} proposed an efficient estimation of exact Wasserstein barycenters.}
\marginpar{R3}\rref{ Iterative algorithms such as \cite{alvarez2016fixed,zemel2019frechet}  were also used to compute Wasserstein Barycenter. Follow-up works such as \cite{chewi2020gradient} addressed their lack of theoretical guarantees.
Anderes \etal \cite{anderes2016discrete}
develops theoretical results for discrete Wasserstein barycenters case.}

The semi-dual formulation of OT  \rtypo{(see  Eq. (3) in \cref{sec:ot_background} )} can be used to cast~\eqref{eq:barycenter} as a minimum over a minimax problem~\cite{Fan2020ScalableNetworks}. Using an input convex neural network (which are universal approximators in the space of convex functions) allows one to search over the set of convex functions. This approach suffers from the usual theoretical intractability of optimization problems associated with neural networks, but in practice, the method appears competitive on toy problems. The computational complexity per training iteration is $\mathcal{O}(NMp)$, where $N$ is the number of marginal distributions, $M$ is the batch size, and $p$ is the neural network size.

\marginpar{R3}\rother{Note that the Wasserstein barycenter can be cast a Multi-marginal Optimal Transport (MOT) problem; a generalization of OT that aligns multiple probability measures (check \cite[Eq.(1),(2)]{pass2015multi} and \cite[10.1]{Peyre2018ComputationalTransport}). Traditionally, the MOT formulation has been used extensively in economics,  physics, and financial mathematics. Recently, the formulation has become popular in machine learning \cite{cao2019multi, mi2001multi, he2019attgan}.}

\textit{Applications}:
\cite{Fan2020ScalableNetworks} represent the Barycenter as a generative model that resembles a WGAN. Due to this, they may generate infinitely many samples from the Barycenter. \cite{Fan2020ScalableNetworks} generate samples from toy Barycenters of MNIST images and re-colored high-resolution images. However, other applications should be possible.  \rother{MOT was used in generative modeling \cite{cao2019multi}, clustering \cite{mi2001multi} and domain adaptation \cite{he2019attgan}. }

\textit{Limitations:} \marginpar{R2/R3}\rref{Computational hardness is the most obvious limitation of  Wasserstein Barycenter \cite{lin2022complexity}. Efforts to innovate scalable algorithms will definitely increase the application range of Wasserstein Barycenter}. \marginpar{R3}\rref{MOT is severely limited by the lack of
efficient algorithms. In some cases \cite{altschuler2023polynomial}, MOT can be solved in polynomial time. 
Yet, in general, MOT requires exponential time in the number of marginals and their support
sizes. Altschuler \etal \cite{altschuler2021hardness} introduced a toolkit that was used to prove the intractability of a number of MOT problems.}

\subsection{Gromov-Wasserstein (GW)}
\label{sec:gromov_wasserstein}
\textit{Motivation:} OT (\ref{eq:kantorovich}) assumes a common meaningful distance $c$ exists for the points in the source and target measures \cite{VayerSlicedGromov-Wasserstein}. Thus implicitly assuming that the two supports lie in the same metric space. This can be limiting in many applications such as cross-domain alignment \cite{TruongTheApproach} and unsupervised bi-lingual lexical induction \cite{Alvarez-Melis2018Gromov-WassersteinSpaces} where we seek an alignment between sets of word embeddings from two \textit{different} languages without access to parallel data. This calls for \textit{relational OT formulation} that measures how distances between pairs of words are mapped across languages.

In the lack of a common distance function, $d$ one might seek a solution that relies on the relevant distance functions $d_{\mathbb{X}}, d_{\mathbb{Y}}$ (i.e., metric spaces) corresponding to the measures considered. Then, define the transportation cost based on the relationships between distances. This turns the problem into a notion of distances \marginpar{R1}\rref{between metric spaces \cite{Memoli2011GromovWassersteinMatching} \cite{lIHESGroupsTits}}. Notably, this offers a way for comparing originally incomparable spaces. Additionally, this notion makes the resulting alignment invariant with respect to arbitrary distance-preserving transformations (e.g., rotations).

\textit{Formulation and Characteristics:} Gromov-Wasserstein \cite{Memoli2011GromovWassersteinMatching} generalizes OT distance to a notion of \marginpar{R1} \rref{\textit{distance between distances} \cite{Memoli2011GromovWassersteinMatching}}
The $\mathcal{GW}$ is defined as
\rtypo{\begin{align*}
    &\phantom{{}={}}\mathcal{GW}(\mu,\nu) \\
    &= \inf\limits_{\pi \in \Pi(\mu, \nu)} \int\int_{(\mathbb{X} \times \mathbb{Y})^2} r(\vector{x},\vector{x}^\prime,\vector{y},\vector{y}^\prime) d\pi(\vector{x}, \vector{y}) d\pi(\vector{x}^\prime, \vector{y}^\prime) , \numberthis \label{eq:GW}
\end{align*}}
where \rtypo{$r(\vector{x},\vector{x}^\prime,\vector{y},\vector{y}^\prime) = d( d_{\mathbb{X}}(\vector{x},\vector{x}^\prime),d_{\mathbb{Y}}(\vector{y},\vector{y}^\prime))$ is the relational distance.} When $d=L_{2}$, $\mathcal{GW}^{\frac{1}{2}}$ is a proper distance \cite{Memoli2011GromovWassersteinMatching}. 
\marginpar{R1}\rref{In addition to the formula above, there exist variations of Gromov Wasserstein for handling unbalanced and partial transport \cite{sejourne2021unbalanced,Chapel2020PartialLearning}}.\rother{Note that $\mathcal{GW}$ is very relevant to the so called Quadratic Assignment Problem (QAP) \cite{koopmans1957assignment} which is also linked with the extensive literature of graph matching problems. Apart from a few special cases solvable in polynomial time, the QAP is known to be NP-Hard. }

\textit{Computation:} Naively solving the $\mathcal{GW}$ formulation (\eqref{eq:GW}) involves prohibitive computation ($\mathcal{O}(N^2M^2)$) 
involving  \rref{fourth-order tensor $ $ product \cite{Memoli2011GromovWassersteinMatching}}
. Recent techniques for scaling $\mathcal{GW}$ to big graph and point cloud datasets are discussed in \cref{sec:scaling_ot}. 

\textit{Applications:} On the one hand, $\mathcal{GW}$, in general, bypasses a classical OT limitation and allows alignment across "incomparable" spaces \cite{Vayer2020ASpaces, LearningSpaces} (e.g. spaces of different dimensionality or data type). Given this, it has found increased adoption in cross-domain (heterogeneous domains) alignment. In single Cell Omics \cite{Demetci2020Gromov-WassersteinData}, $\mathcal{GW}$ was used to align heterogeneous measurements from gene expression and Chromatin accessibility. In \cite{Chapel2020PartialLearning} partial $\mathcal{GW}$ was shown to be efficient for matching point clouds when they exist in different domains and have different features. Very recently, $\mathcal{GW}$ was leveraged in Cross-Domain Imitation
Learning \cite{Fickinger2021Cross-DomainTransport} to align and compare states between the different spaces of RL agents. A Gromov extension of Dynamic Time Warping \cite{CohenAligningSpaces} was used to align incomparable time series by leveraging the intra-relational geometries.  An example is multi-modal data, such as the configuration space of a robotic arm and its representation as pixels of a video frame. 

On the other hand, leveraging within-domain similarity  $( d_{\mathbb{X}}$ and $d_{\mathbb{Y}})$ allows capturing the topological relations. This can be beneficial for learning relations.
Such a relational \rtypo{perspective is also} why $\mathcal{GW}$ is used in graph applications. $\mathcal{GW}$ has thrived in applications such as graph classification\cite{Chen2020OptimalGraphs}, graph partitioning \cite{XuScalableMatching}, graph node embedding \cite{XuGromov-wassersteinEmbedding} and supervised graph prediction \cite{Brogat-MotteLearningBarycenters} and others.

\textit{Limitations:} Relying only on the topological or relational aspects encoded by pairwise similarities during alignment and ignoring the features is one of the limitations of GW (check \marginpar{R1} \rref{Fused Gromov-Wasserstein \cite{Chen2020OptimalGraphs} }below). Computationally, GW turns the OT linear program into a non-convex quadratic program with a prohibitive computational cost.
From a computational perspective, as noted earlier the naive implementation of GW requires operating on a fourth-order tensor. This can be of less concern only in a few cases. For example, when working on small-to-moderate data size \cite{Alvarez-Melis2018Gromov-WassersteinSpaces}.  Also, for certain classes, \cite{PeyreGABRIELPEYRE2016Gromov-WassersteinMatrices}, the cost can be brought down to $\mathcal{O}(n^3)$  where $n$ is the number of points. Going beyond these, GW can be expensive, and searching for ways to speed up the calculation is very active.  More discussion about this is provided in Scaling OT (\cref{sec:scaling_ot}).

\begin{figure*}[!t]
    \centering
    \includegraphics[width= 0.95\linewidth]{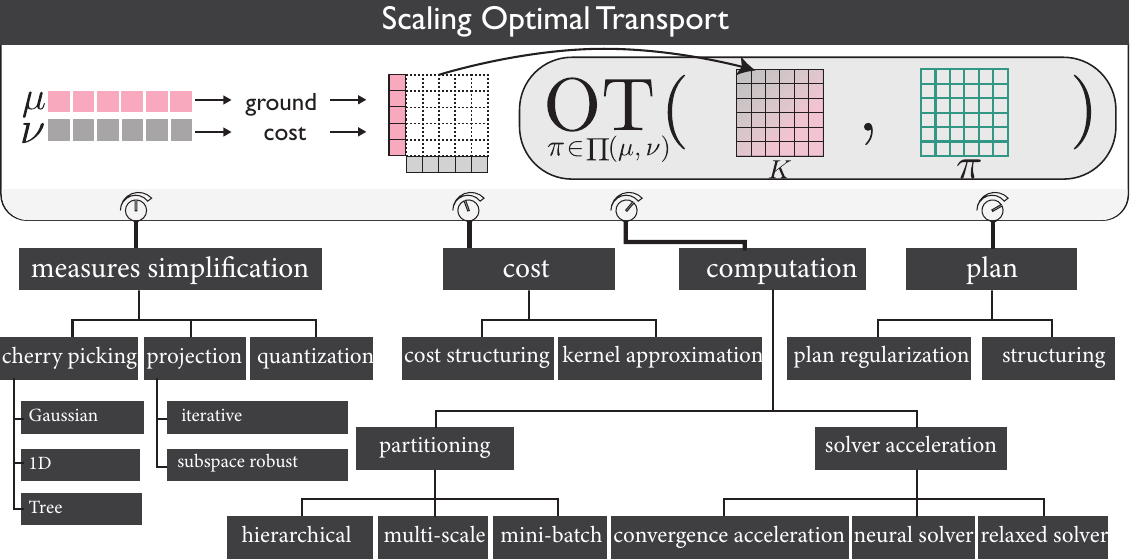}
    \vspace{-0.5em}
    \caption{\textbf{OT Scaling.} Exaggerating a bit, we can scale all OT methods by turning one of the knobs (i.e., optimizing components) of the OT computation machinery. This can be done through  (\cref{sec:scaling_measures} \textbf{measures simplification}) simplifying the input measures, (\cref{sec:scalingot_cost_approx} \textbf{cost}) structuring the ground cost or approximating the associated kernel, (\cref{sec:scalingot_structure} \textbf{plan}) enforcing a specific prior on the optimal plan or (\cref{sec:scalingot_paritition} \textbf{computation}) accelerating the optimization. Of course, it is also possible to turn multiple knobs simultaneously and have (\cref{sec:scalingot_combined})a \textbf{combined} approach. }
    \label{fig:scaling_ot}
    \vspace{-1em}
\end{figure*}

\rtypo{In section \ref{sec:common_ot_formulations}, we covered common OT formulations. In this fast-growing field, it is impossible to turn over every stone.  Formulations that were left out include those that emerged from multiple primitive formulations (
e.g., Fused-Gromov-Wasserstein-Barycenter \cite{Brogat-MotteLearningBarycenters} ),  the dynamic formulations \cite{benamou2000computational}.} \marginpar{R2/R3} \rref{Additionally, there is a continual effort to extend OT to new domains and develop specialized formulations. Examples include functional OT \cite{zhu2021functional}, OT on splines \cite{chen2018measure}, and many others.}

\section{Scaling OT Computation}
\label{sec:scaling_ot}

\rtypo{In this section, we discuss various approaches for scaling OT to handle large datasets and high dimensional data. We focus mainly on OT's computational complexity. OT's statistical complexity is another challenge that we don't cover here. While computational complexity focuses on the efficiency and feasibility of computing the optimal transport solution, statistical complexity deals with the reliability and accuracy of these solutions in the context of data analysis and inference.}
\marginpar{R3} \rref{Readers interested in statistical complexity research can consult sources like \cite{mena2019statistical} , \cite{NadjahiStatisticalDivergences}, and \cite{ahidar2020convergence}. }

\marginpar{R1}\rother{Looking at the big picture of the literature for scaling OT \textit{computation},  one can argue that all the proposed approaches are either approximating the OT solution or tailoring it to a specific case that is easier to handle.} The concrete techniques to achieve this can be categorized into five main themes. Fig.~ \ref{fig:scaling_ot} (top) illustrates a roadmap of OT computation scaling techniques. It is possible to guess where the optimization can happen. One can work on the individual components corresponding to the  \textit{measures}, the \textit{plan}, and the \textit{ground cost} or accelerate the \textit{formulation optimization} by, for example, partitioning. \textit{Combining} some of these approaches can help and some papers fall \marginpar{R1}\rtypo{into multiple groups}.

\subsection{Measures Simplification}
\label{sec:scaling_measures}

A logical first stop for scaling OT is to consider simplifying the input to the OT computation pipeline; the measures. Such a simplification can take the following forms.

\textit{Cherry Picking (measures with closed form solution):}  The simplest form of measure simplification is picking from a special set of measures that have closed form solutions. As discussed earlier, the one-dimensional case has a closed form (\eqref{eq:wasserstein_1d}). The whole line of Sliced OT (\cref{sec:sliced_wasserstein}) builds on this fact. The closed form solution for the 2-Wasserstein metric for multivariate Gaussian distributions with means $\vector{m}_1, \vector{m}_2$ and covariance matrices $\matrix{\Sigma}_1, \matrix{\Sigma}_2$ is given by:
\rtypo{\begin{multline}
    W_2(\mathcal{N}_1, \mathcal{N}_2)^2 = \\ \left\|\vector{m}_1-\vector{m}_2\right\|_{2}^{2}+ \text{Tr}\left(\matrix{\Sigma}_{1}+\matrix{\Sigma}_{2}-2\left(\matrix{\Sigma}_{1}^{\frac{1}{2}} \matrix{\Sigma}_{2} \matrix{\Sigma}_{1}^{\frac{1}{2}} \right)^\frac{1}{2}\right)
   \label{eq:gaussian_wasserstein_2}
\end{multline}}
where $\matrix{\Sigma}_1^\frac{1}{2}$ is the unique positive definite square root matrix such that $\matrix{\Sigma}_1^\frac{1}{2} \matrix{\Sigma}_1^\frac{1}{2} = \matrix{\Sigma}_1$.

\marginpar{R3}\rtypo{Recent fundamental investigations have revealed the existence of closed-form solutions for regularized OT on Gaussians \cite{Mallasto2022Entropy-regularizedMeasures},}
\rtypo{, the entropic Gromov-Wasserstein on Gaussians \cite{LeEntropicDistributions} and others.} Another case is the tree Wasserstein for measures that can be supported on a tree, used mostly in document classification tasks \cite{YamadaApproximatingTrees} as it permits fast comparison of a large number of documents.  
Tree-Wasserstein can be computed in linear time with respect to the number of nodes in the tree \cite{SatoFastTree, TakezawaFixedBarycenter}. Note that this ignores the computational and storage requirements for transforming measures into trees. Noteworthy variants include a sliced version \cite{LeRIKENAIPTree-SlicedDistances}  calculated by averaging the Wasserstein distance between the measures using random tree metrics, a fast supervised version that can be trained end-to-end\cite{Takezawa2021SupervisedDistance}, fixed-support Wasserstein Barycenter on a tree \cite{TakezawaFixedBarycenter} and an efficient solver \cite{SatoFastTree} for unbalanced OT with the capability of processing one-million point measures on CPU.

Since the OT solution for these favorable cases is analytically known, they are employed in many applications. Additionally, they are used to alleviate \rtypo{the computational} burden of other OT formulations \cite{Alvarez-Melis2020GeometricTransport}. Some other usage examples of simple OT problems with analytic solutions include being used as test cases for establishing the correctness of OT solvers \cite{Flamary2021POT:Transport}, and for providing pre-training data for supervised neural OT solvers \cite{bunnesupervised}. Despite the advantages, cherry-picked measures can be unrepresentative of real-world data. Also, solving certain OT problems in higher dimensions can still be expensive even for closed-form situations. For example, calculating the barycenter of Gaussians \cite{AltschulerAveragingDescent} incurs a cubic complexity in the dimension \cite{Korotin2022WassersteinEstimation}.

 \textit{Measures Projection:}  Low-dimensional projection is another conceptually simple candidate for addressing the curse of dimensionality. The recent findings of spiked transport \rtypo{model \cite[Theorem 1]{Niles-WeedEstimationModel} provide a theoretical grounding for this approach. }
 \rtypo{Projection robust methods family resides in this research direction.} Subspace Robust Wasserstein (SRW) \cite{PatySubspaceDistances} builds on the principles of Sliced Wasserstein (SW \cref{sec:sliced_wasserstein})  and extends the projection to a subspace of dimension $k \geq 2$. SRW seeks the subspace that maximizes the OT cost between two measures to capture the major discrepancy (similar in spirit to max-SW \cref{sec:sliced_wasserstein}). SRW shares some properties with the 2-Wasserstein distance while being more robust to random perturbations in data. In Wasserstein Barycenter, \cite{IzzoDimensionalityBarycenter} showed that projecting the distributions into low dimensional spaces can provably preserve the quality of the barycenter. Projection robust Wasserstein barycenter \cite{HuangProjectionBarycenters} solves the fixed-support discrete Wasserstein \rtypo{barycenter} problem. 

One issue when working with subspaces is that the \rtypo{subspace plan} is optimal between the projected and not the original high dimensional measures. This can be addressed by considering only the plan on the original space that is constrained to be optimal after the projection. This is the key idea behind the concept of \textit{subspace detours} that was introduced by \cite{MuzellecSubspaceProjections} and later extended to other formulations such as Gromov-Wasserstein in \cite{Bonet2021SubspaceGromovWasserstein}. 

In cases where the input measures can be grouped \rtypo{and} the grouping is known \rtypo{apriori}, one can further speed up SRW and gain additional advantages. Feature Robust OT (FROT) \cite{PetrovichFeatureData} speeds up OT computation and makes it more robust to noise. While similar in spirit to SRW, FROT is convex and more scalable than SRW.

Another projection-based family of methods uses iterative random projection, where informative projections are determined sequentially. In this category, there is the Projection Pursuit Monge Map (PPMM) method \cite{MengLarge-scalePursuit} that picks the best projection in the $k$th iteration guided by information (e.g., residuals) from $k-1$ projections. \rtypo{Check \cite{Zhang2022ProjectionbasedProblems} for a focused tutorial on iterative projection approaches}.

\textit{Measures Quantization:} Another simplification approach is to quantize the measures before computing OT.  Examples include \cite{BeugnotImprovingQuantization, AboagyeQuantizedSpaces } over-sampling the input measures and applying the solver only on their summary (i.e., a k representative samples acquired by k-means like quantization). Usually, it is easy to obtain more samples when the data come from large datasets or generative models. The approach is solver agnostic in the sense it can be applied to different formulations (OT vs regOT) with no change \cite{BeugnotImprovingQuantization}. Along the same line, a simplification that considers transporting the measures centroids (instead of the actual measures) can be more efficient. For example, Word Centroid Distance \cite{KusnerFromDistances} speeds up the OT computation by representing each document as the weighted average of its word embedding.

In addition to the approaches discussed for measure simplification,  \textit{smoothing} is being explored. Smoothing simplifies measures by adding Gaussian noise. Gaussian Smoothed OT (GOT) \cite{GoldfeldGaussian-smoothedEfficiency} simply convolves the input measures with isotropic Gaussian kernel before performing the computation. Research investigation into the theoretical properties of the formulation \cite{DingAsymptoticsRegime} and potential applications \cite{Rakotomamonjy2021DifferentiallyDistance, Goldfeld2020AsymptoticDistance} is ongoing. A notable trait is alleviating the curse of dimensionality by enjoying a convergence rate $n^{-1/2}$ in all dimensions \cite{GoldfeldConvergenceEstimation}.

\subsection{Structuring the Optimal Plan}
\label{sec:scalingot_structure}

 Going beyond input pre-processing and looking into the OT optimization itself,~\eqref{eq:wasserstein_distance}, OT computation involves an optimization over the infinite-dimensional set of couplings $\pi \in \Pi (\mu,\nu)$ between the measures $\mu$ and $\nu$. Enforcing a structure or regularization on the optimal plan can often lead to structural changes in the optimization \rtypo{problem, making} it simpler to solve. For example, searching for the transport plan only in a low-rank sub-space \cite{Scetbon2021Low-RankFactorization,LiuApproximatingFactorization} can be used to reduce the dimension of the problem, speeding up the optimization process.

\textit{Plan Regularization:} The regularized OT formulation (\cref{sec:regularized_ot}) is the most famous group of plan structuring methods. The entropy regularizer (\eqref{eq:ent_regularized_primal}) biases the optimal coupling towards low entropy couplings. The algebraic properties of the added $\text{KL}$ term enable arranging the unique solution $P_{\lambda}$ in a simple form. This form can be solved using successive matrix scaling iterations. Matrix scaling iterations are significantly faster than using linear programming for solving the original OT problem. In other words, regularization changes the structure of the optimization problem, enabling the use of a faster algorithm. However, this speedup comes with a limitation. In regularized OT, there is a trade-off between convergence speed and precision, controlled by $\lambda$. As reported by many studies \cite{Klatt2020EmpiricalApplications}, larger $\lambda$ results in a faster convergence but poor approximation. On the other hand, very small $\lambda$ leads to numerical instability. To obtain a better approximation, recent works investigate the gradual reduction in $\lambda$ \cite{Xie2019ADistance, Schmitzer2019StabilizedProblems}. This simulates gradually approaching the exact (unregularized) OT which is recovered with $\lambda \to 0 $.

\begin{figure*}[t!]
    \centering
    \includegraphics[width = 0.95\linewidth]{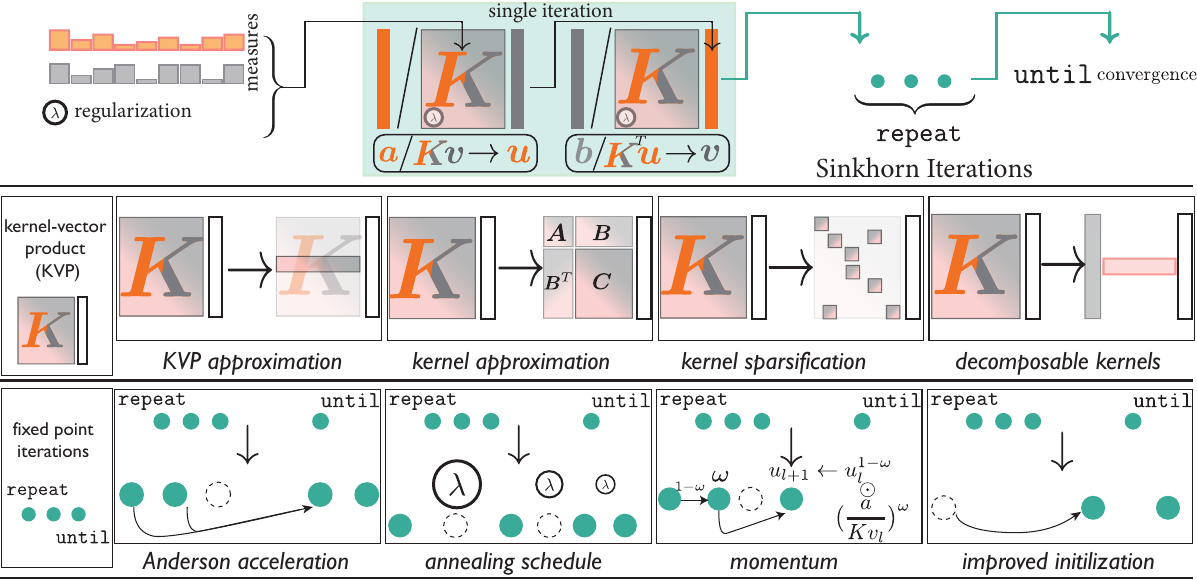}
    \vspace{-0.5em}
    \caption{\rtypo{\textbf{Speeding up Sinkhorn.} Sinkhorn procedure (top row) basically applies kernel ($\matrix{K}$) vector ($\vector{u},\vector{v}$) product operations till convergence. Given this, Sinkhorn can be accelerated through a faster kernel-vector product (KVP) or a reduced number of iterations (bottom row). Example techniques for improved kernel vector product (middle row) include KVP approximation \cite{AltschulerMassivelyMethod}, kernel approximation \cite{AltschulerMassivelyMethod}, kernel sparsification \cite{Klicpera2021ScalableMore}, or decomposable kernels \cite{Scetbon2020LinearFeatures}. Fixed point iterations can be accelerated (discussed in \cref{sec:scaling_stochastic}) using methods such as Anderson acceleration \cite{ChizatFasterDivergence}, annealing schedule \cite{Xie2019ADistance}, momentum \cite{ThibaultOverrelaxedTransport} and improved initialization \cite{Thornton2022RethinkingAlgorithm}}.
    \label{fig:sinkhorn}}
    \vspace{-1em}
\end{figure*}

\textit{Low-Rank Structuring:} Another family of algorithms uses the idea of low-rank approximate factorization of the plan. In this family, the plan is decomposed into a product of low-rank matrices, improving the statistical stability of OT and reducing the sample complexity. Factored Coupling (FC) \cite{Forrow2019StatisticalCouplings} was the first method used to impose a low-rank structure on the transport plan. This is a popular strategy in high-dimensional statistics \cite{WeedSharpDistance}. Their \textit{low-rank transport plan} \rtypo{approximates} the optimal coupling that builds on the assumption that $\mu$ and $\nu$ may only vary across a few dimensions. FC transports factored partitions or ``soft clusters'' (in which the clusters can include fractional points) of the measures, and the problem is solved as regularized 2-Wasserstein barycenters. Building on this, the authors of the Latent OT (LOT) method \cite{Lin2021MakingPoints} observed that the low-rank transport plan is also more resilient to outliers and noise. A low-rank plan transports measures through common ``anchors'' and is observed to be better at preserving clusters and tolerating outliers. In FC, the anchors are shared between the measures, and their number is dictated by the rank $r$. To improve the interpretability of the plan, LOT allows each measure to have a different number of anchors. However, the anchors are additional hyperparameters that need to be chosen carefully. A generalization of FC is the Low-Rank Sinkhorn Factorization method \cite{Scetbon2021Low-RankFactorization}; a formulation compatible with all ground costs and not only the squared Euclidean distance. \rtypo{In the Low-Rank Sinkhorn Factorization formulation, the coupling $\matrix{P}$ is explicitly factorized into a product of two sub-couplings $\matrix{Q}$ and $\matrix{R}$ linked by a common marginal $\vector{g}$, $\matrix{P} = \matrix{Q} \cdot \text{diag}(\vector{g})^{-1} \cdot \matrix{R}^T$. The solution is obtained by jointly optimizing $\matrix{Q}$, $\matrix{R}$, and $\vector{g}$.} The demonstrated computational improvement of the factorization of the transport plan in various applications motivated a further investigation \cite{Scetbon2022Low-rankDebiasing} into its theoretical properties. A debiased version of the Low-Rank Sinkhorn Factorization method \cite{Scetbon2021Low-RankFactorization} was introduced in \cite{Scetbon2022Low-rankDebiasing} and was shown to interpolate between Maximum Mean Discrepancy \cite{Gretton2022ATest} and OT.

An obvious issue with these approaches is the poor approximation when the low-rank assumption is violated (for example, when $\mu$ can be obtained by permuting the support of $\nu$). To attempt to circumvent this limitation, Liu \etal \cite{LiuApproximatingFactorization} proposed OT approximation in which the transport plan can be decomposed into the sum of a low-rank and a sparse matrix as a possible solution.

\subsection{Ground Cost Approximation}
\label{sec:scalingot_cost_approx}

 
 Application-wise, the fact that the OT distance admits a customizable ground cost has been \rtypo{leveraged in many applications. For example, in domain adaptation, we can bias similar (e.g., same-class) source domain points to be transported together to the target domain without}splitting, e.g., using a submodular cost \cite{Alvarez-Melis2018StructuredTransport}. Computationally,  having a structured cost ( perhaps one that can be decomposed) can be leveraged to improve the execution time of OT.

\textit{Structuring Ground Cost:} Enforcing a structure on the ground cost can be seen as an implicit way of reflecting that structure on the optimal plan. 
For example, a structural infinity in the cost matrix maps to a structural zero in the optimal plan, which can be used to sparsify the optimal plan. This relationship between the cost and the plan can be exploited to speed up the OT computation, reducing the memory and time requirement of the various algorithms. Relatedly, in applications that deal with a large number of samples (e.g., document and image retrieval) such that storing the pairwise distances \marginpar{R3} \rtypo{$C_{i,j}$ for all $i$ and $j$ samples can be prohibitive, a useful structure that can be used is the \textit{ground cost saturation}. Specifically, we threshold the ground cost $C_{i,j} = C_{max}$ if $c \notin r$ where $r$ denotes}the relative (nearest) samples. From a bi-partite graph matching (between the measures) perspective, this is equivalent to constraining edge connectivity and can result in significant savings. The approach was used by \cite{WernerSpeedingEmbeddings} and \cite{PeleFastDistances} for speeding up document and image retrieval; respectively.
In Multi-marginal OT (MOT), the structure of the cost tensors is exploited to speed up the computation. In sensor fusion and tracking, \cite{ElvanderMulti-marginalFusion} shows that the cost functions enjoy a structure that allows sequential decoupling, central decoupling, or a combination of both. Computing the Sinkhorn projections in these cases can be done efficiently \cite{Haasler2021MultimarginalProblem}.

\textit{Kernel Matrix Approximation:} another popular strategy is structuring the ground cost \rtypo{$\matrix{C}$} such that the dependent kernel matrix \rtypo{$\matrix{K}(\matrix{C})$} can be approximated efficiently. The Sinkhorn algorithm uses the kernel matrix \rtypo{$\matrix{K}(\matrix{C})$ }by multiplying it (or its transpose) by \rtypo{two} vectors in each iteration. \rtypo{In practice, \rtypo{$\matrix{K}$} can be very large.}
\rtypo{This drives the computational cost of the whole procedure (Figure.~\ref{fig:sinkhorn}), and a naive matrix-vector multiplication will cause each iteration to have a quadratic cost. It is natural to think of reducing the computational burden of the matrix-vector product.}

Reducing the computational cost of kernel matrix-vector product is a common problem with several known workarounds, often used in kernel-based machine learning algorithms such as Gaussian processes. In OT, the same computational tricks were adopted by some works.  Nys-Sink\cite{AltschulerMassivelyMethod} combines Nystr\"{o}m approximation \cite{WilliamsUsingMachines} with Sinkhorn scaling.
LCN-Sinkhorn \cite{Klicpera2021ScalableMore} follows in the same direction and notes that the exponential used in \rtypo{$\matrix{K}$} makes its entries negligibly small everywhere except at each (support) point's closest neighbors. This motivates sparse approximation of \rtypo{$\matrix{K}$} where only nearby neighbors are accounted for. This is equivalent to approximating the cost matrix \rtypo{$\matrix{C}$}  using a "generalized sparse" cost matrix \rtypo{$\matrix{C}^{\text{sparse}}$ where $\matrix{C}^{\text{sparse}}_{ij} = \infty$ }for ``far'' points, i.e. the structural zeros of the sparse matrix have a default value of $\infty$. Filtering ``near'' points is done via Locality Sensitive Hashing (LSH)\cite{WangHashingSurvey}; points within a certain distance $r_1$ are much more likely to be assigned to the same hash bucket than points with a distance $r_2 > r_1$.  Ultimately, this yields a version of Sinkhorn that can scale log-linearly with the number of points (i.e., $\mathcal{O}(n \log n)$). Scetbon \etal \cite{Scetbon2020LinearFeatures} considers the ground cost function of the form  \rtypo{$c(\vector{x},\vector{y}) = -\log \langle \phi(\vector{x}), \phi(\vector{y}) \rangle $ } where $\phi$ is a map from the ground space onto the positive orthant $\mathbb{R}^r_+$, with $r \ll n$. This results in a decomposable kernel ($\matrix{K}$ in algorithm \ref{alg:sinkhorn}), making \rtypo{Sinkhorn} computation more efficient. 

\subsection{Computation Partitioning} 
\label{sec:scalingot_paritition}

Computation Partitioning refers to breaking OT computation into smaller OT problems where the outcomes of the sub-problems \rtypo{are aggregated} into a solution of the original OT. Check the top panel of Fig.~\ref{fig:comp_part_and_acceleration} for visual representation. Computation Partitioning is done in the following ways:

\textit{Multi-scale, recursive, or hierarchical}: several similar methods were proposed in the literature, relying on arranging the points in a tree structure of some sort. "Multi-scale" OT was \marginpar{R3}\rref{proposed in \cite{schmitzer2016sparse}},\cite {Gerber2017MultiscaleTransport, Forum2011ATransport}. In \cite{Gerber2017MultiscaleTransport}, the transport problem is decomposed into a sequence of problems with varying scales and solved in a top-down fashion (i.e., coarse-to-fine scale). These methods proved useful when the ambient dimension $d$ is small. Yet they fail to scale to high dimensional settings as they utilize space discretization.

In ``recursive'' OT, a similar divide-and-conquer strategy performs recursive partitioning of the measures/data. Examples in this category are the recent techniques for scaling GW to big datasets \cite{XuScalableMatching}. The general theme is to partition the measures into smaller blocks, match the blocks based on their representatives (e.g., centroids), then match the paired blocks by proceeding recursively as needed (Fig.~\ref{fig:comp_part_and_acceleration} top). \rtypo{MREC} \cite{BlumbergMREC:Data} adopts this approach using black box clustering for blocking and cluster centers as block representatives. Quantized Gromov-Wasserstein (qGW) \cite{ChowdhuryQuantizedGromov-Wasserstein} supports this simple approach with theoretical guarantees based on \cite{Memoli2018SketchingSpaces} and proposes an algorithm that scales to datasets containing up to $1$ million points. It should be noted that blocking the representatives' choices (e.g., graph data community detection algorithms \cite{Garcia-Gasulla2018FluidAlgorithm} and maximal PageRank \cite{BrinTheEngine}) is crucial, and careful selection of the methods might be needed depending on the data. 
\rother{A similar tree structure is leveraged in ``hierarchical'' OT (check \cref{sec:hierarchical_ot}), but the structure is more intrinsic to the data itself.}

\textit{Mini-batch}: Sommerfeld \etal \cite{Sommerfeld2019OptimalSolvers.} shows that the average OT distances (computed by any black-box solver) on random sub-sampling of the input measures is a fast approximation of the exact OT. Theoretical non-asymptotic guarantees reveal that the approximation error can be independent of the size of the problem in some cases. This surprisingly simple approach can scale well in practice using one percent of the exact computation in images \cite{Sommerfeld2019OptimalSolvers.}. Recent OT acceleration strategies along the same line take inspiration from deep learning. \rtypo{Mini-batch OT mimics the typical training of deep neural networks by performing the computation on mini-batches}. The empirical success of mini-batch OT motivated recent theoretical investigation \cite{Fatras2021DeepAnother, Alvarez-MelisBudget-ConstrainedTransport, FatrasLearningProperties}. One concern here is the mini-batch strategy can lead to an undesirable smoothing effect for which Fatras \etal \cite{Fatras2021UnbalancedAdaptation} suggests relying on unbalanced  OT formulation to counter the issue.

\subsection{Solver Acceleration}
\label{sec:scaling_stochastic}

\begin{figure*}[t!]
    \centering
    \includegraphics[width = 0.95\linewidth]{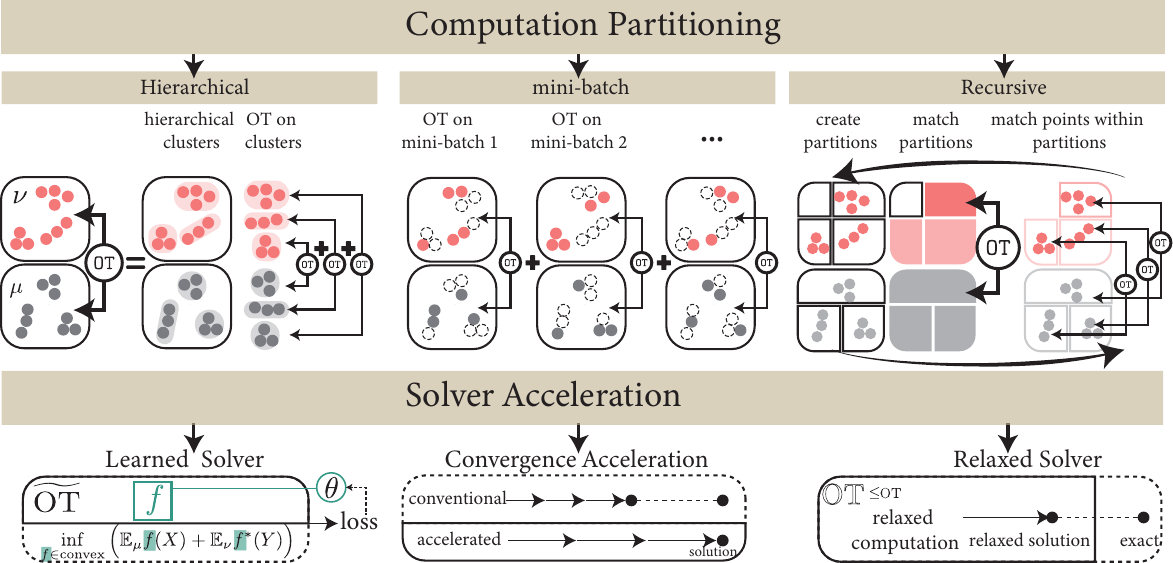}
    \vspace{-1em}
    \caption{
    \rtypo{\textbf{Accelerating OT Computation.} Without optimizing any of the previous components(i.e., the input measures, the plan, and the cost),  the OT computation can be accelerated by either breaking it into smaller OT problems (top) or accelerating the OT solver by predicting the solution quickly (bottom). Notable schemes for OT computation partitioning (top, \textbf{\cref{sec:scalingot_paritition}}) include the decomposing the problem according to a hierarchy  (\cite{Gerber2017MultiscaleTransport}), applying computation in min-batch manner (\cite{FatrasLearningProperties}) and recursive partitioning  (\cite{BlumbergMREC:Data}). Solver Acceleration  (bottom,\textbf{\cref{sec:scaling_stochastic}} leverages the recent neural solvers (\cite{VardhanMakkuva2020OptimalNetworks,Vacher2021ConvexCriterion}), accelerates the optimization (e.g., Anderson Acceleration \cite{ChizatFasterDivergence}), or seeks solution to a relaxed version of OT problem (\cite{KusnerFromDistances, AtasuLinear-complexityAcceleration}).
    }}
    \label{fig:comp_part_and_acceleration}
    \vspace{-1.5em}
\end{figure*}

Solver acceleration targets optimizing the computation process either by learning to predict the solution quickly (usually for specialized problems) or admitting a relaxed version of the original problem, which is easier to compute. In this category, we identify \textit{convergence acceleration}, \textit{learned solvers}, and \textit{relaxed solvers}. A depiction is shown in Fig.~\ref{fig:comp_part_and_acceleration} bottom panel.

\textit{Convergence Acceleration.} Convergence Acceleration aims at bringing the solver to convergence in fewer iterations. In regularized OT, for example,  we can use a number of strategies to accelerate the  Sinkhorn algorithm (see Fig.~\ref{fig:sinkhorn} bottom row). The Sinkhorn algorithm (Alg.~\ref{alg:sinkhorn}) is fundamentally a fixed point iteration (FPI) algorithm for finding fixed points $(\vector{u}^*, \vector{v}^*)$. Therefore, all FPI algorithm acceleration techniques can be investigated. In \cite{ChizatFasterDivergence}, the classical Anderson acceleration for FPI algorithms was used. Anderson acceleration can speed up FPI algorithms by making them more robust to residual oscillation. Besides Andreson acceleration, the over-relaxed Sinkhorn method~\cite{ThibaultOverrelaxedTransport} adopts a relaxed FPI algorithm where the auxiliary vector update $\vector{u}_{l+1} \leftarrow \frac{\vector{a}}{\matrix{K}\vector{v}_{l}} $ is replaced with the relaxation $\vector{u}_{l+1} \leftarrow \vector{u}_l^{1-\omega} \odot \big(\frac{\vector{a}}{\matrix{K}\vector{v}_l}\big)^{\omega}$, where $l$ denotes the iteration and $\omega > 0$ is the chosen relaxation parameter. \cite{ThibaultOverrelaxedTransport, PeyreQuantumTransport} show that for $\omega > 1$ the algorithm can converge faster than traditional Sinkhorn. Recently, \cite{Lehmann2021AAlgorithm} identified an apriori range for $\omega$ for which global convergence is guaranteed. Other fixed point acceleration techniques were investigated in \cite{Tang2022AcceleratingReview}. In the Sinkhorn literature, better initialization has been mostly neglected as the potential speedup seems insignificant given the convex nature of regularized OT. However, a recent work \cite{Thornton2022RethinkingAlgorithm} showed that careful initialization of the Sinkhorn subroutine (in a neural network) initialized by pre-training on  
 problems with closed-form solutions (Gaussians and 1D) can lead to faster convergence. The technique can also be complementary to previous techniques. 
 The previous approaches have little in common apart from seeking faster convergence. Yet, they all are holistic approaches that consider the inner \rtypo{workings} of the solver and a potential revisit of the original formulation \cite{AnEfficientDescent} before formulating an acceleration strategy. 

\textit{Learned Solvers:} A promising direction that \rtypo{has been} accumulating interest recently is based on the idea of ``neural solvers''. Neural solvers are neural networks that attempt to predict the OT output from the input measures. Data-driven neural solvers reduce the expensive traditional computation by leveraging knowledge from past OT solutions. By constructing a model and training it on ground truth OT solutions and parameters, we \rtypo{can quickly predict OT solutions} during test time (single forward pass) without working it out from the measures from scratch (iterative) every time. \marginpar{R1} \rref{Courty \etal \cite{courty2018learning} learns a supervised autoencoder that embeds MNIST images such that the Euclidean distance in the latent space approximates the Wasserstein distance}. The learned approach becomes more attractive in deployment scenarios that \rtypo{require} solving OT repeatedly (e.g., coupling stream of image pairs). \marginpar{R1}\rref{In addition to the computational improvements, a notable advantage of neural \cite{DoBenchmark, LiContinuousBarycenters} and stochastic \cite{GenevayStochasticTransport} solvers is the ability to cope with parametric continuous $\mu$ and  $\nu$ (densities)}. Empirically, neural solvers show great success in generative modeling (\cref{sec:gen_modeling}) and domain adaptation (\cref{sec:domain_adaptation}).

\textit{Grey-box Solvers:} These approaches replace a component of the OT formulation by a neural network, re-parameterizing the optimization to be in terms of the weights and biases of the neural network. Some of these semi-transparent models balance the regular OT computation, which can be expensive and hard to scale, and the data-rich (yet opaque) black-box learned solvers. So far in the literature, the Kantorovich-Rubinstein duality \eqref{eq:kantorovich_rubinstein} and the semi-dual formulation \eqref{eq:regualized_Kantorovich_dual} \cite{SeguyLARGE-SCALEESTIMATION,Arjovsky2017WassersteinNetworks} have received the most attention. This is because the (semi-)dual solution is comprised of 2 pointwise solutions: a dual value for each $\vector{x}$ in the support of the measure $\mu$, and another dual value for each $\vector{y}$ in the support of the measure $\nu$. This pointwise structure motivates using a neural network to replace the mapping from $\vector{x}$ and/or $\vector{y}$ to their respective dual solutions. However, formulation tricks need to be employed here to eliminate any constraints or non-smoothness in the (semi-)dual formulation. Current works alleviate this issue either by relying on soft constraints \cite{LiContinuousBarycenters} or structuring the neural solver to ensure feasibility \cite{VardhanMakkuva2020OptimalNetworks, Arjovsky2017WassersteinNetworks}. 

\textit{Grey-Box Solvers with Soft Constraints:} To understand the strategy of these solvers, we revisit the Kantorovich dual:
\rtypo{\begin{align*}
    \underset{\phi \in L^1(\mu) ,\psi  \in L^1(\nu) } \sup
    \mathbb{E}_{(\vector{\rv{x}},\vector{\rv{y}})\sim \mu \times \nu} \big[ \phi(\vector{\rv{x}}) + \psi (\vector{\rv{y}}) \big] \\
    \text{subject to}~\phi(\vector{x})+\psi(\vector{y}) \leq  c(\vector{x},\vector{y})~\text{for all}~(\vector{x},\vector{y}).
    \numberthis
\label{eq:kantorovich_dual_2}
\end{align*}}
The formulation shows the OT computation as maximizing an expectation \rtypo{term, thus} suggesting that the maximization can be achieved using stochastic gradient methods on sampled batches from the coupling $\mu \times \nu$. However, it is not clear how the constraints on $\phi$ and $\psi$ can be satisfied in this scheme. It turns out that the dual of the entropy regularized OT problem is unconstrained and doesn't exhibit this problem,~\eqref{eq:regularized_primal}, which can be obtained by the Fenchel-Rockafellar theorem:
\rtypo{\begin{align*}
    \underset{\phi \ ,\psi \ }\sup
    \mathbb{E}_{(\vector{\rv{x}},\vector{\rv{y}})\sim \mu \times \nu} \big[ \phi(\vector{\rv{x}}) + \psi (\vector{\rv{y}})  + R_{\lambda} \big] 
\numberthis
\label{eq:regualized_Kantorovich_dual}
\end{align*}}
where $R_{\lambda} = -\lambda e^{\frac{1}{\lambda}(\phi(\vector{x})+\psi(\vector{y})-c(\vector{x},\vector{y}))}$ is a penalty term. Intuitively, $R_{\lambda}$ represents a smooth version of the constraints in  \eqref{eq:kantorovich_dual_2}. Thus, we bypass the need to enforce the hard constraints. Given that the problem is now unconstrained, one can simply parameterize the dual variables $\phi$ and $\psi$ using neural networks and use stochastic \rtypo{gradient} methods to maximize the objective in (\ref{eq:regualized_Kantorovich_dual}) \cite{SeguyLARGE-SCALEESTIMATION}.

Alternatively, the semi-dual formulation \eqref{eq:regualized_Kantorovich_dual}  of the unregularized OT problem is also unconstrained, but its objective is non-smooth, so a smooth approximation of the semi-dual formulation of the unregularized OT problem can also be used. The semi-dual's objective is not smooth because it uses the $\max$ operator. \marginpar{R?} \rref{One way \cite{Blondel2018SmoothTransport} to alleviate this} is to construct a smooth approximation of the $\max$ operator using the conjugate function of some strictly convex regularization/penalty function $\Omega$ over the unit Simplex $\Delta$. The entropy or L2 norm are possible choices for $\Omega$. The following function is known as the conjugate function of $\Omega(\vector{x})$: 
\rtypo{\begin{align}
    \text{max}_{\Omega}(\vector{x}) = \sup_{\vector{y} \in \Delta} x' \cdot y - \Omega(\vector{y}).
\end{align}}
The above conjugate function is smooth and differentiable in $\vector{x}$ and is equal to $\max(\vector{x})$ when $\Omega(\vector{y}) = 0$ for all $\vector{y}$ (i.e. no regularization). Replacing $\max$ with $\text{max}_{\Omega}$ for some choice of $\Omega$ in the semi-dual results in a smooth unconstrained optimization problem. The gradient of the above function can also be analytically derived for common choices of $\Omega$.

\textit{Grey-Box Structured Solvers:} These solvers rely on structural constraints in the neural model itself (e.g., specific layers, training strategies, etc). Thus, we bypass the hard constraints by simply relying on the constrained structure of the neural network. A prominent example here is W-GAN solver \cite{Arjovsky2017WassersteinNetworks}. \rtypo{The potential $f_w$ in  Wasserstein-1 formulation (Eq.(7) in \cref{sec:gen_modeling})  has to satisfy a Lipschitz constraint. This is done by forcing compactness on the parameter space. In practice, this is achieved by weight clipping after each gradient update.}
Another way to import structure that ensures constraints feasibility is to use special layers.  What gives rise to this approach is the formulation in \eqref{eq:kantorovich_dual_2} which allows using Input Convex Neural Networks (ICNN) ~\cite{InputNetworks}. Specifically, one can rephrase \eqref{eq:kantorovich_dual_2} as an optimization problem over the space of convex functions while eliminating the distance constraints. This trick appears in~\cite{VardhanMakkuva2020OptimalNetworks,Vacher2021ConvexCriterion}, although elements of the convexification trick first appear in Villani's work \cite{VillaniTopicsTransportation}.
Reparameterizing the problem in terms of convex functions allows one to leverage a deep learning approach to solving OT problems. 
The formulation is given by
\rtypo{\begin{align*}
    C_{\mu, \nu} - \inf\limits_{\phi \in \mathsf{CVX(\mu)}} \Big[\mathbb{E}_\mu \phi(\vector{\rv{x}}) + \mathbb{E}_\nu \phi^\ast(\vector{\rv{y}}) \Big], \numberthis \label{eq:kantorovich_semidual}
\end{align*}}
where $C_{\mu, \nu} = \frac{1}{2}\Big(\mathbb{E}_{\mu}\Vert \vector{\rv{x}} \Vert_2^2 + \mathbb{E}_{\nu}\Vert \vector{\rv{y}} \Vert_2^2 \Big)$ and $\mathsf{CVX}(\mu)$ is the space of all convex functions integrable with respect to $\mu$. The space $\mathsf{CVX}(\mu)$ can be parameterized by input convex neural networks (ICNNs). Generally, one can compose ICNN from two principal blocks: linear blocks consisting of linear layers, and convexity preserving blocks consisting of linear layers with non-negative weights. The key idea here is to use special parametric models based on deep neural networks to approximate the set of convex functions in~\eqref{eq:kantorovich_semidual}. Note also that~\eqref{eq:kantorovich_semidual} involves only a single convex function $\phi$. ICNNs were used for computing OT maps \cite{VardhanMakkuva2020OptimalNetworks} and Wasserstein Barycenter \cite{Fan2020ScalableNetworks}. A number of variant architectures conforming to the same guidelines were also proposed, e.g., ConvICNN \cite{Korotin2019Wasserstein-2Networks} and DenseICNN \cite{DoBenchmark,Korotin2019Wasserstein-2Networks} were used in generative networks (\cref{sec:gen_modeling}).

 A question pertinent to grey-box solvers is \textit{how to differentiate through the OT computation?}  The OT plan $\pi^\ast:\mathbb{X} \times \mathbb{Y}$ that attains the infimum (\eqref{eq:kantorovich}) (or any discrete and/or regularized variant) is a function defined implicitly as a solution to an optimization problem. As such, given sufficient smoothness, functional operators such as differentiators for such an implicitly defined mapping can be examined. Given an iterative algorithm that estimates the optimal plan, one may estimate such derivatives using automatic differentiation by \emph{unrolling} the steps of the algorithm \cite{Monga2021AlgorithmProcessing} and applying the chain rule to compute derivatives through the algorithm. A popular example of this approach is unrolling the step of the Sinkhorn algorithm \cite{Adams2011RankingPropagation}. Clearly, this approach becomes more expensive as more iterations are required for the algorithm. 

Another approach is to use the \emph{implicit function theorem} \cite{Krantz2002TheApplications} to compute derivatives without any knowledge of the steps involved in the algorithm. This idea has been explored more generally for solutions to \rtypo{optimization} problems~\cite{gould2021deep} and fixed point equations~\cite{bai2019deep}. When used inside neural networks as layers, these are called implicit layers. Layers that solve \rtypo{optimization} problems such as OT problems, are sometimes called Deep Declarative Networks \cite{gould2021deep}.

A unified framework for implicit differentiation of Sinkhorn layers is presented in  \cite{EisenbergerADifferentiation}. They provide expressions for the implicit derivatives of loss functions that accept the result of the pre or post-composition of a neural network with a Sinkhorn layer. This allows for gradients to be backpropagated through the neural network pipeline so that variants of stochastic gradient descent can be used to update the neural network parameters. 
\rother{Such implicitly differentiated Sinkhorn layers have been used to solve the blind Perspective-n-Point problem~\cite{capmbell2020pnp} and in few-shot image classification ~\cite{ZhangDeepemd:Classifiers}}
Exploiting the problem structure with implicit differentiation can further increase computational gains significantly, as shown by \cite{Gould2022ExploitingStudies}. However, it requires case-by-case handling, lengthy mathematical derivations, and custom implementation. Modular implicit differentiation \cite{BlondelEfficientDifferentiation} presents a potential technology for handling such difficulty.

\textit{But, how good are the neural solvers ?}   
While a neural solver can result in computational savings for a pre-specified class of OT problems on which it was trained, neural solvers can fail to approximate OT solution \cite{StanczukWassersteinDistance} or exhibit poor generalization beyond the training dataset (see future directions \cref{sec:future_directions}).  The accuracy of neural solvers is a topic of recent investigation. Recent studies \cite{Korotin2019Wasserstein-2Networks} show that neural solvers might not faithfully recover the optimal map yet continue to perform well in downstream tasks. Architectural constraints (e.g., ICNN) also can be a concern. Despite architectural restrictions,  \cite{chenoptimal} demonstrated the richness of this class by showing its capability to accurately capture the temporal dynamics of complex physical systems. Yet, \cite{Richter-Powell2021InputNetworks} noted that ICNNs are difficult to train, which led \cite{bunnesupervised} to investigate a better initialization.  Beyond the approximation capability, the neural solvers can have characteristic variations. For some solvers, obtaining the OT plan/map is possible. For example,  from the optimal dual potential $\phi$ and $\psi$ in~\eqref{eq:regualized_Kantorovich_dual}, one can recover the optimal plan using the primal-dual relationships \cite{GenevayStochasticTransport}. For other solvers, obtaining the correspondence information (through a plan or a map) isn't feasible.

\textit{Relaxed Solver:}  Sometimes, the context in which OT computation is used provides clues for optimizing the computation. If the OT distance is not the ultimate target, then one can consider relaxing OT computations according to subsequent application requirements. Relaxed Word Movers Distance (RWMD) \cite{KusnerFromDistances} computes a lower bound of WMD  by relaxing one of the marginal constraints on the coupling. This reduces the running time complexity from cubic to quadratic in the number of data points. Linear-Complexity RWMD\cite{AtasuLinear-complexityAcceleration} further pushes it down to a linear complexity.

\subsection{Combined Approaches}
\label{sec:scalingot_combined}

While each of the previous strategies can bring benefits on the computational/statistical side, it can be noticed that there is always an associated side effect.  
Consider, for example, the entropy-based regularization (\cref{sec:scalingot_structure}). It produces an easy-to-parallelize algorithm and a well-behaved optimization problem (e.g., unique optimal coupling and a differentiable objective). Yet, it can produce a blurry optimal plan ill-suited for some applications \cite{Blondel2018SmoothTransport} or numerical instability issues depending on the magnitude of the regularization $\lambda$. 
Given this, a valid line of thought is combining multiple of the previously discussed (compatible) strategies above in a single framework and aggregating the benefits. Schmitzer \etal \cite{Schmitzer2019StabilizedProblems} combines ideas of  $\lambda$-scaling heuristic (\cref{sec:scalingot_cost_approx}), an adaptive truncation of the kernel, and the multi-scale coarse-to-fine scheme (\cref{sec:scalingot_paritition}) in a single algorithm to obtain the combined benefits.  \cite{Scetbon2021Low-RankFactorization} proposes structuring both the coupling (\cref{sec:scalingot_structure}) and the kernel matrix (\cref{sec:scalingot_cost_approx}) to be low rank. 
Similarly, \cite{Scetbon2021Linear-TimeCosts} used the same strategy to obtain a linear time complexity of Gromov Wasserstein. \marginpar{R3}\rref{Korotin \etal \cite{Korotin2022WassersteinEstimation} combines neural solver with the fixed point approach from \cite{alvarez2016fixed} to Wasserstein-2 barycenters of continuous measures.}

\section{Future Directions}
\label{sec:future_directions}

Circumventing OT computational challenges almost a decade ago \cite{CuturiSinkhornTransport} through optimization advancements has led to its emergence in machine learning. 
Subsequent advances \cite{GenevayStochasticTransport} in the same direction enabled embracing OT in many applications. 
Recently, with more adoption, computational limitations have become more pronounced in the big data era. 
OT is being used increasingly as a component of various contemporary learning pipelines \cite{NakagawaGromov-WassersteinAutoencoders,WuMotion-modulatedRecognition}. 
Algorithms with high complexity (e.g. super cubic) can not match scalability ambitions and would be rendered as computational bottlenecks. 
Notably, theoretical advances are no less important in this aspect. 
They pave the way for principled scaling rather than `hacking the way' to a better complexity.  
Below, we highlight, in broad strokes, some challenges we see as worthy of future research investment.

\textbf{Computational complexity:} The emphasis of this survey is on the longest-standing research challenge in OT, the computational budget. 
Despite the algorithmic advancements, more research is needed to develop computational approaches better suited for the massive amounts of data available. 
A recurring pattern observed in computational tricks in \cref{sec:scaling_ot}  is the presence of some sort of compromise. 
In general, the computational tricks involve a transformation of the original transportation problem into an easier \textit{approximate} version that potentially works better in a specific niche of problems. 
Unfortunately, this process typically incurs losing some connections and properties of the original formulation. 
Research in this direction will continue to flourish.

Research at the intersection of optimization and deep learning \cite{Shlezinger2022Model-BasedOptimization} presents one promising direction in this regard. 
The recent trend of {\em learned optimization} \cite{Chen2021LearningBenchmark} and {\em amortized optimization} \cite{Amos2022TutorialDomains} builds on the intuition that we can leverage the structure of multiple related optimization problems and learn a model to predict the solutions faster (without running the naive optimization from scratch). 
Evidently, this was shown to produce reliable results with orders of magnitude faster than classical optimization. 
Recently, some works have started exploring this direction\cite{Amos2022MetaTransport}\cite{Lacombe2021LearningBarycenters} in the OT literature. 
Interesting research questions in this direction are generalizing predicted solutions to out-of-(training data) distribution and, more importantly, addressing the lack of theoretical convergence guarantees in this setup. 
Formulations that can benefit the most from these efforts are the complex (higher order) formulations such as multi-marginal OT,  Gromov Wasserstein, and hierarchical formulations. 
To date, leveraging these formulations in learning tasks is subject to careful judgment that weighs the potential benefits against the computational price. 

\textbf{\textbf{Benchmarking.}} Given the wealth of OT formulations and solvers, standardized evaluation of solvers would greatly benefit future research.  
Yet, very few works \cite{DongATransport, schrieber2016dotmark, DoBenchmark} have addressed OT solver benchmarking in the literature. 
These works are mostly \cite{DongATransport, DoBenchmark} focused on discrete OT solvers with low-dimensional inputs that don't reflect the current nature of big datasets.
Benchmarking OT solvers is challenged by the diversity of the OT landscape in terms of formulations and machine learning applications considered. 
One potential direction is to initiate specialized benchmarks that focus on specific tasks in which OT is used extensively (e.g.  domain adaptation \cref{sec:domain_adaptation} ) or specific group of formulations. 
Another benchmarking challenge is designing metrics that assess the OT computational accuracy. 
This is difficult given that in some cases (such as OT between continuous measures\cite{Korotin2022KantorovichTransport}) it isn't straightforward to obtain the ground truth OT solution including the cost and plan. 
Recent efforts \cite{Korotin2022KantorovichTransport} address this issue for special cases. 
It would be interesting to see more research in this direction. 
Another issue with the metric design is that the performance on the downstream task isn't an indicator of a good approximation of the Wasserstein distance \cite{StanczukWassersteinDistance}.

\textit{\textbf{New data types.}}
Extending the scalability of OT beyond data quantity to new data types is a promising future direction. \marginpar{R2}\rref{There are recent efforts in this direction such as applying OT on functional data \cite{zhu2021functional}}
\marginpar{R1}\rref{Complex data such as structured objects (e.g. graphs\cite{petric2019got,Chen2020OptimalGraphs} and trees) and sequences and temporal streams \cite{CohenAligningSpaces}} have received little attention in the literature. OT on structured objects requires generalizing the notion of transport to accommodate both nodes and relations while ensuring other properties. 
Current approaches imitate OT on these objects either by having \rother{multiple levels of computation \cite{Brogat-MotteLearningBarycenters}} 
or embedding the data in a new space that preserves the structure (e.g. hyperbolic embedding \cite{Alvarez-MelisUnsupervisedSpaces}). 
Both approaches come with increased computational challenges. 
A native formulation that embraces the structure is yet to be realized. 
Potential advantages of native formulations include having the structure reflected in marginal constraints and the capacity to perform interpolation and Barycentric mapping on structured data.

\textbf{\textit{Novel applications and better tools.}}
By treating the weights of deep models as probability measures, one can cast the framework of OT on problems involving transportation among (trained) model objects. 
This can be useful, for example, in model fusion and interpolation without requiring re-training. 
For example, OT can be used \cite{PalSinghModelTransport} for fusing pre-trained neural networks as an easy 'one-shot' transfer mechanism. 
Another use of this can be in {\em federated learning} \cite{Li2020FederatedDirections}, in which multiple clients collaborate to train a global model in a coordinated server without exchanging their local data. 
Traditional clients model aggregation (i.e. averaging) on the coordinated server can be replaced with OT-based advanced averaging (e.g. Wasserstein barycenter) that can address clients heterogeneity \cite{RakotomamonjyPersonalisedSpaces}. 
Along the same line, OT can be used as a geometry-aware analytics tool for studying the weights of  deep models and addressing issues of interpretability.

A number of powerful OT tools such as POT\cite{Flamary2021POT:Transport} and OTT \cite{CuturiOptimalWasserstein} exist.  Tools like OTT and KeOps \cite{FeydyFastMatrices} are leveraging GPUs and TPUs.
KeOps can handle a very large number of samples by storing computation as formulas and streaming them on the fly. 
Yet, it is difficult to extend the package and use it and deep learning \cite{Fatras2021UnbalancedAdaptation}. Looking forward, overcoming the limitations of current tools will drive wider applicability of optimal transport.

\section{Conclusion}
Optimal Transport provides a coherent mathematical toolbox for formulating  and solving problems whose core is about similarity quantification and correspondence estimation between objects. Problems like these, which arise naturally in machine learning, are addressed by transporting collections of samples or features in a geometric sensible way. We discussed various transportation mechanisms (i.e. formulations), their characteristics, their application value, and their limitations. Then, a formulation-agnostic taxonomy for scaling OT computation to large datasets was presented. We reviewed a broad spectrum of techniques ranging from those native to the optimization framework (e.g. additional constraints or structures) all the way to those inspired by recent advances in machine learning (e.g. neural learned solvers).

\appendices

\section{Extended Background }
\label{sec:ot_background}
\subsection{Kantorovich formulations}

\rtypo{Recall from the main paper the Kantorovich formulation of the OT problem,}
\rtypo{\begin{align*}
    \mathcal{K}_c(\mu, \nu) = 
 & \inf\limits_{\pi \in \Pi(\mu, \nu)} \int_{\mathbb{X} \times \mathbb{Y}}  c(\vector{x},\vector{y}) d\pi(\vector{x},\vector{y}) \\
 =& \inf\limits_{\pi \in \Pi(\mu, \nu)} \mathbb{E}_{(\vector{\rv{x}},\vector{\rv{y}}) \sim \pi} \ [c(\vector{\rv{x}},\vector{\rv{y}})].
 \numberthis 
\label{eq:kantorovich_app}
\end{align*}}
The benefit of the Kantorovich formulation in~\eqref{eq:kantorovich_app} is that multiple computational approaches and interpretations of the same problem can be obtained. The remainder of this section is devoted to discussing some of these reformulations and interpretations.

We can invoke the basic concept of ``duality'' \footnote{Any convex minimization problem admits a \textit{dual} maximization problem whose optimal value lower-bounds that of the minimization problem. The optimal values coincide if the problem is a linear program. } from convex optimization to obtain a \textbf{$\sup$} (dual) formulation equivalent to the \textbf{$\inf$} (primal) formulation in \eqref{eq:kantorovich_app}. The Kantorovich dual of~\eqref{eq:kantorovich_app} \cite[Theorem 5.10]{Berlin2009OptimalNew} is :
\rtypo{\begin{align*}
    \mathcal{K}_c(\mu, \nu) &= \sup\limits_{(\varphi, \psi) \in \Phi_c} \Big[ \int_{\mathbb{X}} \varphi(\vector{x}) d\mu(\vector{x})  + \int_{\mathbb{Y}} \psi(\vector{y}) d \nu(\vector{y}) \Big] \\
    &= \sup\limits_{(\varphi, \psi) \in \Phi_c} \mathbb{E}_{\vector{\rv{x}} \sim \mu} \big[ \varphi(\vector{\rv{x}}) \big] + \mathbb{E}_{\vector{\rv{y}} \sim \nu} \big[ \psi (\vector{\rv{y}}) \big], \numberthis
\label{eq:kantorovich_dual_app}
\end{align*}}
where \rtypo{$$ \Phi_c = \{ (\varphi, \psi) \in L^1(\mu) \times L^1(\nu) \mid \varphi(\vector{x}) + \psi(\vector{y}) \leq c(\vector{x},\vector{y})\}.$$}
The dual is obtained by noting that~\eqref{eq:kantorovich_app} is a linear programming problem. For more on linear programs, dual problems, and the Kantorovich problem, refer to the appendix section on optimization (\cref{sec:app_optimization}).

Given that~\eqref{eq:kantorovich_dual_app} bears little resemblance to~\eqref{eq:kantorovich_app}, one might wonder how intuitively the original minimum-effort transportation can be interpreted as a maximization problem. 
This can be understood by imagining that the transportation is carried out by a third-party beneficiary. 
Their goal then would be to maximize the total sending \rtypo{$\int_{\mathbb{X}} \varphi(\vector{x}) d\mu(\vector{x})$ and receiving $\int_{\mathbb{Y}} \psi(\vector{y}) d \nu(\vector{y})$} costs as long their rate stays below a baseline cost \rtypo{$c(\vector{x}_i,\vector{y}_j)$} for all the points $\vector{x}_i$ and $\vector{y}_j$. Otherwise, the interested parties would opt out (assuming a competitive market climate).

Note also that in~\eqref{eq:kantorovich_dual_app}, we no longer have the optimal plan $\pi$ but rather the variables $\varphi(\vector{x})$ and $\psi(\vector{y})$, known as the \textit{dual potentials}. They still encode the optimal plan and the optimal plan can be recovered from them. However, the dual problem~\eqref{eq:kantorovich_dual_app} can be more computationally tractable than the primal problem~\eqref{eq:kantorovich_app}. In the discrete problem in \fixedref{\begin{math}\mathbf{\S 2}\end{math}},
the $n \times m$ memory footprint of the optimal plan is reduced to a linear (w.r.t the input) footprint required by the dual potential. This can be very useful when scaling the computation to thousands or even millions of points. On the other hand, we still have $n \times m$ constraints to check against when solving~\eqref{eq:kantorovich_dual_app}.

A simplification that makes the objective dependent only on one potential $\varphi$ (or $\psi$) instead of two potentials $\varphi$ and $\psi$ is called the \textit{semi-dual}:


\rtypo{\begin{align*}
    \mathcal{K}_m(\mu, \nu) &= \sup\limits_{\varphi \in L^1(\mu), \varphi \in L^1(\nu)} \Big[ \int_{\mathbb{X}} \varphi(\vector{x}) d\mu(\vector{x})  + \int_{\mathbb{Y}} \varphi^{c}(\vector{y}) d \nu(\vector{y}) \Big] \\
    &= \sup\limits_{\varphi \in L^1(\mu), \varphi \in L^1(\nu)} \mathbb{E}_{\vector{\rv{x}} \sim \mu} \big[ \varphi(\vector{\rv{x}}) \big] + \mathbb{E}_{\vector{\rv{y}} \sim \nu} \big[ \varphi^{c}(\vector{\rv{y}}) \big], \numberthis
\label{eq:kantorovich_semi_dual_app}
\end{align*}}
where \rtypo{$\psi(\vector{y})$ was substituted with $\varphi^{c}(\vector{y}) = \inf\limits_{\vector{x}} [ c(\vector{x},\vector{y}) - \varphi(\vector{x})]$ which is the c-transform of $\varphi$.} The term \textit{semi-dual} was named as such by \cite{Cuturi2018Semi-dualTransport} since this formulation underlies semi-discrete OT methods. Notably, the flexibility by which this formulation allows us to pick one potential over the other can be useful in applications. In some problems (e.g. crowd counting \cite{Ma2021LearningTransport}) that require transporting a sparse measure (e.g. a few annotation points) to a dense one (e.g. many image pixels), one can eliminate the dense potential and optimize over the other. 

\rtypo{If additionally the cost function $c(\vector{x}, \vector{y})$ is a distance function $d(\vector{x}, \vector{y})$ satisfying the triangle inequality, e.g. $c(\vector{x}, \vector{y}) = d(\vector{x}, \vector{y}) = \Vert\vector{x} - \vector{y}\Vert_2$, the c-transform becomes $\varphi^{c}(\vector{y}) = -\varphi(\vector{y})$ where $\varphi(\vector{y})$} is a 1-Lipschitz function w.r.t the distance $d$:
\rtypo{\begin{align*}
    |\varphi(\vector{\rv{x}}) - \varphi(\vector{\rv{y}})| \leq d(\vector{\rv{x}},\vector{\rv{y}}) \quad \forall (\vector{\rv{x}},\vector{\rv{y}})  
\end{align*}}
We denote the set of 1-Lipschitz functions w.r.t the distance $d$ by $L_d(1)$. So the semi-dual formulation \cite[Theorem 5.10]{Berlin2009OptimalNew} becomes:
\rtypo{\begin{align*}
    \mathcal{K}_c(\mu, \nu) &= \sup\limits_{\varphi \in L_d(1)} \Big[ \int_{\mathbb{X}} \varphi(\vector{x}) d\mu(\vector{x}) - \int_{\mathbb{Y}} \varphi(\vector{y}) d \nu(\vector{y}) \Big] \\
    &= \sup\limits_{\varphi \in L_d(1)} \mathbb{E}_{\vector{\rv{x}} \sim \mu} \big[ \varphi(\vector{\rv{x}}) \big] - \mathbb{E}_{\vector{\rv{y}} \sim \nu} \big[ \varphi(\vector{\rv{y}}) \big], \numberthis
\label{eq:kantorovich_rubinstein}
\end{align*}}


This is known as the \textit{Kantorovich-Rubinstein formulation} or the \textit{Kantorovich-Rubinstein norm} \cite{hanin1992kantorovich}.  
Note that this only applies to \rtypo{$c(\vector{\rv{x}},\vector{\rv{y}}) = d(\vector{\rv{x}},\vector{\rv{y}})$, not to the more general (and popular) case $c(\vector{\rv{x}},\vector{\rv{y}}) = d(\vector{\rv{x}},\vector{\rv{y}})^p$ for some power $p$}.

 The exact formulations discussed here are typically solved using classic linear programming solvers. Unfortunately, these methods can't exploit the parallel computing power of modern hardware. As we will see later, approximate formulations that embrace parallel computation and leverage GPUs (e.g. Regularized Optimal Transport \fixedref{\begin{math}\mathbf{\S 4.1}\end{math}}
 ) had gained more popularity in many recent applications.

\subsection{Hierarchical OT Formulation}
\label{sec:hierarchical_ot}

\rother{\textit{Motivation:} Hierarchical Optimal Transport (HOT) extends the concept of OT by introducing a hierarchical structure to the problem. In HOT, the transportation problem is considered at multiple scales or levels of granularity. By grouping data points with similar features or predefined structures, the transport task can be handled at the group level before dealing with individual data points. This approach is particularly useful when dealing with complex data structures with inherent hierarchical relationships, such as in certain types of data analysis and machine learning tasks.}

\rother{\textit{Formulation and Characteristics:} There are several techniques known as HOT. It is an OT problem where the ground metric itself is defined through an OT problem. Generally, these \cite{LeeHierarchicalAlignment,Alvarez-Melis2020GeometricTransport} involve a two-tiered approach: solving smaller, local assignments within each group and then addressing a larger, overarching assignment across groups. 
This layered approach simplifies the overall task and allows for flexibility in methods used at each level.}

\rother{\textit{Applications}  Problems involving hierarchical data, like matching documents, aligning datasets, and adapting to different domains, can benefit from using hierarchical optimal transport (OT) methods. One example is HOTT \cite{YurochkinHierarchicalRepresentation}, which uses the natural structure of documents (documents are modeled as distributions over topics that are modeled as distributions over words) to make comparing documents using OT faster.  Similarly, HiWA\cite{LeeHierarchicalAlignment} approaches domain adaptation in the multi-modal setup as a hierarchical transportation process. A notable development here is the introduction of a scalable ADMM algorithm, which can be divided across cluster pairs and allows for parallel computations. This two-level transportation approach has been applied to other areas such as unsupervised domain adaptation \cite{ElHamri2022HierarchicalAdaptation}, semi-supervised domain adaptation \cite{Taherkhani2020TransportingLearning} and measuring dataset distances \cite{Alvarez-Melis2020GeometricTransport}.}

\section{Applications of OT in Machine Learning}
\label{sec:ot_applictions}

In this section, we review the applications of OT. We focus primarily on domain adaptation and generative models.


\subsection{Domain Adaptation (DA)}
\label{sec:domain_adaptation}

\begin{figure*}[!t]
    \centering
    \includegraphics[width=1\linewidth]{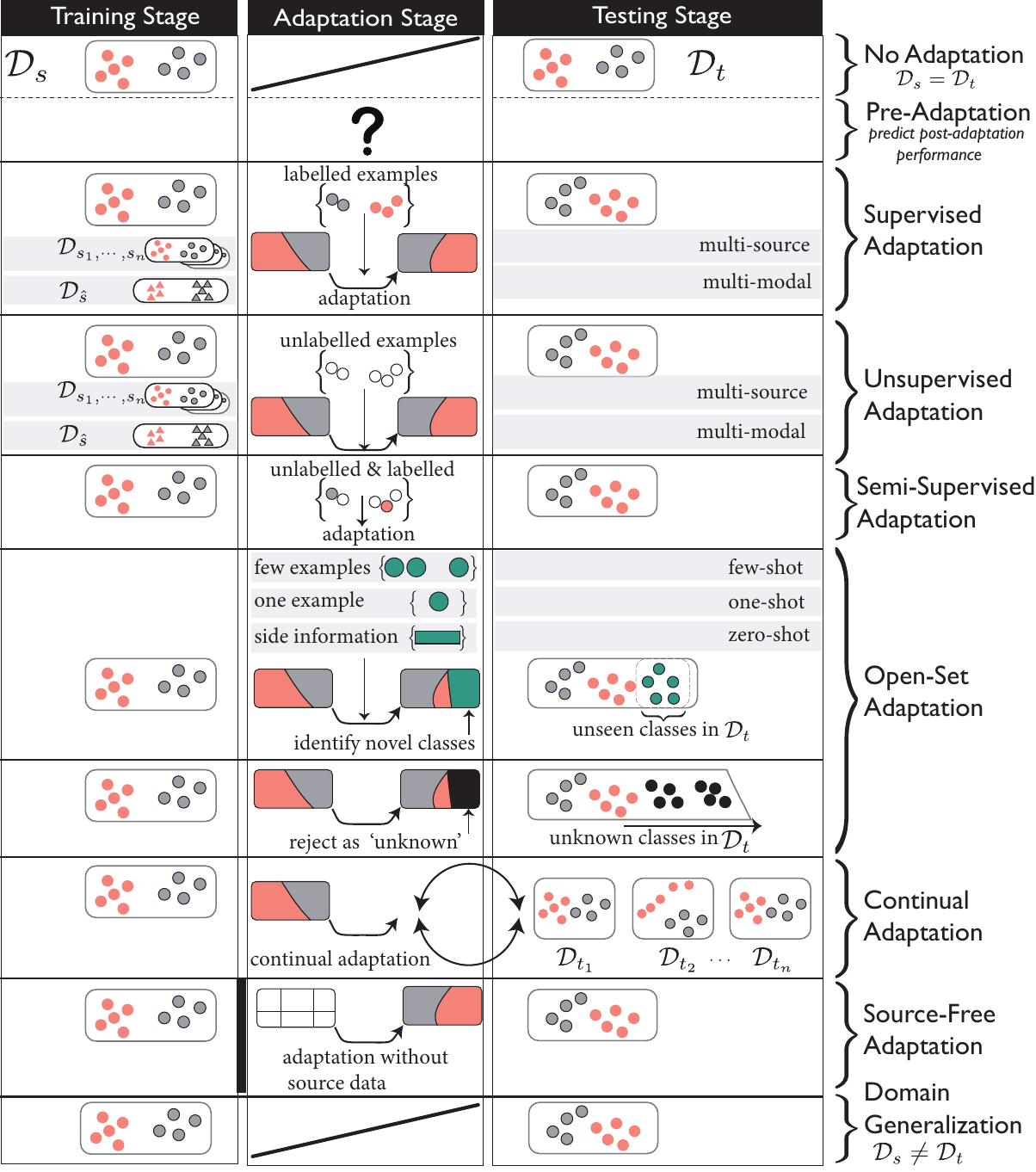}
    \caption{\textbf{OT Domain Adaptation.} Various domain adaptation configurations addressed by OT works in \cref{sec:domain_adaptation}}
    \label{fig:da_flavors}
    
\end{figure*}

Domain adaptation is a very vast and vibrant sub-field of machine learning. Its motivation is the frequent violation of a  classical learning assumption in the real-world application, namely, the assumption that the data used for model training and testing data come from the same probability distributions. A common example of this assumption failure is an image classification task in the acquisition conditions causing a non-negligible (distributional) shift between training and test datasets \cite{Saenko2010AdaptingDomains}. In this vast topic, one can find these excellent recent surveys focusing on specific themes of domain \rtypo{adaptation} whether it is generalization \cite{WangGeneralizingGeneralization}, open set adaptation\cite{YangGeneralizedSurvey} or theoretical advances \cite{RedkoAGuarantees}. Here, we rather collect the many domain \rtypo{adaptation} paradigms to which optimal transport contributes. It should be noted that the subtle but crucial differences between these domain \rtypo{adaptation} paradigms can be confusing. Figure~\ref{fig:da_flavors} provides a quick visual connection between various domain \rtypo{adaptation} 
paradigms and highlights the differences.

Formally, domain \rtypo{adaptation} considers two domains \textit{source domain} and \textit{target domain} denoted as $s$ and $t$; respectively. The goal is to learn a robust model given that the source and target domains have different data distributions (i.e. $\mathcal{D}_{s} \neq \mathcal{D}_{t}$). In the most common setup, the labels for the two domains are assumed to be available in abundance. In this case, it is called \textit{supervised \rtypo{adaptation}}. Yet, the problem also comes in many other variations.  For example, the target data can be unlabeled (\textit{unsupervised \rtypo{adaptation}}) or partly labeled (\textit{semi-supervised \rtypo{adaptation}}). The classes in the source and target domain can be disjoint (\textit{open set \rtypo{adaptation}}). The learning goal, in this case, may go beyond merely rejecting new classes as `unknown' to include learning the new classes from one (\textit{one-shot \rtypo{adaptation}}) or multiple (\textit{few-shot \rtypo{adaptation}}) samples. When the target domain is very dynamic, domain \rtypo{adaptation} can be done \rtypo{continuously} (\textit{continuous domain \rtypo{adaptation}}). All the previous configurations assume that the training samples from the source domain are available during adaptation. This is unrealistic in cases where it's impractical to take training data to deployment due to privacy or storage considerations. This motivates \textit{source-free domain \rtypo{adaptation}}, which makes different assumptions as shown in Figure \ref{fig:da_flavors}.

Since the first use of OT in domain \rtypo{adaptation} problems by \cite{Courty2017OptimalAdaptation}, OT has been considered a strong baseline for domain adaptation. The mathematical flexibility of optimal transport formulations allows for catering to the various domain \rtypo{adaptation} configurations. Before proceeding with the review of these works, it is instructive to see why one might consider OT in domain \rtypo{adaptation} problems. The following are potential motivations:

\begin{itemize}
    \item Distributions matching is at the core of domain adaptation techniques.  Given that, the problem naturally lends itself to optimal transport. Moreover, the previously discussed OT properties (\rtypo{e.g.,} geometry awareness and diversity of formulations) can offer a new perspective on the problem.  
    \item As a metric on the space of probability measures, the Wasserstein distance was used to derive theoretical bounds \footnote{For the interested reader, Redko \etal \cite{RedkoAGuarantees} nice exposition of theoretical domain adaptation covers learning bounds based on various statistical frameworks and divergences (Wasserstein included).} on the domain adaptation error\cite{Redko2017TheoreticalTransport} and deep learning generalization \cite{ChuangMeasuringTransport}. Thus, many techniques building on OT for \rtypo{adaptation} 
    \cite{RostamiLifelongDistribution} provide \rtypo{adaptation} theoretical guarantees.
    \item It is possible to encode a prior structure useful for domain \rtypo{adaptation} (\rtypo{e.g.,} preserving labels or graphical relations between samples \cite{Courty2017OptimalAdaptation}) in the OT mapping between the source and target distributions.
    \item A degree of interpretability can be gained by having explicit correspondences, represented in a transport plan, between domains/datasets. 
    This might be even more true in works that consider the matching in the raw input space \cite{Alvarez-Melis2020GeometricTransport,Redko2020CO-OptimalTransport} rather than (a less interpretable) feature space.
\end{itemize}

The \rtypo{above factors} are possible reasons for OT to be a versatile building block in domain \rtypo{adaptation} areas. Below is a selective review of these areas.

\textbf{\textit{Transferability Assessment using OT.}} Transfer learning across domains or tasks is one the most promising and widely adopted machine learning methods and it remains to be a topic of extensive investigation \cite{ZhuangALearning}. Yet, judging the effectiveness of a specific transfer learning setting prior to the actual transfer in a quantifiable way has received much less attention. In multi-task learning, one may want to save the joint training burden of weakly related tasks as the final performance will be actually worse compared to \rtypo{single-task training}. Also, this can be used in source model selection \cite{ScalableLearning} (\rtypo{i.e.,} select from a pool of source pre-trained models the best one for a target task). Rather than relying on human experience for this kind of relatedness assessment, estimating a transferability metric \cite{TanOTCE:Representations,TranTransferabilityTasks} could be much more ubiquitous and practical.
General metrics for transferability can be categorized into the \textit{analytical transferability metrics} \cite{TranTransferabilityTasks, BaoAnLearning, Nguyen2020LEEP:Representations} and \textit{empirical transferability} \cite{ZamirTaskonomy:Learning}. The first has stricter assumptions but is computationally efficient and enjoys theoretical generalization bounds. The latter is scalable but lacks theoretical guarantees. 

Optimal Transport-based Conditional Entropy OTCE \cite{TanOTCE:Representations} was the first analytical transferability metric in a simultaneous cross-domain cross-task. It adopts the popular “retrain head” transfer model in which a source model with \rtypo{a frozen} feature extractor is transferred to target data by fine-tuning only the classifier head following the extractor. The proposed metric predicts the model’s performance after the tuning without actual re-training.

\textbf{\textit{Unsupervised DA (UDA) using OT.}} Several works applied OT in UDA in which labels are available only for the source while samples are available from both the source and target domain. Courty \etal \cite{Courty2017OptimalAdaptation} established a framework that was extended by others later for using optimal transport as a principled method for UDA. The key idea is to unify the training and test samples in one domain in which the classifier can be learned. First, OT \rtypo{aligns} the empirical measures of source and target domains. A process that can be done without using any labels. Then, \rtypo{the training samples are transported to the target distribution using the estimated transport plan}. Finally, a classifier can be learned only on the target domain. A graph-inspired regularization is used to preserve the after-transportation proximity of the source samples sharing the same label.

Instead of sample alignment, follow-up works seek to align the model's internal features corresponding to$\mathcal{D}_s$ and $\mathcal{D}_t$. Sliced Wasserstein Discrepancy (SWD) \cite{LeeSlicedAdaptation} builds on Maximum Classifier Discrepancy (MCD) \cite{SaitoMaximumAdaptation} 
that maximizes the discrepancy between task-specific classifiers as a part of an adversarial \rtypo{adaptation} strategy.
As the discrepancy is a key component of MCD, SWD proposes upgrading it with the Sliced Wasserstein distance.  Interestingly, by merely replacing the $L_1$  discrepancy in MCD with SWD, the system consistently delivers better performance 
in image classifications, segmentation, and object detection tasks.

Enhanced Transport Distance (ETD) \cite{TanOTCE:Representations} uses a neural solver 
(\fixedref{\begin{math}\mathbf{\S 5.5}\end{math}}) for OT-based features alignment and notes the mini-batch instability issue. Specifically, optimizing OT on mini-batch can lead to inconsistent transport plans across iterations. To mitigate the bias caused by mini-batch,
the system weighs (calibrates) the ground cost using a network-integrated 
attention.

 Wasserstein Guided Representation Learning (WDGRL) \cite{Shen2017WassersteinAdaptation} proposes augmenting  Adversarial Domain Adaptation (ADA) architectures, such as DANN \cite{Ganin2016Domain-adversarialNetworks}, with additional Wasserstein loss to guide the domain \rtypo{adaptation} process. ADA extends the concept of Generative Adversarial Networks (GAN) to domain adaptation. While GANs leverage the adversary to generate samples indistinguishable from the real data, ADA seeks a mapping that makes source features indistinguishable from target domain features. WDGRL adds a WGAN-like domain critic (WGAN discussed in \cref{sec:gen_modeling}) for calculating the 1-Wasserstein distance between the source and target features. The motivation behind this is the possibility of having vanishing gradients when the supports of the mapped features are disjoint, a setting in which OT is assumed to work better compared to other divergences. In fact, WDGRL significantly outperforms DAN in this setting, albeit \rtypo{the results are based} on synthetic data. 

\textbf{\textit{Unsupervised DA using OT (multi-source).}} JCPOT \cite{Redko2019OptimalShift} considers the mutli-source DA problem with a focus on cases where the label proportions between source and target domains are different. Target shift (or class imbalance) has practical values in applications like anomaly detection. Theoretically \cite{Mansour2009DomainAlgorithms}\rtypo{,} the error on the target domain can be minimized by re-weighting the distribution of the source classes to match the target one. Motivated by the impracticality of acquiring the target class distribution, an optimal transport matching that automatically acquires the weights is proposed. Entropy \rtypo{regularized} OT (\fixedref{\begin{math}\mathbf{\S 4.1}\end{math}})
is used for mapping between every single source and the target, and the problem is formulated as a Bregman projection that has a simple solution (proposition 1 \cite{Redko2019OptimalShift}). The approach shows improvement over vanilla OT on the satellite imagery dataset.

\begin{figure}
    \centering
    \includegraphics[width = 0.95\linewidth]{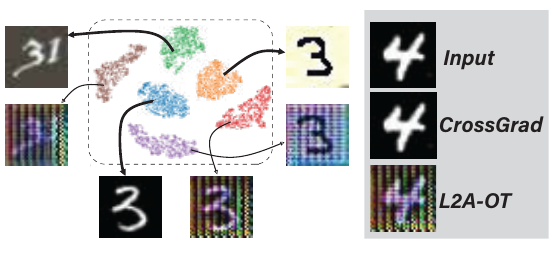}
    \caption{(Left) L2A-OT augments source domains (thick arrows) with synthesized novel domains (thin arrows) by learning to maximize the Wasserstein distance between the two sets. (Right) synthesized novel samples are qualitatively better compared to adversarial perturbation approach CrossGrad \cite{Shankar2018GeneralizingTraining} (figures adapted  from \cite{Zhou2020LearningGeneralization})}
    \label{fig:l2aot}
\end{figure}

\textbf{\textit{Open Set DA using OT.}} In open set domain \rtypo{adaptation}, the target domain can contain classes that weren't present in the source domain. While more practical, it is a challenge where domain \rtypo{adaptation} methods fail due to interference with extra unknown classes. The key goal in open set domain \rtypo{adaptation} \cite{BustoOpenAdaptation} (OSDA) is to correctly classify known classes while rejecting the extra classes as ``unknown''. Xu \etal \cite{Xu2020JointAdaptation} noticed that aligning the whole source and target domains followed by previous OT methods is actually inferior to the adaptation in an open set setting. Causing a problem known as \textit{negative transfer} \cite{Pan2010ALearning}.  Thus,  Joint Partial Optimal Transport (JPOT) \cite{Xu2020JointAdaptation} employs partial optimal transport to align only ``well-matched''/most-correlated samples and avoid far-fetched pairings that can cause the negative transfer. To obtain the partial matching, the mean cost of the complete optimal transport matrix \cite{Damodaran2018DeepJDOT:Adaptation} is used as a threshold. Coupled pairs whose distance is greater than this threshold are marked as non-transferable.

\textbf{\textit{Open Set DA using OT (zero-shot).}} For the more challenging cases of totally unseen classes (\rtypo{i.e.,} zero shot),  GZSL\cite{WangZero-ShotTransport} proposes synthesizing features for the unseen classes from the corresponding auxiliary textual descriptions. The feature generator is encouraged to match real feature distributions by minimizing regularized OT distance. OT-based matching was motivated by the concern of sampling outliers (brought by domain shift) that can make point-wise matching approaches less robust. IPOT \cite{Xie2019ADistance} is used instead of Sinkhorn as it maintains near-linear time complexity while being less sensitive to regularization weight.

\textbf{\textit{Multi-modal and heterogeneous DA using OT.}} Applications that require models to operate across different modalities motivate the need for heterogenous domain \rtypo{adaptation}. 
Clearly, aligning features living in heterogeneous (multimodal) spaces is more complicated than the alignment of homogeneous (unimodal) features.
In a neural population decoding application, HiWA \cite{LeeHierarchicalAlignment} used data from populations of (human brain) neurons to predict the arm movement direction in reach-out tasks. This involves aligning two domains of different modalities; the neural data distribution and the 3D movement data distribution. Given the difficulty of the \rtypo{adaptation}, they suggest leveraging the additional clustering structure of the data to regularize OT and constrain the solution space. Thus, a nested formulation of OT is proposed for aligning clustered and multi sub-spaces data and the framework hierarchical optimal transport (\fixedref{\begin{math}\mathbf{\S 4.6}\end{math}})
was invoked.

 Knowledge Distillation (KD) is a common option for cross-modal domain \rtypo{adaptation}. That is; aligning latent features of two models, each of them trained on a different modality yet solving the same or related tasks. Wasserstein Contrastive Knowledge Distillation (WCKD) \cite{ChenWassersteinDistillation}  encourages a student (compact) model features $h^s$ to be distributionally similar to those of an advanced teacher $h^t$. 
 Specifically, a GAN-like discriminator ensures that source samples are not distinguishable from target samples.  The alignment is done using a neural version of OT (WGAN\cite{Arjovsky2017WassersteinNetworks}).

\textbf{\textit{Continual DA using OT.}} Very few works developed techniques that address gradual distributional shifts in the target domain. Continual Domain \rtypo{Adaptation} \cite{RostamiLifelongDistribution}  is a problem that resembles Continual Learning (CL)\cite{SlabaughATasks} albeit without assuming the availability of labels in the target domain.  LDAuCID \cite{RostamiLifelongDistribution} learns a source model capable of adapting to several sequential target domains without labels. A strong overlap between the source and target classes is assumed. LDAuCID aligns the \textit{consolidated} internal learned distribution with the target \rtypo{domain using} Sliced Wasserstein. To mitigate the catastrophic forgetting issue, representative samples from all tasks are stored and replayed in the adaptation stage. This technique is known as \textit{experience reply} in CL literature.

\begin{figure}[t!]
    \centering
    \includegraphics[width = 1\linewidth]{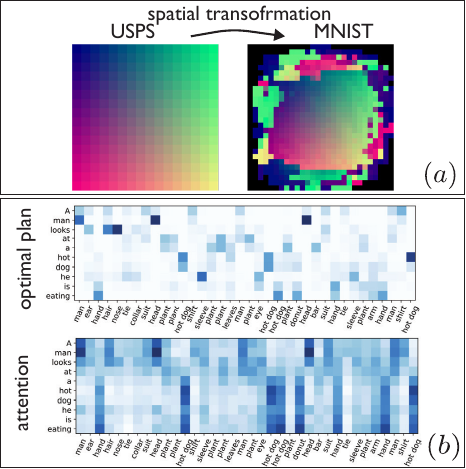}
    \caption{\textbf{Interpretability of optimal transport plan}. Optimal plan is used to reveal in (a) features (represented by raw pixels) correspondences between USPS and MNIST datasets using an unsupervised approach \cite{Redko2020CO-OptimalTransport} and (b) images object-to-word correspondences \cite{YuanWeaklyTransport}. The latter was shown to be favorable from an interpretability perspective compared to Transformer's attention due to sparsity. The figures were adapted from original references}
    \label{fig:interp_plan}
\end{figure}

\textbf{\textit{Source-free DA using OT.}} Traditional UDA performs the \rtypo{adaptation} between domains using the source and target data. Requiring the presence of source data during \rtypo{adaptation} is impractical in cases when the data is inaccessible due to privacy (e.g., Federated Learning clients \cite{LiAProtection}), storage limitations (e.g., on edge device), or other reasons. Augmented Self-Labelling (ASL) \cite{YanAugmentedAdaptation} uses the pre-trained source model (instead of the source data) and the unlabeled target data for \rtypo{adaptation}. ASL generates pseudo-labels for target data using the pre-trained model to be used during the model adaptation. It builds on a technique \cite{AsanoSelf-labellingLearning}  that frames \textit{self-labeling} as an optimal assignment between the model's predictions (soft-labels) and the set of possible labels under an equipartition constraint (that avoids degenerate solution where a single label is assigned to all samples).

\textbf{\textit{Domain generalization using OT.}} UDA assumes that data can be collected in advance (prior to training ) from the target domain since this is required for training. This assumption doesn't hold in practice. Domain Generalization (DG) thus is envisioned as a more practical alternative that learns a model capable of operating in unseen target domains without requiring model update (adaptation). Learning to Augment by Optimal Transport (L2A-OT) \cite{Zhou2020LearningGeneralization} proposes learning to synthesize novel domains from multiple source domains. By augmenting source domains with synthesized ones, DG can be realized through increased training data diversity. The problem is framed as image-to-image \rtypo{translation despite} being similar in spirit to gradient-based perturbation\cite{Shankar2018GeneralizingTraining}(i.e., perturbing input using adversarial gradients from a domain classifier).
Interestingly, the conditional generator in their pipeline is trained to maximize (not minimize) the optimal transport distance between the source domain and the synthesized novel domains. Such divergence loss is balanced with cycle consistency loss \cite{ZhuUnpairedPhotos} and task loss to ensure preserving semantic content. L2A-OT advocates \rtypo{using} Wasserstein distance compared to f-divergences for divergence maximization as the near-zero denominators for f-divergences tend to generate large but numerically unstable divergence values.

\textbf{\textit{DA interpretability  using OT.}}  One of the virtues of OT-based domain \rtypo{adaptation} is the possibility of inspecting the optimal plan (when available) to gain insights into the domains' correspondence. This offers an \rtypo{opportunity beyond} merely detecting and handling the distributional shift between the source and target domains. Specifically, an explanation of the domain shift between the source and target datasets through the samples or features correspondences
can be attained.  

We note that many OT works \cite{Redko2020CO-OptimalTransport,Kulinski2022TowardsShifts,KulinskiTowardsShifts,Lin2021MakingPoints,Blondel2018SmoothTransport, YuanWeaklyTransport} refer to \textit{interpretability} of the optimal plan without formal specification of what are the interpretability requirements. Yet, there is consensus among them that some characteristics are desirable and favorable from an interpretability point of view. For example,  sparse optimal plans \cite{Blondel2018SmoothTransport, Kulinski2022TowardsShifts} are favorable. Sparsity is motivated by the principle of \rtypo{parsimony,} in which simple solutions should be preferred. For example, a sparse optimal plan in a weakly-supervised \rtypo{adaptation} setup  \cite{YuanWeaklyTransport} between \rtypo{image-sentence} pairs can be easier to interpret compared to the dense attention from Transformers. As shown in Fig.~\ref{fig:interp_plan}(b), the two constructs (the optimal plan and the attention) visualize the strength of association between words (e.g., `man') on the vertical axis and visual objects, denoted by words on the horizontal axis. Clearly, having the word strongly associated with a few visual objects (e.g., the `man' in the optimal plan) is easier to interpret.  Note, however, that solving OT can yield an arbitrarily complex plan that's not necessarily interpretable \rtypo{per the abovementioned characteristics.} Thus, a number of works explicitly place additional constraints on the optimal plan reflecting the interpretability requirements. For example, \cite{Kulinski2022TowardsShifts,KulinskiTowardsShifts} suggest enforcing user-defined interpretable mappings (e.g., k-sparsity) on the optimal plan as additional constraints.

Also, it is easier to interpret the correspondences when they can be traced back to the raw samples \cite{Redko2020CO-OptimalTransport,Alvarez-Melis2020GeometricTransport} . For example, CO-Optimal Transport \cite{Redko2020CO-OptimalTransport} simultaneously estimates the correspondence between samples (images) and features (independent pixels values across the dataset images) from two handwritten digit datasets. The features of \rtypo{the transport plan} (Fig.~\ref{fig:interp_plan}(a) shows the spatial transformation of USPS into MNIST dataset. Although it is oblivious to the geometric structure of the images, it reveals how pixels would spatially rearrange in transportation.

\textbf{\textit{DA  theoretical guarantees using OT.}} Another virtue of OT is the possibility of leveraging theoretical error bounds that build on the  Wasserstein distance. 
 For example,  Courtey \etal \cite{CourtyJointAdaptation} proved that minimizing the joint distribution optimal transport (JDOT) quantity is equivalent to minimizing the learning bound on the domain adaptation problem. Wasserstein Guided Representation Learning (WDGRL) \cite{Shen2017WassersteinAdaptation}  builds on \cite{Redko2017TheoreticalTransport} and proves that the target error is bounded by the \marginpar{R1}\rtypo{Wasserstein distance for empirical measures under the assumption that hypothesis class is $K$-Lipschtiz continuous for some $K$.}

\subsection{Generative Modeling}
\label{sec:gen_modeling}

Learning models that generate images, audio, video, text, or other data is a major \marginpar{R1} \rtypo{subfield} of machine learning. 
Generative models have a vast range of possible applications and enormous potential can result from training on endlessly available unlabeled data. 
Let $X\sim P_r$, where $P_r$ be a fixed distribution from which we have access to samples. 
The fundamental problem of generative \rtypo{modeling} is to make distributions $P_z$ and $P_{r}$ as close as possible, where $P_z$ is a distribution we may sample from. 
In high-dimensional and large data regimes, this is often done by minimizing a divergence between the two distributions.

\textbf{\textit{Generative Adversarial Networks (GANs).}} The GAN consists of two networks --- a generator network $g_\theta \in \mathcal{G}$ and a $(0,1)$-valued discriminator network $d \in \mathcal{D}$ --- that are jointly trained according to a procedure that may be understood as a minimax game~\cite{GoodfellowGenerativeNets}.
Given some easy-to-sample distribution $P_0$, the generator implicitly defines a sample from a distribution \rtypo{$g_\theta(\vector{\rv{z}}_0)$}, where \rtypo{$\vector{\rv{z}}_0 \sim P_0$.}
The discriminator network $d$ is then used to evaluate whether the sample \rtypo{$g_\theta(\vector{\rv{z}}_0)$} comes from $P_z$ or $P_r$.
Whether or not the sample comes from $P_z$ or $P_r$ is measured through a quantity
\rtypo{\begin{equation}
\label{eq:gan_}
    \mathcal{L}_{\text{GAN}}(g_\theta) = \sup_{d\in\mathcal{D}}  \mathbb{E}_{\vector{\rv{X}}}[\log(d(\vector{\rv{x}}))] +
    \mathbb{E}_{\vector{\rv{z}}_0}[\log(1-d(g_\theta(\vector{\rv{Z}}))] .
\end{equation}}
The GAN's minimiax objective between the generator $g_\theta$ and the discriminator $d$ is then
\begin{equation}
\label{eq:gan}
    \inf_{\theta} \mathcal{L}_{\text{GAN}}(g_\theta).
\end{equation}
The quantity $\mathcal{L}_{\text{GAN}}$ provides a lower bound on the Jensen-Shannon divergence (JSD), up to a multiplicative and additive constant~\cite{GoodfellowGenerativeNets}. Furthermore, if $\mathcal{D}$ were replaced by the set of all $(0,1)$-valued functions, the lower bound would be an equality, and~\eqref{eq:gan} amounts to \rtypo{minimizing} the JSD~\cite{bousquet2017optimal}. 


\textbf{\textit{WGAN and followups.}} The notorious training instability of the vanilla GAN has been argued to be an issue associated with the JSD metric \cite{Arjovsky2017WassersteinNetworks}. Specifically, the JSD remains constant when the two distributions are disjoint (not overlapping). This can happen when $P_g$ and $P_r$ are low dimensional manifold in high dimensional space~\cite{bousquet2017optimal}. Additionally, even if they overlap, the limited sampling (during training) might render them disjoint. The WGAN~\cite{Arjovsky2017WassersteinNetworks} \rtypo{minimization} objective with respect to a generative network $g_\theta$ is 
\rtypo{\begin{equation}
\begin{aligned}
\mathcal{L}_{\text{WGAN}}(g_\theta)&= \sup_{f_w\in\mathcal{F}} \, \mathbb{E}_{\vector{\rv{x}} \sim P_r} [f_w(\vector{\rv{x}})] - \mathbb{E}_{\vector{\rv{z}} \sim P_g}  [f_w(g_{\theta}(\vector{\rv{Z}}))],
\label{eq:kr_dual_wgan}
\end{aligned}
\end{equation}} 
where $\mathcal{F}$ is a space of $1$-Lispchitz neural networks.
$f_w$ here replaces GAN's $(0,1)$-valued discriminator and is called the critic. 
Recall that the equivalent form of Kantorovich-Rubinstein duality \cite{Berlin2009OptimalNew} is
\rtypo{\begin{equation}
\begin{aligned}
W_1(\mu,\nu)&= \sup_{f\in\overline{\mathcal{F}}} \, \mathbb{E}_{\vector{\rv{x}} \sim \mu} [f(\vector{\rv{x}})] - \mathbb{E}_{\vector{\rv{z}} \sim \nu}  [f(\vector{\rv{Z}})],
\label{eq:kr_dual_wgan_}
\end{aligned}
\end{equation} }
where $\overline{\mathcal{F}}$ is the space of all $1$-Lispchitz functions.
Since $\mathcal{F} \subset \overline{\mathcal{F}}$, it follows that the GAN objective~\eqref{eq:kr_dual_wgan} is a lower bound of the $1$-Wasserstein distance~\eqref{eq:kr_dual_wgan_}.
The $1$-Lipschitz constraint ($\mathcal{L}ip(f)$) in \eqref{eq:kr_dual_wgan} also ensures a smooth critic and, thus, a more meaningful gradient for updating the generator. In practice, WGAN enforced this constraint using \rtypo{the weight clipping heuristic to enforce the parameters} to be in a compact space ($[-0.01,0.01]$). WGAN-LP \cite{petzkaregularization} augments the loss with a regularization term that penalizes the deviation of the critic's gradient norm from one. Followup works argued that explicit enforcement of $\mathcal{L}ip(f)$ might be unnecessary. WGAN-GP \cite{gulrajani2017improved} observed that the optimal critic has \rtypo{a gradient} norm equal to 1 almost everywhere. They replace the weight clipping with a gradient penalty term that encourages this characteristic. 
 Sobolev WGAN\cite{Artificial2021TowardsGANs} argues that the strong Lipschitz constraint might be unnecessary for optimization.
Another direction for tackling the instability that can be caused \rtypo{by imperfect optimization} of the critic is to use a fully tractable divergence. Genavy \etal \cite{Genevay2018LearningDivergences} used \rtypo{regularized} OT evaluated on mini-batches. 

\begin{figure}[t!]
    \centering
    \includegraphics[width = 1\linewidth]{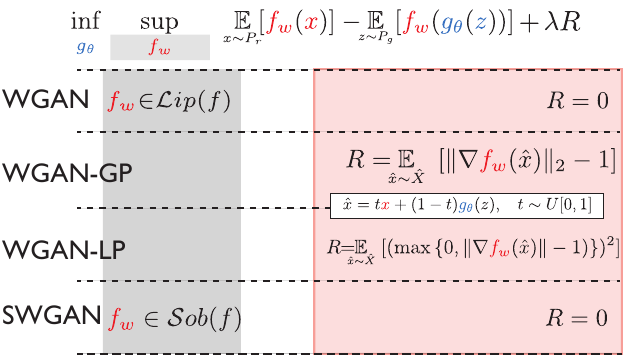}
    \caption{Connection between WGAN\cite{Arjovsky2017WassersteinNetworks} and the followup works WGAN-GP\cite{gulrajani2017improved}, WGAN-LP\cite{petzkaregularization} and SWGAN\cite{Artificial2021TowardsGANs}. $\mathcal{L}ip$ and $\mathcal{S}ob$ denote $1$-Lipschitz constraint and Sobolev integral constraint; respectively. }
    \label{fig:wgan_formulas}
\end{figure}

\textbf{\textit{Diffusion and score based methods.}} \marginpar{R3}\rref{Score based~\cite{song2019generative, song2020score} and diffusion generative models~\cite{sohl2015deep}} both rely on a similar mechanism. 
Input data (such as images or text) is perturbed by incrementally adding small amounts of noise by a procedure called the forward process.
The forward process is a so-called diffusion process.
The input data is recovered by running another diffusion process called the reverse process.
The reverse process involves the score function --- the derivative of the log probability density function of the perturbed data distribution --- which is unknown. 
Instead of using the exact score function, score-based methods use a neural network approximation.
The parameters of the neural network are learned using the so-called (conditional) score-matching loss, which is coarsely an expected weighted mean-squared error between the neural network and the true score function~\cite{song2021maximum}.
Samples from the approximate data distribution can be generated by running the reverse process.

The forward and reverse mechanisms do not explicitly minimize the divergence between the input data and the output of the reverse process.
Surprisingly, however, score-based methods are equivalent to minimizing an upper bound on the KL divergence between the reverse process and the input data distribution~\cite{song2021maximum}.
More recently, it has been shown that the 2-Wasserstein distance is upper bounded by the (conditional) score-matching loss, up to multiplicative and additive constants~\cite{Wu2022Score-basedDistance}.

\marginpar{R1}\rother{Several works have revealed the interesting connections between OT and diffusion models. Providing alternative interpretations and solutions to Diffusion-based generative modeling. The seminal work by Jordan \etal \cite{jordan1998variational} interpreted Langevin dynamics (a basic principals underlying diffusion models) as a gradient flow in the Wasserstein space of probability measures. Diffusion
Schrodinger Bridges  \cite{gushchin2022entropic, BortoliDiffusionModeling, chen2016relation} were shown to be strongly related to the entropy-regularized OT problem. The formulation can be solved using a modified version of Iterative
Proportional Fitting (a continuous analog of the Sinkhorn algorithm). Khrulkov \etal \cite{khrulkov2022understanding}  shows that, in specific cases,  the optimal transport (Monge) map
coincides with the Denoising Diffusion Probabilistic Models (DDPM) \cite{ho2020denoising} encoder map. Rectified Flow generative model \cite{liu2209flow} finds a transport map between two distributions by learning a flow (ordinary differential equation) that favours the transportation in straight paths.}

\marginpar{R1\\R3}\rother{In addition to the above, OT was utilized to optimize the diffusion model's pipeline. For example, to speed up the data sampling. Typically, diffusion models require many evaluation iterations to generate data samples.  OT methods were used to enable efficient sampling in Diffusion Models. DPM-OT \cite{li2023dpm} views the inverse diffusion as an OT problem, significantly improving sample quality and speed. 
The model leverages semi-discrete optimal transport maps to better capture the discontinuity of the target data manifold. The model is currently limited to unconditional data generation.}

\subsection{Other applications}
\label{sec:app_other_applications}

\rother{\textit{Graphs Matching:} Graph comparison and matching is concerned with estimating the structural correspondences of nodes between graphs. Due to its combinatorial nature, the problem is NP-complete, and approximate solutions are commonly sought.  In addition to the quadratic formulations for graph comparison discussed in Gromov Wasserstein (\fixedref{\begin{math}\mathbf{\S 4.6}\end{math}}), some approaches relax the problem into a linear assignment. Wang \etal \cite{wang2019learning} learns a supervised model for graph matching using a permutation loss that can work for an arbitrary number of nodes for graph matching. As the Sinkhorn layer was shown \cite{yu2019learning} to be the approximate and differentiable version of the Hungarian algorithm, it was used to ``soft'' assign graph correspondences. An issue here is that the ``soft'' output isn't necessarily a discrete permutation matrix. This is fixed by applying the Hungarian algorithm \cite{kuhn1955hungarian} on the output matrix. Other works \cite{wang2020graduated} used this post-processing trick for a similar purpose. In graph correspondence learning application, \cite{FeyDEEPCONSENSUS} uses Sinkhorn to initiate soft correspondence between graph nodes before refining it using message passing.}

\rother{\textit {Reinforcement Learning (RL):}  Reinforcement Learning (RL) is a framework that aims at effectively solving sequential decision-making tasks. In this framework, the learning agent interacts with the environment to improve its performance through trial and error\cite{sutton2018reinforcement}. So far,  there is limited adoption of optimal transport techniques in reinforcement learning. Of the few works in this area, \cite{klink2022curriculum,huang2022curriculum} applied OT to Curriculum Reinforcement Learning (CRL). CRL approaches complex and challenging RL scenarios by gradually increasing the difficulty of tasks presented to the learning agent. This way, it evades extensive engineering of reward functions. One interpretation of this paradigm considers the curriculum as a sequence of task distributions interpolating between auxiliary (intermediate) task distributions and target task distributions. Algorithmically, the process involves using KL divergence, which OT replaced in \marginpar{R2} \rref{\cite{klink2022curriculum,huang2022curriculum}} to improve the learning performance.}

\section{Optimization Background} \label{sec:app_optimization}

\rtypo{This section provides a concise description of some of the most important concepts from optimization theory relevant to new OT researchers and machine learning researchers learning and using OT. The section is not meant to be comprehensive although effort was made to ensure its accessibility to a wide audience not familiar with mathematical optimization theory. While this section can serve as a starting point for the uninformed readers, we also refer the readers to the following more comprehensive textbooks on optimization theory  \cite{boyd2004convex,nocedal2006numerical,bertsekas2016nonlinear}.}

\subsection{Minimum cost flow problems}

An exact optimal transport problem on discrete measures \rtypo{$\mu$ and $\nu$, with probability masses $\vector{a}$ and $\vector{b}$ over their $d_{\mathbb{X}}$ and $d_{\mathbb{Y}}$ points respectively,} can be formulated as the classic minimum cost flow (MCF) network problem which is a graph-theoretic problem that can be formulated using the following linear program:
\begin{mini}|l|[3]
  {\matrix{P}}{\rtypo{\mathcal{L}_c(\bm{\mu, \nu}) = \sum_{i=1}^{d_{\mathbb{X}}} \sum_{j=1}^{d_{\mathbb{Y}}} c(\vector{x}_i, \vector{y}_j) \matrix{P}_{ij}}}{}{}
  \addConstraint {\rtypo{\sum_{j=1}^{d_{\mathbb{Y}}} \matrix{P}_{ij} = a_i}}{\quad \forall i \in \rtypo{[1, d_{\mathbb{X}}]}}{}
  \addConstraint {\rtypo{\sum_{i=1}^{d_{\mathbb{X}}} \matrix{P}_{ij} = b_j}}{\quad \forall j \in \rtypo{[1, d_{\mathbb{Y}}]}}{}
  \addConstraint {\rtypo{\matrix{P}_{ij}} \geq 0}{\quad \forall i \in \rtypo{[1, d_{\mathbb{X}}]}, j \in \rtypo{[1, d_{\mathbb{Y}}]}}{}
\end{mini}
where $c$ is a cost function, usually a distance measure between \rtypo{$\vector{x}_i$ and $\vector{y}_j$}. General linear programs can be solved using a number of algorithms with the Simplex method \cite{Dantzig1997Linear1,Dantzig2003Linear2,Bradley1977AppliedProgramming} and the interior point method \cite{Wright1997Primal-DualMethods} being 2 of the most popular choices. The computational complexity of the Simplex method is exponential in the number of decision variables in the worst case while the interior point method has polynomial time complexity in the number of decision variables. However, in practice, the Simplex method tends to be significantly faster for most practical problems. Besides computational complexities and running time, the main difference between the Simplex method and the interior point method from a user's perspective is that the Simplex method always finds a so-called "extreme" or corner point of the feasible domain as the optimal solution, while the interior point method finds a point in the relative interior of the feasible domain as the optimal solution. An extreme point is a point at the intersection of $N$ hyper-planes where $N$ is the number of decision variables \rtypo{($d_{\mathbb{X}} \times d_{\mathbb{Y}}$ above)}. Since the optimal transport problem has \rtypo{$d_{\mathbb{X}} + d_{\mathbb{Y}}$} equality constraints, assuming \rtypo{$d_{\mathbb{X}} \times d_{\mathbb{Y}} > d_{\mathbb{X}} + d_{\mathbb{Y}}$, the remaining $d_{\mathbb{X}} \times d_{\mathbb{Y}} - d_{\mathbb{X}} - d_{\mathbb{Y}}$} constraints to be satisfied at equality will be bound constraints of the form $\matrix{P}_{ij} \geq 0$. The Simplex method will therefore naturally tend to find sparse optimal plans. The relative interior of a feasible domain is the set of points that satisfy all the linear equality constraints (e.g. \rtypo{$\sum_{j=1}^{d_{\mathbb{Y}}} \matrix{P}_{ij} = a_i$}) at equality, but with all the inequality constraints ($\matrix{P}_{ij} \geq 0$) satisfied at a strict inequality.


That said, using the standard Simplex method or interior point method for generic linear programs is not the most efficient way to solve minimum cost flow (MCF) problems. MCF problems have a wide variety of more efficient algorithms with polynomial time complexities \cite{Kelly1991TheMethod}, one of the most famous of which is the so-called network Simplex method \cite{Orlin1997AFlows}. The network Simplex method is a variation of the general Simplex method specialized for MCF problems to have a \rtypo{lower cost per iteration} and a polynomial bound on the number of iterations needed to converge. The general Simplex algorithm is both more expensive per iteration and has an exponential worst case complexity for the number of iterations needed to converge. More recently, faster algorithms for the MCF problem have also been proposed \cite{Chen2022MaximumTime}.

\subsection{Convex optimization}

\subsubsection{Set convexity}

A set of points $\mathcal{S} = \{ \bm{x} \}$ is said to be convex if every weighted average of some points inside the set is also inside the set. Equivalently, for a set $\mathcal{S}$ to be convex, the following must be satisfied for all $(\bm{x}_2, \bm{x}_2) \in \mathcal{S}$ and for all $0 < \alpha < 1$.
\begin{align}
    \alpha \times \bm{x}_1 + (1 - \alpha) \times \bm{x}_2 \in \mathcal{S}
\end{align}
The above weighted average is also known as the convex combination of $\bm{x}_1$ and $\bm{x}_2$.

The intersection of any number of convex sets is either the empty set or another convex set. Therefore, in an optimization problem with multiple constraints, if the set of feasible points to each constraint individually is a convex set, the set of feasible points to all the constraints is either the empty set (problem is infeasible) or another convex set.

\subsubsection{Function convexity}

\begin{figure}[ht]
  \centering
  \includegraphics[width=0.7\linewidth]{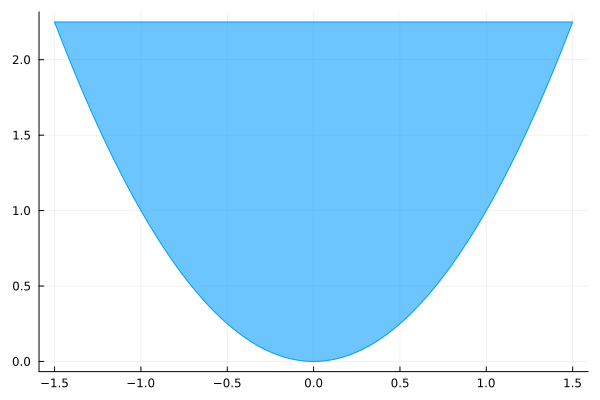}
  \caption{Epigraph of $f(\vector{x}) = x^2$.}
  \label{fig:epigraph}
\end{figure}

A function $f(\bm{x})$ with domain $\mathcal{X}$ is called convex if the set of all points in $\mathcal{X}$ satisfying the constraint \rtypo{$f(\bm{x}) \leq \alpha$} is either an empty or a convex set for all \rtypo{$\alpha \in \mathbb{R}$}. Note that the set of such points doesn't have to be a compact set, that is it can extend to $\pm \infty$ along one or more directions. Equivalently, one can say that a function is convex if the set of points in its epigraph is a convex set. The eipgraph of the function $f(\bm{x})$ is the set of points \rtypo{$\{(\bm{x}, \alpha) \in \mathcal{X} \times \mathbb{R}: f(\bm{x}) < \alpha \}$}. Figure \ref{fig:epigraph} shows the epigraph of $f(x) = x^2$.


Showing that a function is convex can be done by proving any of the above properties. There are however other ways to check if a function is convex. Another common check for twice differentiable functions is to check if the function's Hessian is positive semi-definite everywhere in $\mathcal{X}$. Identifying and building convex functions in a disciplined way is a well-studied field, often termed "disciplined convex programming" (DCP). For more on DCP, the readers are advised to check the excellent online resources available in \url{https://dcp.stanford.edu}.

\subsubsection{Constraint convexity}

In optimization, a constraint is said to be a convex constraint if the set of points satisfying this constraint is a convex set. One common way to construct a convex constraint is using a convex function $f(\bm{x})$. If $f$ is a convex function, the following constraint is convex:
\begin{align}
    f(\bm{x}) \leq \rtypo{\alpha}
\end{align}
for any constant \rtypo{$\alpha$}, assuming there exists at least 1 point satisfying this constraint. Other common convex constraints are affine (aka linear) constraints of the form:
\begin{align}
    f(\bm{x}) = 0
\end{align}
where $f$ is an affine function of $\bm{x}$, e.g. $f(\bm{x}) = \bm{b}^T \bm{x} + a$ for some constants $\bm{b}$ and $a$. Note that it's customary to use the term "linear" to refer to affine functions and/or constraints in optimization literature.

A constrained optimization problem with a convex \textit{minimization} objective function and convex constraints is called a "convex program". Formulating decision problems as convex programs and solving them with convex optimization algorithms is often called "convex programming". Convex programs are theoretically appealing because under mild assumptions, they have efficient algorithms that can find a so-called \textit{global optimal solution}. A global optimal solution is a (not necessarily unique) feasible point $\bm{x}$ that absolutely minimizes the objective function, i.e. no other solution can do strictly better.

Note that it has to be a minimization problem. Maximizing a general convex function is NP-hard. If your objective $f(\bm{x})$ is a quantity you want to maximize, that's equivalent to minimizing $-f(\bm{x})$ instead. If $-f(\bm{x})$ is a convex function, $f$ is known as a concave function. Therefore for a maximization optimization problem to be convex, the objective function must be concave and the constraints must be convex.

\subsubsection{Structured constraints}

Among convex constraints, there are some special classes of constraints that are generally considered nicer than "generic" convex constraints. These constraints have a specific structure which can be exploited in more efficient algorithms. These are broadly known as "\textit{structured constraints}". A simple class of structured constraints are the affine/linear constraints, e.g:
\begin{align}
    \bm{b}^T \bm{x} + a \quad (\leq / =) \quad 0
\end{align}
There are some algorithms that can handle linear inequality constraints more efficiently than generic convex constraints. Additionally, linear equality constraints are the only equality constraints that are convex. Any other equality constraint where the function in the constraint is not affine/linear can never be convex. Linear constraints are therefore among those special classes of so-called \textit{structured constraints}.

Another broad class of structured constraints is conic constraints. A conic constraint is a convex constraint whose feasible set is a special convex set known as a cone. A convex set $S$ is considered a cone if for every $\bm{x}$ inside the set $S$, \rtypo{$\alpha \bm{x}$} is also inside the set $S$ for all \rtypo{$\alpha \geq 0$}. Note how a cone must be a non-compact set since \rtypo{$\alpha$} is allowed to go to $\infty$. However, a conic program can have multiple conic constraints whose intersection is a compact set.

A structured conic constraint can be written as:
\begin{align}
    \bm{x} \in \mathcal{C}
\end{align}
where $\mathcal{C}$ is a cone. There are various algorithms that can handle conic constraints more efficiently than regular convex constraints. It is therefore often appealing to convert regular mathematical constraints that are known to be convex to their conic equivalents to make use of these efficient algorithms \cite{mosekcookbook}. 

\subsubsection{Algorithms}

There are various classes of optimization algorithms for solving each class of optimization problems. For linear programs, the most popular algorithms are:
\rtypo{
\begin{itemize}
  \item Primal/dual Simplex algorithm \cite{hillier2020introduction}
  \item Interior point algorithm \cite{nesterov1994interior,boyd2004convex}
  \item Primal-dual hybrid gradient algorithm \cite{pdhg_2021_linear}
\end{itemize}
}

\rtypo{
For unstructured convex and general nonlinear programs, the following are popular algorithms:
\begin{itemize}
  \item Interior point algorithm \cite{nesterov1994interior,boyd2004convex,Wachter2006}
  \item Sequential quadratic programming (with trust region) \cite{nocedal2006numerical}
  \item Method of moving asymptotes algorithm \cite{svanberg_1987_mma,svanberg_2002_mma}
  \item Augmented Lagrangian algorithm \cite{bertsekas1982constrained}
\end{itemize}
}

\rtypo{
For convex and general nonlinear programs with an unstructured objective but structured convex (conic) and linear constraints, the following are popular algorithms:
\begin{itemize}
  \item (Stochastic) Frank-Wolfe algorithm \cite{frank1956algorithm,xiao2016stochastic,FW_2023_RG}
  \item Alternating direction method of multipliers (ADMM) algorithm \cite{admm_2011}
  \item Projected gradient descent algorithm \cite{parikh_proximal}
  \item Mirror descent algorithm \cite{BECK2003167}
\end{itemize}
}

\rtypo{
More generally, "proximal algorithms" \cite{parikh_proximal} are a family of algorithms for handling this class of problems which encompasses multiple of the above algorithms, such as ADMM, projected gradient descent, mirror descent, and many more algorithms.
}
\rtypo{
For structured convex (conic) programs, the following are popular algorithms:
\begin{itemize}
  \item Interior point algorithm \cite{nesterov1994interior,boyd2004convex}
  \item ADMM with conic operator splitting \cite{Garstka_2021}
\end{itemize}
}

\subsection{Differentiability}

A function is any one-to-one or many-to-one (but not one-to-many) mapping from a set called the domain to another set called the range of the function. The uniqueness of the value in the range for each value in the domain is a characteristic property of what a function is. For some parametric optimization problems with parameters $\vector{\theta}$, there exists a unique optimal solution $\vector{x}^*(\vector{\theta})$. In such cases, $\vector{x}^*$ can be considered a proper function of $\vector{\theta}$. Functions can be continuous, \rtypo{discontinuous} or semi-continuous. Continuity of a function means that for every infinitesimal change in the inputs, an infinitesimal change takes place in the output. In the context of optimization problems, this means that for every small change in $\vector{\theta}$, $\vector{x}^*$ only changes by a small amount. Such continuity assumption is generally violated if the solution $\vector{x}^*$ is discrete since discrete solutions can only move in non-infinitesimal steps, except in the trivial case where the solution $\vector{x}^*$ does not change at all with $\vector{\theta}$. Semi-continuous functions are continuous almost everywhere but have a zero-measure subset of the domain where the function has a \rtypo{discontinuity}. Optimal solutions to non-convex optimization problems with multiple modes can exhibit such semi-continuity.

\subsection{Duality}

\subsubsection{What is duality?}

Duality is one of the most important concepts in optimization. Sometimes, instead of solving an optimization problem in its "natural formulation", also known as the primal formulation, one can solve a related optimization problem, known as the dual problem, or one can solve the primal and dual problems together. Let the primal optimization problem be:
\begin{mini}|l|[3]
  {\vector{x}}{f(\vector{x})}{}{}
  \addConstraint {\vector{x} \in \mathcal{P}}{}
\end{mini}
where $\vector{x}$ is the vector of primal decision variables, $f$ is the primal objective function, and $\mathcal{P}$ is the feasible domain of the primal problem, i.e. the set of all points that satisfy any constraints on $\vector{x}$. Additionally, let $\vector{x}^*$ be the unknown optimal solution of the primal optimization problem. The dual problem would be another optimization problem of some so-called dual variables \rtypo{$\vector{\lambda}$}:
\begin{maxi}|l|[3]
  {\rtypo{\vector{\lambda}}}{\rtypo{h(\vector{\lambda})}}{}{}
  \addConstraint {\rtypo{\vector{\lambda}} \in \mathcal{D}}{}
\end{maxi}
where \rtypo{$h$} is the dual problem's objective and $\mathcal{D}$ is the feasible domain of the dual problem such that for any feasible \rtypo{$\vector{\lambda} \in \mathcal{D}$, $h(\vector{\lambda}) \leq f(\vector{x}^*)$}. The fact that every dual feasible solution \rtypo{$\vector{\lambda}$} gives a lower bound \rtypo{$h(\vector{\lambda})$} on the optimal primal objective value $f(\vector{x}^*)$ is what makes the primal and dual problems entangled with each other. The optimal dual solution \rtypo{$\vector{\lambda}^* \in \mathcal{D}$ that maximizes $h$} gives the best/tightest lower bound on the optimal primal objective. In problems where the so-called strong duality exists, the best bound \rtypo{$h(\vector{\lambda}^*)$} is exactly equal to the best primal objective value $f(\vector{x}^*)$. For linear programs and convex programs (subject to a mild constraint qualification condition), strong duality holds.

\subsubsection{Importance of the dual problem}

There are various reasons to formulate and solve the dual problem. One reason is to identify if a primal solution is near optimal. Assuming the primal problem is a minimization problem, a dual solution that's feasible to the dual problem provides a lower bound on the optimal objective value of the primal problem. Having a good (tight) lower bound on the optimal value of our primal problem helps us to know how far we are from optimality in the worst case. For instance, let $\vector{x}$ be the best feasible primal solution found and let \rtypo{$\vector{\lambda}$} be a feasible dual solution. In this case, we know that \rtypo{$h(\vector{\lambda}) \leq f(\vector{x}^*) \leq f(\vector{x})$}. A feasible dual solution with a higher \rtypo{$h(\vector{\lambda})$} or a feasible primal solution with a lower $f(\vector{x})$ provide tighter bounds around the optimal value $f(\vector{x}^*)$. Therefore knowing \rtypo{$\vector{\lambda}$}, we know that $f(\vector{x})$ can never be more than \rtypo{$f(\vector{x}) - h(\vector{\lambda})$} away from $f(\vector{x}^*)$. When strong duality applies, the best bound \rtypo{$h(\vector{\lambda}^*)$} is exactly equal to the best primal objective value $f(\vector{x}^*)$. In this case, the dual solution can be used to provide a proof/certificate of convergence to the optimal solution if we evaluate $f(\vector{x})$ and \rtypo{$h(\vector{\lambda})$} for the best $\vector{x} \in \mathcal{P}$ and \rtypo{$\vector{\lambda} \in \mathcal{D}$} found and they turn out to be equal. However, strong duality doesn't always exist for all classes of optimization problems.

Another use of the dual problem is that sometimes it's beneficial to solve the primal and dual problems simultaneously like in the primal-dual hybrid gradient algorithm for large scale linear programs or the primal-dual interior point method for unstructured convex and nonlinear programs. In these cases, the dual problem can guide the optimization algorithm improving the convergence rate of the primal solution.

In some other cases, \rtypo{e.g.,} some linear and convex/conic programs, the dual can also have a special structure enabling more efficient algorithms or faster convergence compared to solving the primal formulation. In such cases, it's also common that there exists an easy way to get the optimal primal solution $\vector{x}^*$ corresponding to the optimal dual solution \rtypo{$\vector{\lambda}^*$}. In such cases, it might make sense to solve the dual problem alone instead of the primal one and then recover $\vector{x}^*$ from \rtypo{$\vector{\lambda}^*$} at the end.

\subsubsection{Deriving the dual problem (general)}

To derive the dual of an optimization problem, the following steps are followed:
\begin{enumerate}
    \item Derive the Lagrangian function
    \item Derive the Lagrangian relaxation problem
    \item Derive the Lagrangian dual function
    \item Derive the Lagrangian dual problem
    \item Optional: simplify the dual problem
\end{enumerate}
The \textbf{Lagragnian function} is a function of the objective and some (or all) of the constraints of the primal optimization problem such that the minimum of the Lagrangian function, subject to the non-relaxed constraints, is a lower bound on the optimal value of the primal problem. The standard way to construct the Lagrangian function is by relaxing some or all of the constraints in the primal problem and adding a function of them to the objective that \rtypo{ensures} the above property. For instance, let the primal problem be an inequality constrained optimization problem:
\begin{mini}|l|[3]
  {\vector{x}}{f(\vector{x})}{}{}
  \addConstraint {\vector{g}(\vector{x}) \leq \vector{0} \, (\text{element-wise})}{}
  \addConstraint {\vector{x} \in \rtypo{\mathcal{\tilde{P}}}}{}
\end{mini}
where $f$ is the objective function, $\vector{g}$ is the constraint function \rtypo{with $d_g$ outputs, and $\mathcal{\tilde{P}}$} encodes all the "other constraints" if any exist or just the Euclidean space if no other constraints exist. The Lagrangian function resulting from relaxing the inequality constraint would be:
\begin{align}
  \rtypo{L(\vector{x}, \vector{\lambda}) = f(\vector{x}) + \langle \vector{\lambda}, \vector{g}(\vector{x}) \rangle}
\end{align}
where \rtypo{$\vector{\lambda} \in \mathbb{R}_+^{d_g}$} is a fixed vector of non-negative coefficients known as the \textbf{Lagrangian multipliers} or the \textbf{dual decision variables}. The so-called \textbf{Lagrangian relaxation} of the primal problem would then be a parametric optimization problem parameterized by $\vector{\lambda}$ as follows:
\begin{mini}|l|[3]
  {\vector{x}}{L(\vector{x}, \vector{\lambda}) = f(\vector{x}) + \langle \vector{\lambda}, \vector{g}(\vector{x}) \rangle}{}{}
  \addConstraint {\vector{x} \in \rtypo{\mathcal{\tilde{P}}}}{}
\end{mini}

Since any solution $\vector{x} \in \rtypo{\mathcal{\tilde{P}}}$ that is \textit{feasible to the primal problem} will satisfy $\vector{g}(\vector{x}) \leq \vector{0}$ and since $\vector{\lambda}$ is non-negative, $L(\vector{x}, \vector{\lambda}) \leq f(\vector{x})$ is true for all primal feasible solutions $\vector{x}$. This is true for the optimal solution to the primal problem $\vector{x}^*$ as well, i.e. $L(\vector{x}^*, \vector{\lambda}) \leq f(\vector{x}^*)$.

Let the minimum objective value of the Lagrangian relaxation problem for some choice of $\vector{\lambda}$ be:
\begin{align}
  L^*(\vector{\lambda}) = \min_{\vector{x} \in \rtypo{\mathcal{\tilde{P}}}} L(\vector{x}, \vector{\lambda})
\end{align}
$L^*(\vector{\lambda})$ is known as the \textbf{Lagrangian dual function}. It is important to note that the $\vector{x} \in \rtypo{\mathcal{\tilde{P}}}$ that minimizes $L(\vector{x}, \vector{\lambda})$ in the Lagrangian relaxation problem may or may not satisfy the relaxed inequality constraint $\vector{g}(\vector{x}) \leq \vector{0}$.

The key condition that a Lagrangian dual function must satisfy by definition is that $L^*(\vector{\lambda}) \leq f(\vector{x}^*)$ for any $\vector{\lambda} \in \rtypo{\mathbb{R}_+^{d_g}}$. A sufficient condition to ensure this property is satisfied is that $L(\vector{x}, \vector{\lambda}) \leq f(\vector{x})$ for any $\vector{x}$ that is also feasible to the primal problem and any $\vector{\lambda} \in \rtypo{\mathbb{R}_+^{d_g}}$, since $L^*(\vector{\lambda}) \leq L(\vector{x}^*, \vector{\lambda}) \leq f(\vector{x}^*)$ would then be true by definition for all $\vector{\lambda} \in \rtypo{\mathbb{R}_+^{d_g}}$.

\rtypo{To generalize the above Lagrangian dual function to different constraint types, other than $\vector{g}(\vector{x}) \leq \vector{0}$, the following quantities and sets can be defined differently for each constraint type to ensure a generalization of the aforementioned property is satisfied:
\begin{enumerate}
    \item Lagrangian multipliers $\vector{\lambda}$,
    \item Multiplier domain $\mathcal{D}_L$ (instead of \rtypo{$\mathbb{R}_+^{d_g}$}), and
    \item Lagrangian function $L(\vector{x}, \vector{\lambda})$
\end{enumerate}}

\rtypo{Note that the above terms can all be trivially generalized to combinations of constraint types by breaking down $\vector{\lambda}$ into multiple components, one per relaxed constraint type. The Lagrangian dual feasible domain $\mathcal{D}_L$ can therefore be more generally defined as the Cartesian product of the domains of the dual variables associated with all the relaxed constraint sets (grouped by constraint type). Each relaxed set of constraints has a dual variable domain associated with it to ensure the lower bound property of the Lagrangian. The following are some of the most common dual variable domains for common constraint types:
\begin{center}
\begin{tabular}{||c c||} 
 \hline
 Constraint type & Dual variable domain $\mathcal{D}_L$ \\ [0.5ex] 
 \hline\hline
 $\vector{g}(\vector{x}) \leq \vector{0}$ element-wise & $\mathbb{R}_+^{d_g}$: non-negative numbers \\ 
 \hline
 $\vector{g}(\vector{x}) \geq \vector{0}$ element-wise & $\mathbb{R}_-^{d_g}$: non-positive numbers \\
 \hline
 $\vector{g}(\vector{x}) = \vector{0}$ element-wise & $\mathbb{R}^{d_g}$: all real numbers \\
 \hline
 $\vector{x} \in \mathcal{C}$ where $\mathcal{C}$ is a cone & The polar cone $\mathcal{C}^o$ of $\mathcal{C}$ \\ [1ex] 
 \hline
\end{tabular}
\end{center}}
The polar cone of a cone $\mathcal{C}$ is the set of points $\{ \vector{\lambda} \}$ in the Euclidean space such that the inner product $\langle \vector{\lambda}, \vector{x} \rangle \leq 0\, \forall \vector{x} \in \mathcal{C}$. This definition ensures that the Lagrangian is a lower bound on the objective function for any feasible $\vector{x}$ and $\vector{\lambda} \in \mathcal{D}_L$.

Using the more general notation, the Lagrangian dual function $L^*(\vector{\lambda})$ is a lower bound on the optimal primal objective value $f(\vector{x}^*)$ for every $\vector{\lambda} \in \mathcal{D}_L$. Since for every $\vector{\lambda} \in \mathcal{D}_L$, $L^*(\vector{\lambda}) \leq f(\vector{x}^*)$, the choice of $\vector{\lambda} \in \mathcal{D}_L$ that gives the tightest bound is the one that maximizes $L^*(\vector{\lambda})$. This optimization problem in $\vector{\lambda}$ is known as the \textbf{dual problem}.
\begin{maxi}|l|[3]
  {\vector{\lambda}}{L^*(\vector{\lambda}) = \min_{\vector{x} \in \rtypo{\mathcal{\tilde{P}}}} L(\vector{x}, \vector{\lambda})}{}{}
  \addConstraint {\vector{\lambda} \in \mathcal{D}_L}{}
\end{maxi}
This is a max-min optimization problem which from a first look may seem more complicated than the primal problem. However, for classes of optimization problems where the inner minimization of the Lagrangian function \rtypo{with respect to} $\vector{x}$ has a closed form solution, the dual problem can be significantly simplified. We will see examples in the next section.

When no closed form solution exists for the Lagrangian relaxation, instead of solving the dual problem as a max-min problem, primal-dual algorithms solve for $\vector{x}$ and $\vector{\lambda}$ simultaneously. For instance in the primal-dual interior point method, \rtypo{the optimality conditions of the outer and inner optimizations} are written out as a system of (potentially nonlinear) equations in terms of $\vector{x}$ and $\vector{\lambda}$ which are then solved using a variation of Newton's method.

\subsection{Dual of a general linear program}

In this section, we will apply the general steps from the previous section to a linear program in its canonical form. Consider the following primal linear program:
\begin{mini}|l|[3]
  {\vector{x}}{\langle \vector{c}, \vector{x} \rangle}{}{}
  \addConstraint {\matrix{A}\vector{x} \geq \vector{b}}{}
\end{mini}
for some constants $\vector{c}$, $\matrix{A}$ and $\vector{b}$ of the appropriate dimensions where $\langle .. \rangle$ is the inner product operator. There is no loss in generality in assuming the above form. Any $\leq$ constraints can be trivially transformed to $\geq$ constraints by multiplying the constraints by -1. \rtypo{Additionally, equality constraints of the form $\matrix{\tilde{A}} \vector{x} = \vector{d}$ can be transformed to the following 2 sets of constraints:
\begin{enumerate}
    \item $\matrix{\tilde{A}} \vector{x} \geq \vector{d}$
    \item $\matrix{\tilde{A}} \vector{x} \leq \vector{d}$ which is equivalent to $-\matrix{\tilde{A}} \vector{x} \geq -\vector{d}$
\end{enumerate}
Therefore setting $\matrix{A} = \begin{bmatrix} \matrix{\tilde{A}} \\ -\matrix{\tilde{A}} \end{bmatrix}$ and $\vector{b} = \begin{bmatrix} \vector{d} \\ -\vector{d} \end{bmatrix}$, we recover the canonical form.}

The Lagrangian relaxation of the above linear program is:
\begin{mini}|l|[3]
  {\vector{x}}{L(\vector{x}, \vector{\lambda}) = \langle \vector{c}, \vector{x} \rangle + \langle \vector{\lambda}, \vector{b} - \matrix{A}\vector{x} \rangle}{}{}
  \breakObjective{= \langle \vector{x}, \vector{c} - \matrix{A}^T \vector{\lambda} \rangle + \langle \vector{b}, \vector{\lambda} \rangle}
  \addConstraint{\vector{\lambda} \in \mathbb{R}_+^{m}}{}
\end{mini}
\rtypo{where $m$ is the number of constraints in $\matrix{A} \vector{x} \geq \vector{b}$}, and
$\vector{\lambda}$ is the vector of Lagrangian multipliers. This is just minimizing an affine function of $\vector{x}$. The optimal value $L^*(\vector{\lambda})$ of the above relaxation has the following closed form solution:
\begin{align}
  L^*(\vector{\lambda}) = \begin{cases}
   \langle \vector{b}, \vector{\lambda} \rangle & \text{if } \vector{c} - \matrix{A}^T \vector{\lambda} = \vector{0} \\
   -\infty & \text{otherwise}
  \end{cases}
\end{align}
The general form of the Lagrangian dual problem can be written as:
\begin{maxi}|l|[3]
  {\vector{\lambda}}{\min_{\vector{x}} L(\vector{x}, \vector{\lambda})}{}{}
  \addConstraint{\vector{\lambda} \in \mathbb{R}_+^m}{}
\end{maxi}
but it can then be simplified given the closed form solution as follows:
\begin{maxi}|l|[3]
  {\vector{\lambda}}{\langle \vector{b}, \vector{\lambda} \rangle}{}{}
  \addConstraint{\matrix{A}^T \vector{\lambda}}{ = \vector{c}}
  \addConstraint{\vector{\lambda}}{\in \mathbb{R}_+^m}
\end{maxi}
Note that the $-\infty$ branch when $\vector{c} - \matrix{A}^T \vector{\lambda} \neq 0$ can be ignored in the simplified form since $-\infty$ is clearly less than $\langle \vector{b}, \vector{\lambda} \rangle$ and we are only interested in the $\vector{\lambda}$ that maximizes the dual function $L^*(\vector{\lambda})$. Also note that the simplification led to some additional constraints on $\vector{\lambda}$ compared to the general Lagrangian dual formulation. The above problem is known as the dual linear program. Similar derivation steps can be followed for conic and other special classes of optimization problems to derive their dual formulation.

In some cases, it can be more efficient to solve the dual linear program instead of the primal one. For linear programs, strong duality exists, and there exists an efficient way to recover the optimal primal solution $\vector{x}^*$ from the optimal dual solution $\vector{\lambda}^*$ so one has the choice to solve either the primal or dual problems. The dual can also be exploited in algorithms directly without changing the problem's formulation. For instance, the dual Simplex method is an efficient way to perform the regular Simplex method on the dual formulation but without having to explicitly derive the dual formulation.

\subsection{Dual of the exact optimal transport problem}

In this section, we shall derive the dual problem of the exact discrete optimal transport problem:
\begin{mini}|l|[3]
  {\matrix{P}}{\langle \matrix{P}, \matrix{C} \rangle}{}{}
  \addConstraint{\matrix{P} \vector{1}}{ = \vector{a}}
  \addConstraint{\matrix{P}^T \vector{1}}{ = \vector{b}}
  \addConstraint{\matrix{P}_{ij}}{\geq 0 \quad \forall (i, j) \in [1, d_{\mathbb{X}}] \times [1, d_{\mathbb{Y}}]}
\end{mini}
\rtypo{where $\matrix{C}_{ij} = c(\vector{x}_i, \vector{y}_j)$.}

The Lagrangian relaxation of the above linear program is:
\rtypo{\begin{mini}|l|[3]
  {\matrix{P}}{\langle \matrix{P}, \matrix{C} \rangle + \langle \vector{\lambda}_{\vector{a}}, \vector{a} - \matrix{P} \vector{1} \rangle + \langle \vector{\lambda}_{\vector{b}}, \vector{b} - \matrix{P}^T \vector{1} \rangle}{}{}
  \breakObjective{+ \langle \vector{\lambda}_{-}, \matrix{P} \rangle}
  \addConstraint{\vector{\lambda}_{\vector{a}}}{ \in \mathbb{R}^{d_{\mathbb{X}}}}
  \addConstraint{\vector{\lambda}_{\vector{b}}}{ \in \mathbb{R}^{d_{\mathbb{Y}}}}
  \addConstraint{\vector{\lambda}_{-}}{ \in \mathbb{R}_-^{d_{\mathbb{X}} \times d_{\mathbb{Y}}}}
\end{mini}}
Note that $\langle \vector{a}, \matrix{P} \vector{b} \rangle$ can be re-written as $\langle \bm{P}, \bm{a} \bm{b}^T \rangle$ for any vectors $\bm{a}$ and $\bm{b}$:

\begin{align*}
\langle \bm{a}, \bm{P} \bm{b} \rangle & = \bm{a}^T \bm{P} \bm{b} \\
& = \text{tr}(\bm{a}^T \bm{P} \bm{b}) \\
& = \text{tr}(\bm{P} \bm{b} \bm{a}^T) \\
& = \langle \bm{P}^T, \bm{b} \bm{a}^T \rangle \\
& = \langle \bm{P}, \bm{a} \bm{b}^T \rangle
\end{align*}

Similarly, $\langle \bm{a}, \bm{P}^T \bm{b} \rangle = \langle \bm{P}, \bm{b} \bm{a}^T \rangle$.

\vline

Re-writing the Lagrangian relaxation using these identities and removing the \rtypo{$\in \mathbb{R}$} constraints:
\rtypo{\begin{mini}|l|[3]
  {\matrix{P}}{L(\matrix{P}, \vector{\lambda}_{\vector{a}}, \vector{\lambda}_{\vector{b}}, \vector{\lambda}_{-})}{}{}
  \breakObjective{ = \langle \matrix{P}, \matrix{C} - \vector{\lambda}_{\vector{a}} \vector{1}^T - \vector{1} \vector{\lambda}_{\vector{b}}^T + \vector{\lambda}_{-} \rangle}
  \breakObjective{ + \langle \vector{\lambda}_{\vector{a}}, \vector{a} \rangle + \langle \vector{\lambda}_{\vector{b}}, \vector{b} \rangle}
  \addConstraint{\vector{\lambda}_{-}}{ \in \mathbb{R}_-^{d_{\mathbb{X}} \times d_{\mathbb{Y}}}}
\end{mini}}

This is just minimizing an affine function of $\matrix{P}$. The optimal value \rtypo{$L^*(\vector{\lambda}_{\vector{a}}, \vector{\lambda}_{\vector{b}}, \vector{\lambda}_{-})$} of the above relaxation has the following closed form solution:
\rtypo{\begin{multline}
  L^*(\vector{\lambda}_{\vector{a}}, \vector{\lambda}_{\vector{b}}, \vector{\lambda}_{-}) = \\ \begin{cases}
   \langle \vector{\lambda}_{\vector{a}}, \vector{a} \rangle + \langle \vector{\lambda}_{\vector{b}}, \vector{b} \rangle & \text{if } \matrix{C} - \vector{\lambda}_{\vector{a}} \vector{1}^T - \vector{1} \vector{\lambda}_{\vector{b}}^T + \vector{\lambda}_{-} = \vector{0} \\
   -\infty & \text{otherwise}
  \end{cases}
\end{multline}}
The general form of the Lagrangian dual problem can be written as:
\rtypo{\begin{maxi}|l|[3]
  {\vector{\lambda}_{\vector{a}}, \vector{\lambda}_{\vector{b}}, \vector{\lambda}_{-}}{\min_{\matrix{P}} L(\matrix{P}, \vector{\lambda}_{\vector{a}}, \vector{\lambda}_{\vector{b}}, \vector{\lambda}_{-})}{}{}
  \addConstraint{\vector{\lambda}_{-}}{ \in \rtypo{\mathbb{R}_-^{d_{\mathbb{X}} \times d_{\mathbb{Y}}}}}
\end{maxi}}
but it can then be simplified given the closed form solution as follows:
\rtypo{\begin{maxi}|l|[3]
  {\vector{\lambda}_{\vector{a}}, \vector{\lambda}_{\vector{b}}, \vector{\lambda}_{-}}{\langle \vector{\lambda}_{\vector{a}}, \vector{a} \rangle + \langle \vector{\lambda}_{\vector{b}}, \vector{b} \rangle}{}{}
  \addConstraint{\matrix{C} - \vector{\lambda}_{\vector{a}} \vector{1}^T - \vector{1} \vector{\lambda}_{\vector{b}}^T + \vector{\lambda}_{-}}{ = \vector{0}}
  \addConstraint{\vector{\lambda}_{-}}{ \in \rtypo{\mathbb{R}_-^{d_{\mathbb{X}} \times d_{\mathbb{Y}}}}}
\end{maxi}}
Simplifying this further by eliminating $\vector{\lambda}_{-}$ from the constraints, we can write the same problem using inequality constraints:
\rtypo{\begin{maxi}|l|[3]
  {\vector{\lambda}_{\vector{a}}, \vector{\lambda}_{\vector{b}}}{\langle \vector{\lambda}_{\vector{a}}, \vector{a} \rangle + \langle \vector{\lambda}_{\vector{b}}, \vector{b} \rangle}{}{}
  \addConstraint{\matrix{C} - \vector{\lambda}_{\vector{a}} \vector{1}^T - \vector{1} \vector{\lambda}_{\vector{b}}^T}{ \geq \vector{0}}
\end{maxi}}

The main advantage of the dual optimal transport problem is that the decision variables now scale linearly with the number of points in the supports of the measures \rtypo{$O(d_{\mathbb{X}} + d_{\mathbb{Y}})$}. Additionally, the local constraints in the dual formulation over each $(\vector{x}_i, \vector{y}_j)$ pair enable certain stochastic optimization methods, \rtypo{e.g., the stochastic Frank-Wolfe algorithm}.

\subsection{Dual of entropy regularized optimal transport problem}

The final example of duality and its application in optimal transport which we shall present is that of the entropy regularized optimal transport problem:
\begin{mini}|l|[3]
  {\matrix{P}}{\langle \matrix{P}, \matrix{C} \rangle - \lambda H(\matrix{P})}{}{}
  \addConstraint{\matrix{P} \vector{1}}{ = \vector{a}}
  \addConstraint{\matrix{P}^T \vector{1}}{ = \vector{b}}
\end{mini}
where $H(\matrix{P}) = -\sum_{ij} \matrix{P}_{ij} (\log \matrix{P}_{ij} - 1)$. The partial derivative of $H$ with respect to $\matrix{P}_{ij}$ is:
\begin{align*}
\frac{\partial H}{\partial \matrix{P}_{ij}} = - \log \matrix{P}_{ij}
\end{align*}

The Lagrangian relaxation of the above convex program is:
\rtypo{\begin{mini}|l|[3]
  {\matrix{P}}{\langle \matrix{P}, \matrix{C} \rangle - \lambda H(\matrix{P})}{}{}
  \breakObjective{ + \langle \vector{\lambda}_{\vector{a}}, \vector{a} - \matrix{P} \vector{1} \rangle + \langle \vector{\lambda}_{\vector{b}}, \vector{b} - \matrix{P}^T \vector{1} \rangle}
  \addConstraint{\vector{\lambda}_{\vector{a}}}{ \in \mathbb{R}^{d_{\mathbb{X}}}}
  \addConstraint{\vector{\lambda}_{\vector{b}}}{ \in \mathbb{R}^{d_{\mathbb{Y}}}}
\end{mini}}
which can be re-written as:
\rtypo{\begin{mini}|l|[3]
  {\matrix{P}}{L(\matrix{P}, \vector{\lambda}_{\vector{a}}, \vector{\lambda}_{\vector{b}})}{}{}
  \breakObjective{ = \langle \matrix{P}, \matrix{C} - \vector{\lambda}_{\vector{a}} \vector{1}^T - \vector{1} \vector{\lambda}_{\vector{b}}^T \rangle}
  \breakObjective{ + \langle \vector{\lambda}_{\vector{a}}, \vector{a} \rangle + \langle \vector{\lambda}_{\vector{b}}, \vector{b} \rangle}
  \breakObjective{ - \lambda H(\matrix{P})}
\end{mini}}

This is just minimizing an unconstrained strongly convex function of $\matrix{P}$. The unique optimal solution \rtypo{$\matrix{P}^*(\vector{\lambda}_{\vector{a}}, \vector{\lambda}_{\vector{b}})$} minimizing $L$ can be obtained using the stationarity conditions:
\rtypo{\begin{align*}
\lambda \log \matrix{P^*}_{ij} + \matrix{C}_{ij} - \vector{\lambda_{a_i}} - \vector{\lambda_{b_j}} & = 0 \\
\matrix{P^*}_{ij} & = e^{\frac{1}{\lambda} (\vector{\lambda}_{\vector{a}_i} + \vector{\lambda}_{\vector{b}_j} - \matrix{C}_{ij})}
\end{align*}}

The corresponding optimal value \rtypo{$L^*(\vector{\lambda}_{\vector{a}}, \vector{\lambda}_{\vector{b}})$} is therefore:
\rtypo{\begin{multline}
  L^*(\vector{\lambda}_{\vector{a}}, \vector{\lambda}_{\vector{b}}) = \langle \vector{\lambda}_{\vector{a}}, \vector{a} \rangle + \langle \vector{\lambda}_{\vector{b}}, \vector{b} \rangle \\ + \sum_{ij} \matrix{P^*}_{ij} \times (\matrix{C}_{ij} - \vector{\lambda}_{\vector{a}_i} - \vector{\lambda}_{\vector{b}_j} + \lambda \log \matrix{P^*}_{ij} - \lambda)
\end{multline}}
Since \rtypo{$\lambda \log \matrix{P^*}_{ij} = -\matrix{C}_{ij} + \vector{\lambda}_{\vector{a}_i} + \vector{\lambda}_{\vector{b}_j}$} (from the stationarity conditions above), the above expression simplifies to:
\rtypo{\begin{align*}
  L^*(\vector{\lambda}_{\vector{a}}, \vector{\lambda}_{\bm{b}}) & = \langle \vector{\lambda}_{\vector{a}}, \vector{a} \rangle + \langle \vector{\lambda}_{\vector{b}}, \vector{b} \rangle - \lambda \sum_{ij} \matrix{P^*}_{ij} \\
  & = \langle \vector{\lambda}_{\vector{a}}, \vector{a} \rangle + \langle \vector{\lambda}_{\vector{b}}, \vector{b} \rangle - \lambda \sum_{ij} e^{\frac{1}{\lambda} (\vector{\lambda}_{\vector{a}_i} + \vector{\lambda}_{\vector{b}_j} - \matrix{C}_{ij})}
\end{align*}}

Let \rtypo{$\matrix{R}_{\lambda}(\vector{\lambda}_{\vector{a}}, \vector{\lambda}_{\vector{b}})$} be the matrix whose $(i,j)^{th}$ element is \rtypo{$-\lambda e^{\frac{1}{\lambda} (\vector{\lambda}_{\vector{a}_i} + \vector{\lambda}_{\vector{b}_j} - \matrix{C}_{ij})}$}. The Lagrangian dual problem can therefore be written as:
\rtypo{\begin{maxi}|l|[3]
  {\vector{\lambda}_{\vector{a}}, \vector{\lambda}_{\vector{b}}}{\langle \vector{\lambda}_{\vector{a}}, \vector{a} \rangle + \langle \vector{\lambda}_{\vector{b}}, \vector{b} \rangle + \langle \matrix{R}_{\lambda}(\vector{\lambda}_{\vector{a}}, \vector{\lambda}_{\vector{b}}), \vector{1} \rangle}{}{}
\end{maxi}}
This is an unconstrained maximization problem amenable to classical stochastic gradient descent and its family of stochastic optimization algorithms.

\bibliographystyle{ieeetr}
\bibliography{references}


\vspace{-4em}

\begin{IEEEbiography}
[{\includegraphics[width=1in,height=1.25in,clip,keepaspectratio]{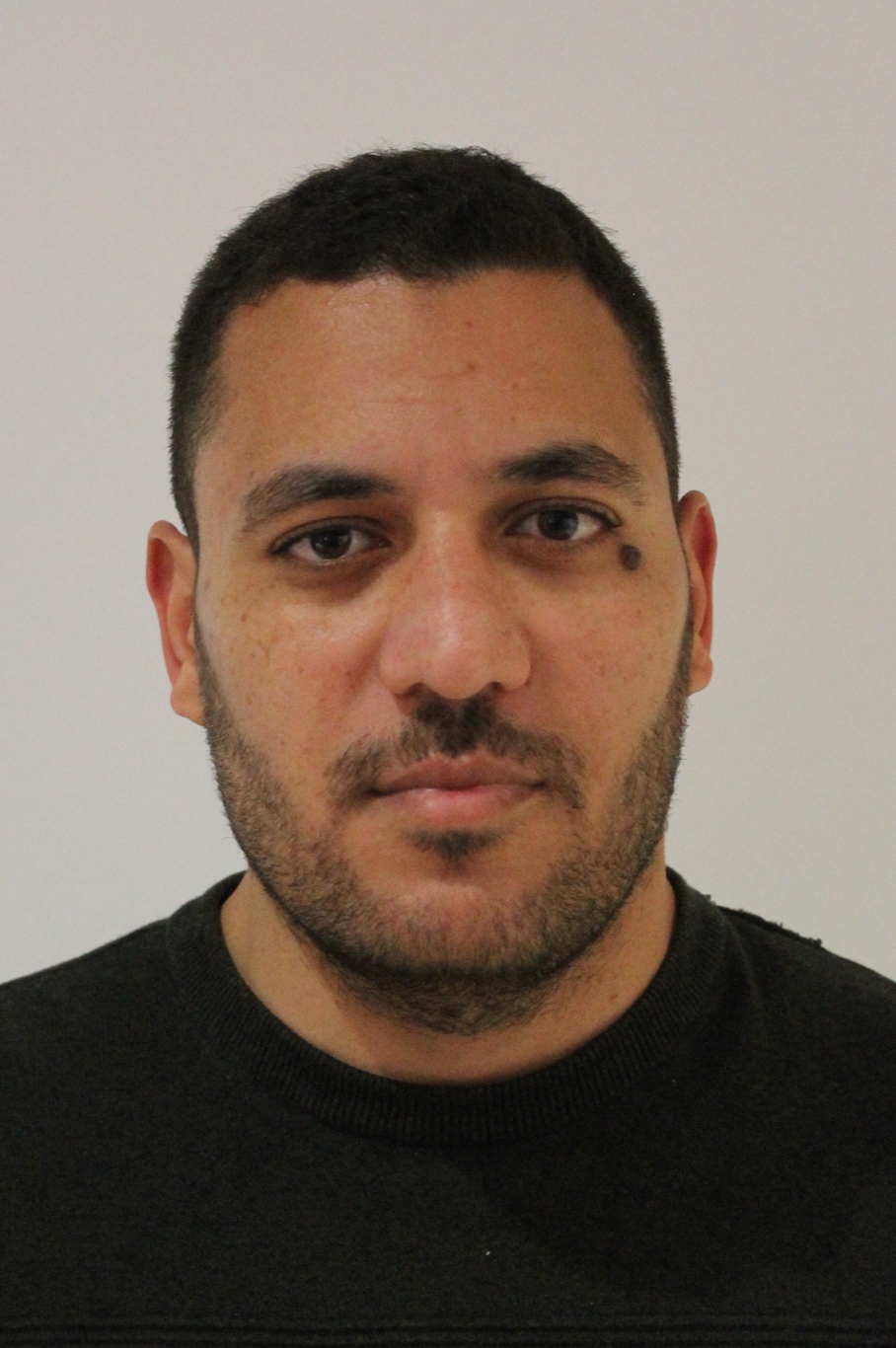}}]{Abdelwahed Khamis, PhD}
Dr. Abdelwahed Khamis is a research scientist at the Commonwealth Scientific and Industrial Research Organisation (CSIRO), Data61 in Australia. Abdelwahed completed his Ph.D. in Computer Science and Engineering from UNSW, Sydney in 2020.  His current research interests include AI on the edge,  RF and device-free sensing, and optimal transport. 
\vspace{-4em}
\end{IEEEbiography}
\begin{IEEEbiography}[{\includegraphics[width=1in,height=1.25in,trim={8cm 8cm 8cm 8cm},clip,keepaspectratio]{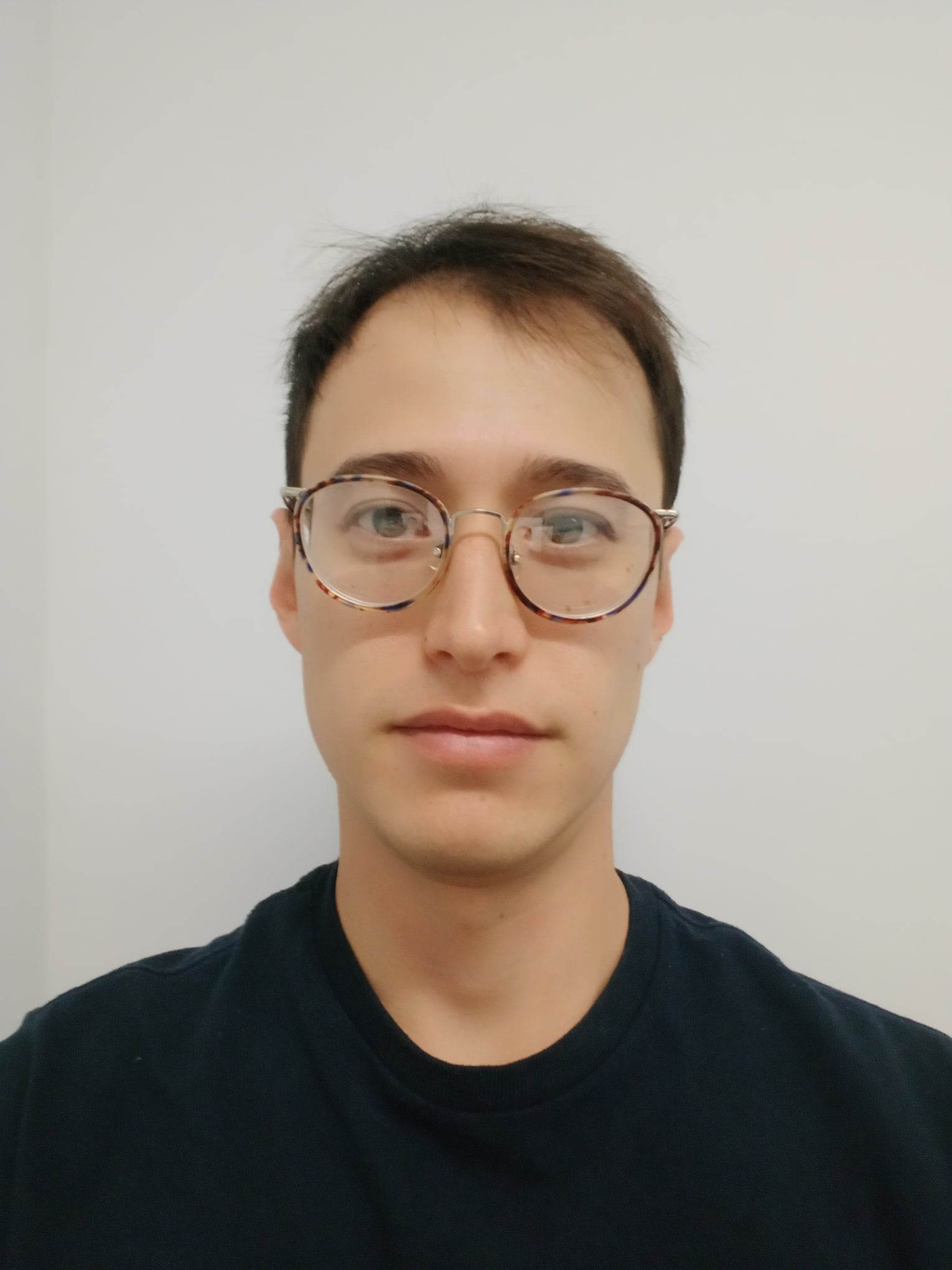}}]{Russell Tsuchida, PhD} Dr. Russell Tsuchida is a research scientist at the Commonwealth Scientific and Industrial Research Organization (CSIRO), Data61. Russell received his PhD from The University of Queensland (2020). His research interests include kernel methods, Gaussian processes, probabilistic machine learning and neural networks.
\vspace{-6em}
\end{IEEEbiography}
\begin{IEEEbiography}[{\includegraphics[width=1in,height=1.25in,clip,keepaspectratio]{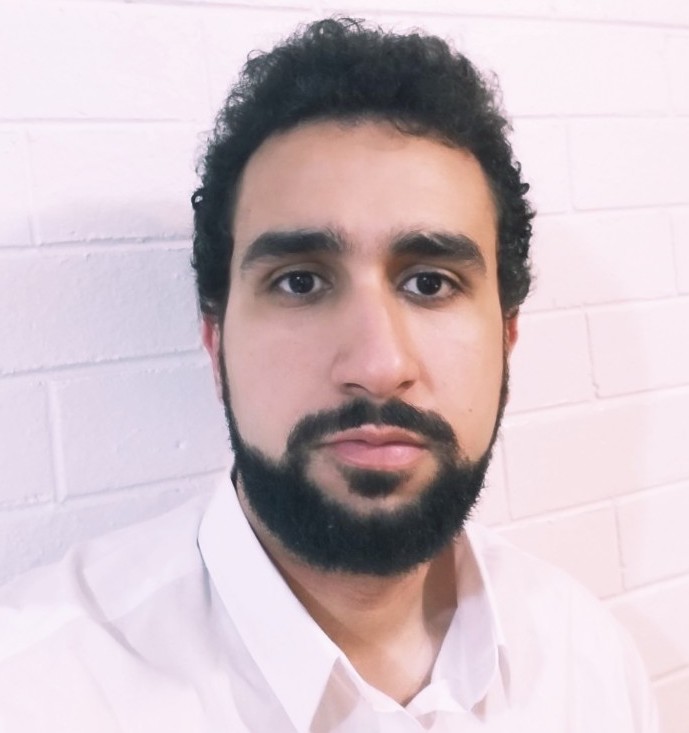}}]{Mohamed Tarek, PhD}
Dr. Mohamed Tarek is a senior product engineer at Pumas-AI Inc. and a research affiliate at the University of Sydney Business School. Mohamed received his PhD in Computer Science from University of New South Wales (2022). His current research interests include: statistical learning of hierarchical models, Bayesian inference and applied mathematical optimization.
\vspace{-6em}
\end{IEEEbiography}

\begin{IEEEbiography}[{\includegraphics[width=1in,height=1.25in,clip,keepaspectratio]{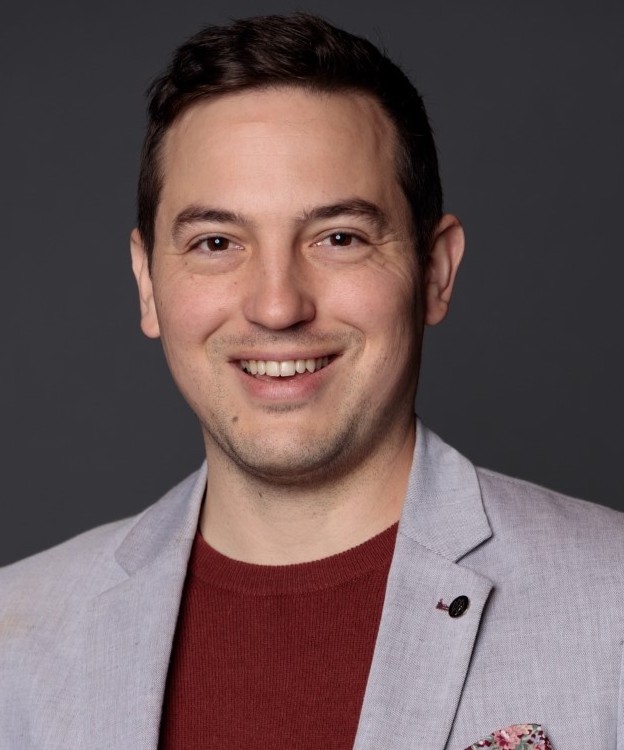}}]{Vivien Rolland, PhD} Dr. Vivien Rolland is a Senior Research Scientist at the Commonwealth Scientific and Industrial Research Organisation (CSIRO), Agriculture and Food, where he is using machine learning and computer vision to accelerate the rate of innovation in the Agrifood sector. 
\vspace{-6em}
\end{IEEEbiography}

\begin{IEEEbiography}[{\includegraphics[width=1in,height=1.25in,clip,keepaspectratio]{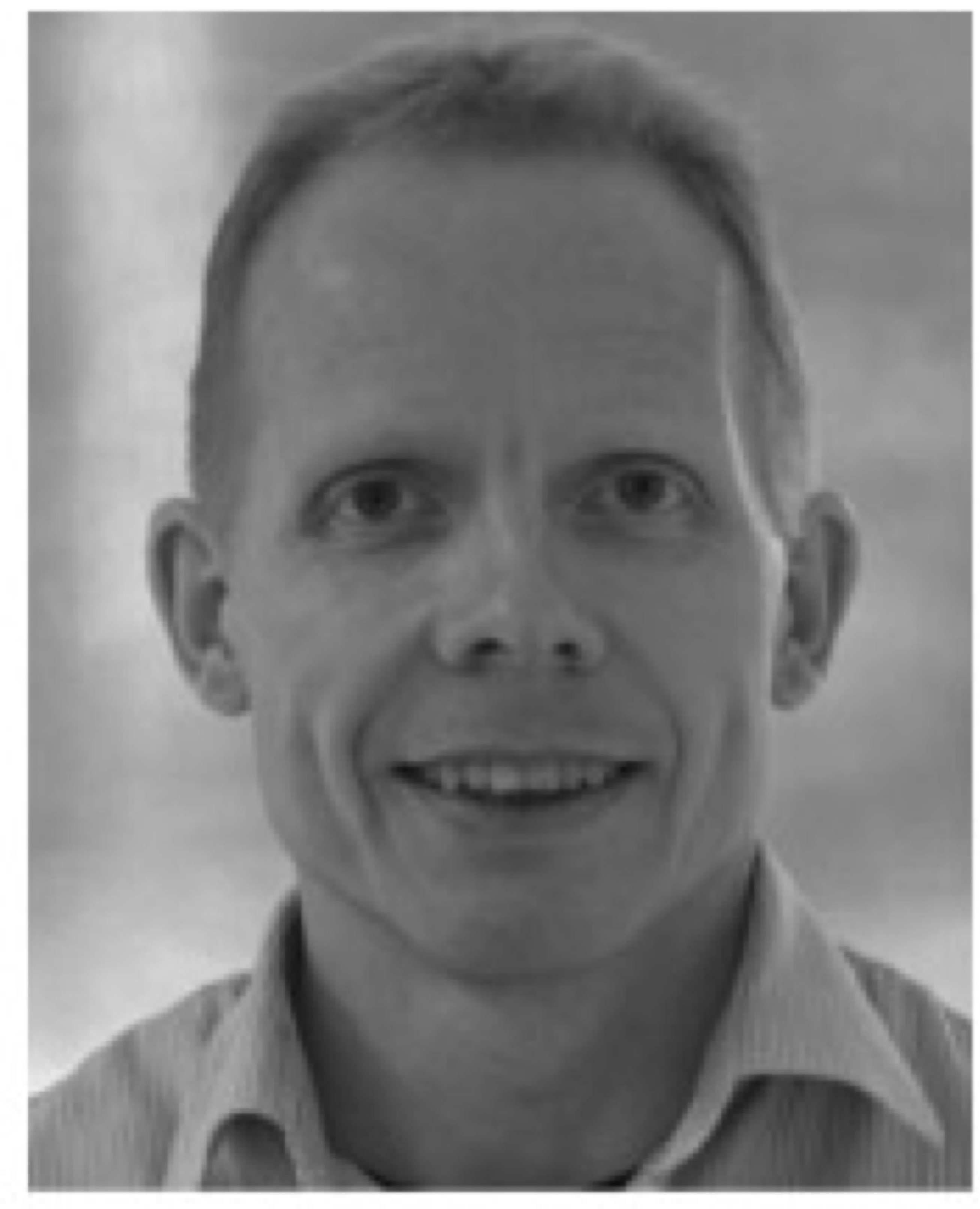}}]{Lars Petersson, PhD}
Dr. Lars Petersson is a Senior Principal Research Scientist leading the Imaging and Computer Vision Group in Data61, CSIRO, Australia. His research interests span machine learning and computer vision across a number of domains, from the smallest of microscopy scales to the largest of astronomical scales. 
He received his PhD in March 2002 from KTH, Stockholm, Sweden, where he also received his Master’s degree in Engineering Physics.
\end{IEEEbiography}

\end{document}